\documentclass[5p,final,authoryear,onecolumn]{elsarticle}

\usepackage{siunitx}
\usepackage{amssymb}
\usepackage{graphicx}
\usepackage{amsmath}

\usepackage[latin1]{inputenc}
\usepackage{color, colortbl}
\usepackage{multirow}
\usepackage[hypcap=false]{caption}

\usepackage{url}
\usepackage{subcaption}
\usepackage{xcolor}

\usepackage{breqn}
\usepackage{rotating}

\def\zz#1{%
	\ifdim#1pt<0.001pt\cellcolor{blue!40}\else
	\ifdim#1pt<0.01pt\cellcolor{blue!20}\else
	\ifdim#1pt<0.05pt\cellcolor{blue!10}\else
	\fi\fi\fi
	#1}

\begin{document}
\begin{frontmatter}
\let\ACMmaketitle=\maketitle
\title{Gaussian Process Uncertainty in Age Estimation as a  Measure of Brain Abnormality \corref{accept} } 
\author[aimed,camp]{Benjamin Gutierrez Becker\corref{cor1}}
\ead{benjamin.gutierrez\_becker@med.uni-muenchen.de}
\author[sap]{Tassilo Klein}
\ead{tassilo.klein@sap.com}
\author[aimed]{Christian Wachinger}
\ead{christian.wachinger@med.uni-muenchen.de}
\author[adni]{\\for the Alzheimer's Disease Neuroimaging Initiative}
\cortext[accept]{\textbf{This paper has been accepted for publication in Neuroimage.}}

\cortext[cor1]{Corresponding author.  Waltherstrasse 23. KJP Forschungsabteilung. 80337, Munich, Germany}
\address[aimed]{Artificial Intelligence in Medical Imaging (AI-Med), Department of Child and Adolescent Psychiatry, LMU M\"unchen, Germany}
\address[sap]{SAP SE Berlin, Germany}
\address[camp]{CAMP, Technische Universit\" at M\" unchen, Germany}

\cortext[adni]{Data used in preparation of this article were obtained from the Alzheimer's Disease
	Neuroimaging Initiative (ADNI) database (adni.loni.usc.edu). As such, the investigators
	within the ADNI contributed to the design and implementation of ADNI and/or provided data
	but did not participate in analysis or writing of this report. A complete listing of ADNI
	investigators can be found at:
	http://adni.loni.usc.edu/wp-content/uploads/how\textunderscore to\textunderscore apply/ADNI\textunderscore Acknowledgement\textunderscore List.pdf}

\begin{abstract}

Multivariate regression models for age estimation are a powerful tool for assessing abnormal brain morphology associated to neuropathology.  
Age prediction models  are built on cohorts of healthy subjects and are built to reflect normal aging patterns. The application of these multivariate models to diseased subjects usually results in high prediction errors, under the hypothesis that neuropathology presents a similar degenerative pattern as that of accelerated aging.
In this work, we propose an alternative to the idea that pathology follows  a similar trajectory than normal aging. Instead, we propose the use of metrics which measure deviations from the mean aging trajectory. We propose to measure these deviations using two different metrics: uncertainty in a Gaussian process regression model and a newly proposed age weighted  uncertainty measure.
Consequently, our approach assumes that pathologic brain patterns are different to those of normal aging. 
We present results for subjects with autism, mild cognitive impairment and Alzheimer's disease to highlight the versatility of the approach to different diseases and age ranges. We evaluate volume, thickness, and VBM features for quantifying brain morphology. Our evaluations are performed on a large number of images obtained from a variety of publicly available neuroimaging databases. Across all features,  our uncertainty based measurements yield a better separation between diseased subjects and healthy individuals than the prediction error. Finally, we illustrate differences in the disease pattern to normal aging, supporting the application of uncertainty as a measure of neuropathology. 

\end{abstract}
\end{frontmatter}
\definecolor{lightgray}{gray}{0.95}


\newcommand{\cov}  {\textbf{cov}  (\mathbf{\hat{y}} )}
\newcommand{\covw} {\textbf{cov}_w(\mathbf{\hat{y}} )}

\section{Introduction}
The brain is a complex organ whose morphology varies substantially across the population.
The causes of  morphological variation have not yet been fully understood, but several studies have reported on potential causal factors including age \citep{Guttmann1998, Franke2010, Ziegler2012,Wachinger2015}, sex \citep{ingalhalikar2014}, pathologies like dementia \citep{Gaser2013, Wachinger2016}, and even environmental factors such as education and physical activity \citep{steffener2016}. Among all these variables, age was shown to be the main factor determining brain morphology \citep{Potvin2017}. 
Due to the wide impact of aging on brain morphology, multivariate regression methods using features based on brain morphology can in turn be used to estimate a subject's age.
A recent volume of work has focused on modeling the normal aging of healthy individuals to predict a subject's age. Obtaining a prediction of the age with imaging features was shown to be useful to derive imaging biomarkers, which can potentially be used to predict brain anomaly caused by disease \citep{Cole2017}.

\begin{figure*}[h!]
	\begin{minipage}[b]{0.46\linewidth}
		\centering\includegraphics[width=\textwidth]{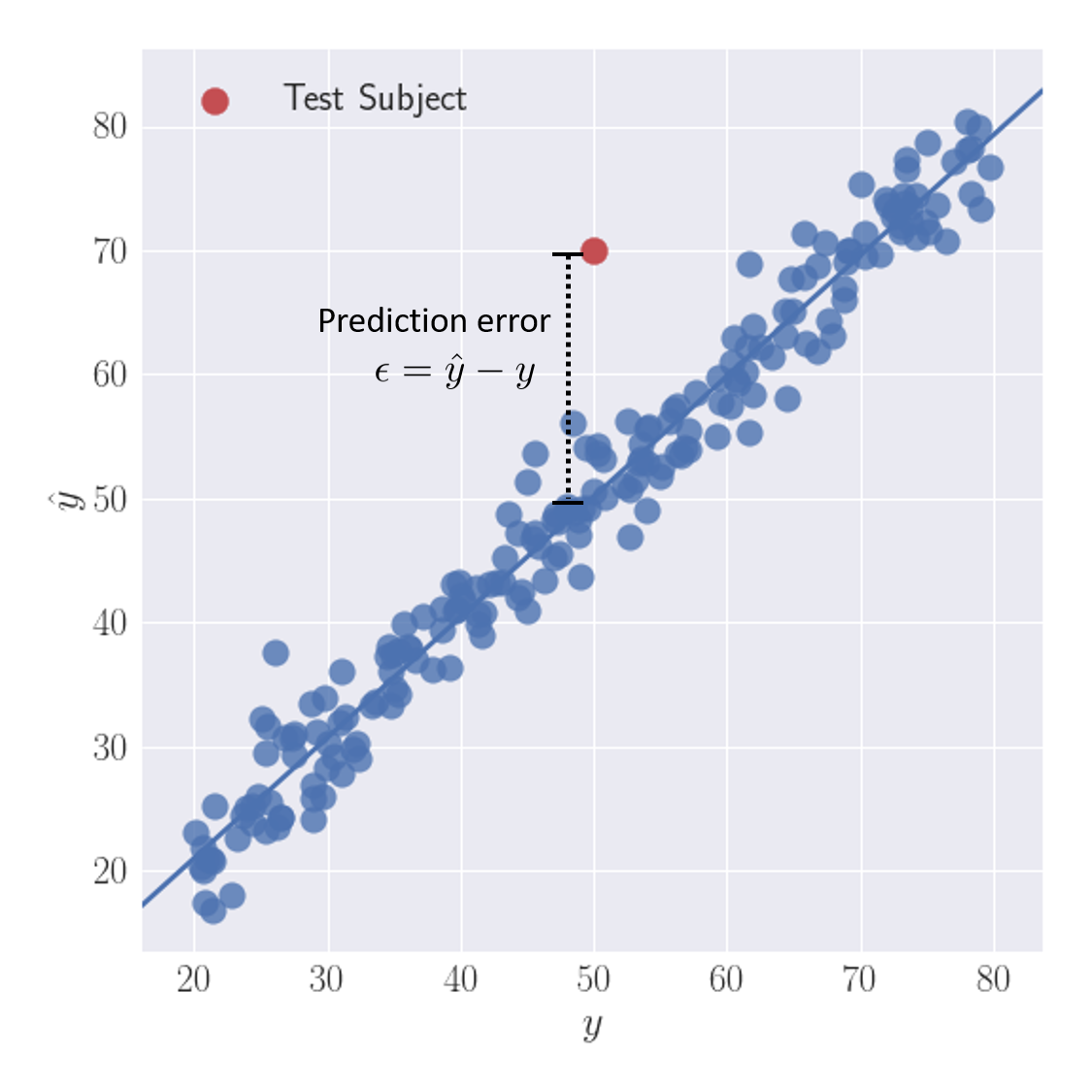}
		\subcaption{\textbf{Prediction error $\epsilon$}. A series of data points are plotted showing their chronological age $y$ and their predicted biological age $\hat{y}$. Training points used to train the model are shown in blue. Prediction error $\epsilon$ for a test point is defined as the difference between its predicted biological age and its chronological age.  } 
		\label{fig:prediction_error}
	\end{minipage}
	\hfill
	\begin{minipage}[b]{0.50\linewidth}
		\centering\includegraphics[width=\textwidth]{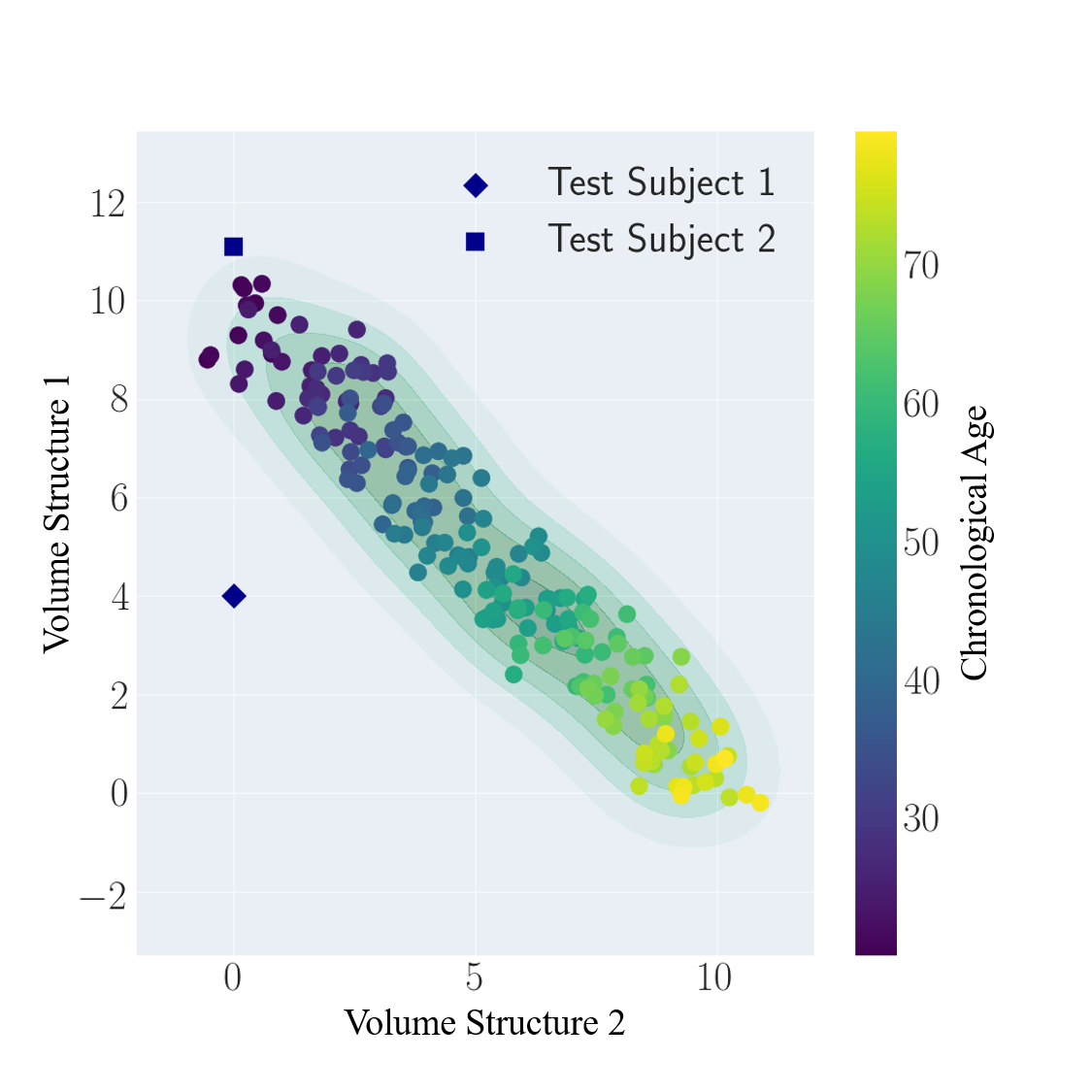}
		\subcaption{\textbf{Prediction uncertainty $\cov$}. Data points are plotted on the feature space determined by two volumetric features. Circles correspond to training subjects and non circles to testing subjects. Uncertainty $\cov$ measures the distance of a testing point to all training points in the feature space. The background color corresponds to uncertainty in the feature space. Areas close to the training points have lower degrees of uncertainty. The color of each subject encodes its chronological age. } \label{fig:prediction_uncertainty}
	\end{minipage} \\
	
	\begin{minipage}[b]{1\linewidth}
		\centering\includegraphics[width=0.5\textwidth]{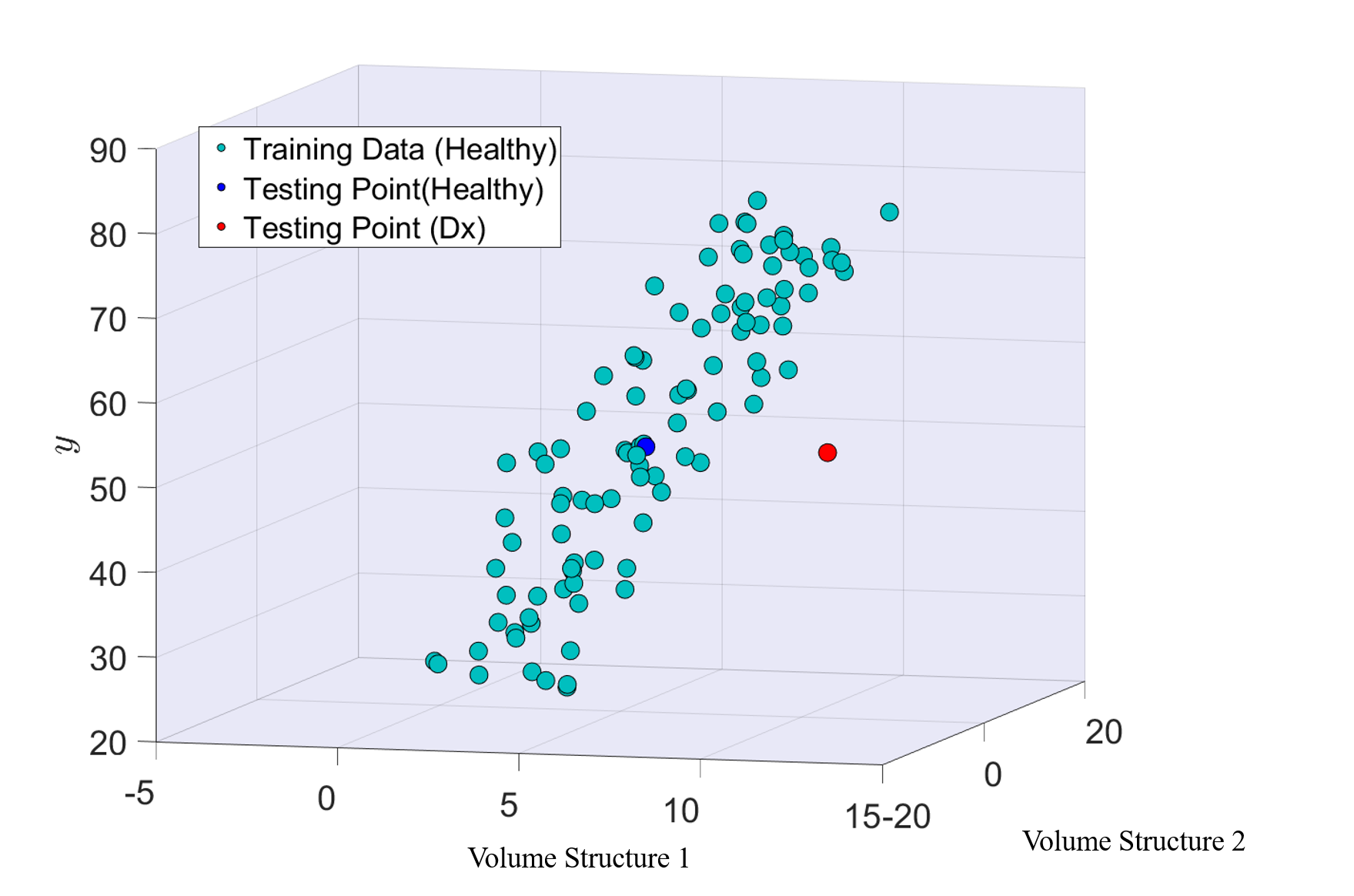}
		\subcaption{\textbf{Weighted Uncertainty $\covw$}. Data points are shown on an extended feature space given by two volumetric features and their corresponding chronological age $y$. Weighted uncertainty $\covw$ measures the distance of a testing point to all training points in this extended feature space.  }\label{fig:weighted_uncertainty}
	\end{minipage}
	
	\caption{Comparison between the three evaluated anomaly metrics: prediction error $\epsilon$, GPR uncertainty $\cov$ and GPR age-weighted uncertainty $\covw$.  }\label{fig:Gaussian_fit}
\end{figure*}

The task of predicting age from brain images has been formulated as multivariate regression, where a predictive model is trained to relate structural information obtained from brain MR images  to the chronological age of healthy subjects \citep{Gaser2013, Wang2014, Kondo2015, Valizadeh2017, liem2017predicting}. The prediction from these models is interpreted as an estimate of a subject's biological age, in contrast to a subject's chronological age. Of particular interest is the \emph{prediction error}, which is defined as the difference between the biological and chronological age (figure \ref{fig:prediction_error}). 
When predicting the age of healthy subjects, the prediction error is assumed to be small, while the prediction on subjects with neuropathology is assumed to result in large positive prediction errors. The error could  therefore serve as a personalized marker of pathological processes \citep{Franke2010, Gaser2013}. 
The main assumption behind using the prediction error as a measure of pathology  is that changes caused by neuropathology are equivalent to an accelerated aging process.  Following this hypothesis, \cite{Gaser2013} and \cite{Habes2016} showed that changes related to Alzheimer's disease (AD) resemble \emph{accelerated aging},   since differences between biological and chronological age are larger for individuals with AD than for healthy controls. Similar results on age differences have also been reported for individuals diagnosed with schizophrenia \citep{Nenadic2017} and depression \citep{koutsouleris2013accelerated}.  In these studies, the age prediction model is trained using only images from healthy individuals. This means that contrary to their discriminative counterparts, a single age prediction model can be used to assess differences between healthy controls and individuals diagnosed with different conditions.

Although the findings of these studies show the big potential of using models of healthy aging to assess brain abnormality,  a potentially limiting factor when quantifying neuropathology through the difference between chronological and predicted age, is the assumption that morphological changes caused by disease follow an accelerated aging process. The assumption that brain anomaly is equivalent to accelerated aging may hold true for specific brain regions that accommodate neural systems with high susceptibility to deleterious factors, which are therefore affected by aging and disease processes. However, the assumption of accelerated aging does likely not extend to the whole brain,  given that differences in brain morphology are caused by a variety of neurobiological processes that are complex and non-linear  \citep{Fjell2014,buckner2004,hedden2004}. This is potentially problematic, as current approaches for predicting the brain age are based on multivariate regression models that operate on gray matter maps or morphological features across the entire brain.

In this work, we build on the idea of modeling neuropathology as deviations from the healthy development of the brain. 
Our main hypothesis is that disease and aging result in brain-wide patterns of change. These patterns are not independent from each other and are essentially similar for several brain regions where disease results in patterns resembling accelerated aging. However, these accelerated aging patterns do not extend to the whole brain, making the assessment of deviations from healthy aging solely through prediction error problematic. We propose instead the use of  Gaussian process regression (GPR). GPR  can measure how a new subject deviates from previous observations used to construct the model by means of the posterior prediction uncertainty. Different to prediction error, GPR uncertainty is able to measure deviations from the healthy aging model without the implicit assumption of a brain-wide accelerated aging pattern.  Particularly for the task of age prediction, we introduce a variation to traditional Gaussian processes regression that takes the known chronological age into account. This modification yields a weighted uncertainty measure. We evaluate our new method on a large collection of images obtained from several   public datasets for assessing the variation to normal aging in mild cognitive impairment, Alzheimer's disease, and autism. Our results support the use of Gaussian process uncertainty and the age weighted uncertainty as tools to measure neuropathological  patterns that deviate from  healthy aging. Similar to previous age prediction models, our evaluations are done with a single age prediction model which is trained only on healthy controls, showing its versatility across different age ranges and diseases.

\setcounter{subfigure}{0}
     
\section{Materials and Methods}\label{sec:methods}

\subsection{Method overview} \label{sec:method-overview}

\begin{figure*}[h!]
	\centering\includegraphics[width=\textwidth]{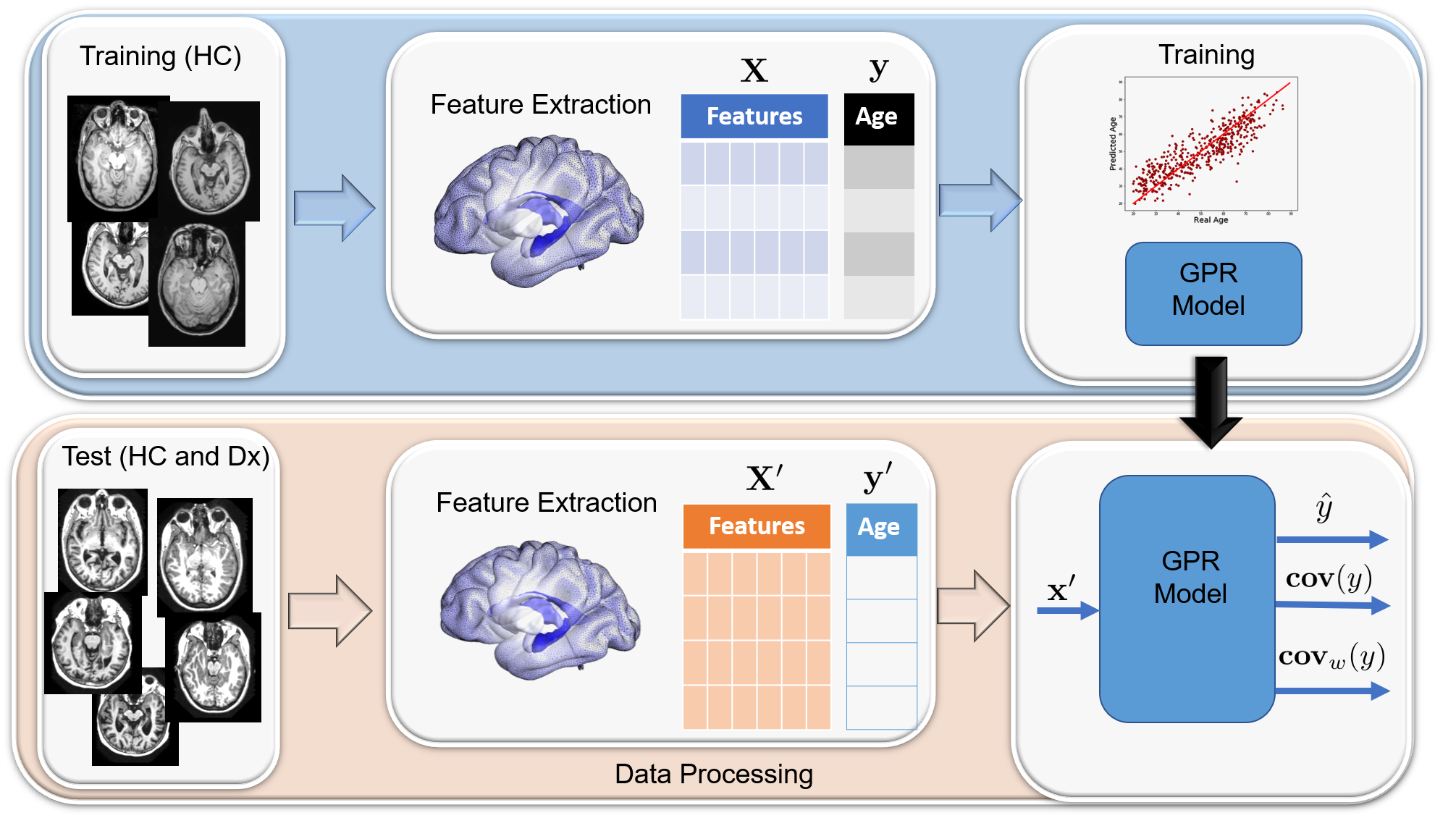}
	\caption{Overview of our brain anomaly prediction model. The top part corresponds to the training stage where a set of images from healthy individuals is used to build a GPR age prediction model. The bottom part corresponds to the age prediction stage where the GPR model is used to predict the age of a set of test images. A predicted  age $\hat{y}$ as well as the uncertainty measures $\cov$ and $\covw$ are obtained. These measurements can be used to find differences between the HC and Dx groups. }\label{fig:overview}
\end{figure*}

In this section, we describe our method for assessing neuropathology based on GPR uncertainty. Figure \ref{fig:overview}  presents an overview of our method, which consists of two stages. In the training stage (top section of figure \ref{fig:overview}), we build a GPR  model that estimates the chronological age of healthy subjects. This model is built using  a dataset of MRI scans of healthy controls (section \ref{sec:data}). Images are processed and segmented to extract a set of features describing brain morphology (section \ref{sec:features}). Finally, a GPR model mapping the extracted features to a predicted age is trained on these features (section \ref{sec:regression}).

In the testing stage (bottom section of figure \ref{fig:overview}), we use the GPR model trained on healthy subjects to quantify deviations from the normal aging pattern on previously unseen subjects. In this stage, morphological features are extracted from the MR images of the test subjects, and these features are then used to obtain an estimate of the age of the subject using the GPR model. From the GPR model, we obtain the estimated age $\hat{y}$, an uncertainty measure of the estimation $\cov$, and a weighted uncertainty measure $\covw$ (see section \ref{sec:gpr} for details on these measurements). We will show in our experiments (section \ref{sec:results}) that these measurements based on the uncertainty of the GPR model can be used to assess the similarity between subjects in the testing set and the healthy population in the training set.
\setcounter{subfigure}{0}

\subsection{Data} \label{sec:data}
Similar to previous work on age prediction, we train an age regression model based on T1-MR images of healthy individuals. The training images for our age regression model are extracted from three different databases:  IXI \footnote{http://brain-development.org/ixi-dataset/}, ABIDE~\citep{ABIDE}, and AIBL~\citep{aibl}. Details for each training dataset are shown in  table \ref{table:datasets-training}. We perform evaluation on 3 different test datasets summarized in table \ref{table:datasets-testing}. The training and testing groups are therefore extracted from different databases and are independent from each other. For the first and second groups, obtained from the OASIS \citep{oasis} and ADNI \citep{adni} datasets, we aim at finding differences between Healthy Controls (HC), individuals diagnosed with Mild Cognitive Impairment (MCI), and subjects diagnosed with Alzheimer's disease (AD). For the third dataset, ABIDE II, we look for differences between HC and individuals diagnosed with autism. In total, 1,543 images were used for training and  4,819 images for testing. 

\begin{table*}[ht]
	\centering
	\begin{tabular}{ l c  c  c  c}
		Training Dataset    & No. Images  & Female/Male  &  Age (Min-Max) & Age Quantiles \\ 
		\rowcolor{lightgray} 
		IXI  & 581 &  311/270   & 19-87  & 33.7 - 48.6 - 62.2  \\ 
		ABIDE I& 573 &  99/474   &  6-64  & 14.6 - 17.0 - 20.1 \\
		\rowcolor{lightgray}
		AIBL & 409&  209/200    & 55-92  & 67.0 - 73.0 - 79.0\\
		
	\end{tabular}
	\caption{Summary of the datasets used for training the age prediction model.}
	\label{table:datasets-training}
\end{table*}

\begin{table*}[ht]
	\centering
	\resizebox{\columnwidth}{!}
	{
	\begin{tabular}{ l c  c  c  c c}
		Testing Dataset    & No. Images  & Female/Male  &  Age (Min-Max)& Age Quantiles & Target Dx\\ 
		\hline
		\rowcolor{lightgray} 
		ADNI  & 3591 &  1422/2169  & 54-90& 71.2 - 75.1 -79.7  & MCI - AD \\ 
		OASIS & 196 &  129/67   &  60-82 & 71.0 - 76.0 - 82.0 & MCI - AD \\
		\rowcolor{lightgray}
		ABIDE II &  1032   & 247/785  & 5-64& 9.5 - 11.4 - 15.2 & Autism  \\
	\end{tabular}
	}
	\caption{Summary of the datasets used for testing.}
	\label{table:datasets-testing}
\end{table*}
\def\real{{\mathbb{R}}}
\subsection{Feature Extraction} \label{sec:features}

As mentioned in the overview section of our method, we require to extract  features from structural MR images to quantify the brain morphology. Several image based features have previously been used for the age estimation task: \cite{Good2001}, \cite{Franke2010} and \cite{Gaser2013} used Voxel Based Morphometry (VBM) features, which identify differences on the composition of brain tissue by registering all structural images to the same space. After segmenting gray and white matter, the voxel values of the gray matter extracted images are used as features. Finally the dimensionality of the feature space is reduced using Principal Component Analysis (PCA) and keeping only the principal modes of variation.   \cite{Valizadeh2017} and \cite{Wang2014} use volumetric, thickness and curvature measurements of the brain, derived from the brain segmentation with FreeSurfer \citep{Fischl2012}. These different sets of features have been presented in separate studies but, to the best of our knowledge, have not yet been directly compared on the same age estimation task.
In summary, we use three different types of features in our approach:

\begin{itemize}
	\item VBM features (50 Principal Components), 
	\item Thickness of 70 cortical structures. 
	\item Volume of 50 brain structures.
\end{itemize}
Additionally we build a prediction model combining VBM, thickness and volume features together. VBM features were extracted using the CAT12  toolbox \footnote{http://www.neuro.uni-jena.de/cat/} together with the  SPM12 toolbox \footnote{http://www.fil.ion.ucl.ac.uk/spm/} for segmentation. The preprocessing of the images, the segmentation of gray matter, and post processing were implemented in line with the pipeline proposed by \cite{Franke2010}. Dimensionality reduction was performed using the PCA library included in the scikit-learn toolbox \citep{scikit}. The principal component directions were estimated using only the training sample, and the testing data was projected to this estimated lower dimensional space.
For all the analyses on thickness and volume features, FreeSurfer version 5.3 was used. The default Deskian/Killiany atlas was used for the parcellation to obtain thickness measurements. We are using all subcortical volume measurements as provided by FreeSurfer and described in the FreeSurfer subcortical segmentation pipeline. 

\subsection{Uncertainty Estimation with Gaussian Process Regression} \label{sec:regression}
 
Several multivariate regression techniques have been previously used for the task of age prediction from brain MR images. A detailed comparison of the performance of neural networks, random forests, k-nearest neighbors, support vector machines, multiple linear regression and ridge regression was presented by \citep{Valizadeh2017}. In the work by  \cite{Franke2010} relevance vector regression was preferred. 

In our case, we are interested in modeling the age regression problem with a model that does not only provide estimates of the biological age, but also provides  uncertainties of these estimates. Gaussian process regression achieves a comparable accuracy in age regression than standard regression techniques, while offering the advantage of providing an estimate of the uncertainty of each prediction. GPR models have been used successfully before as age prediction models \citep{cole2015prediction, Cole2016}, but the potential of using uncertainty based measurements as biomarkers has not been explored yet. In this section, we will briefly introduce GPR models, focusing particularly on the calculation of uncertainty, where we refer the reader to \citep{Rasmussen2005} for a more detailed explanation, and introduce our modification for computing an age-weighted uncertainty. 

\subsubsection{Gaussian Process} \label{sec:gpr}

A Gaussian process is defined as a collection of random variables, any finite number of which have a joint Gaussian distribution \citep{Rasmussen2005} with:

\def\featurematrix{{\mathbf{X}}}
\def\bfx{{\mathbf{x}}}
\def\bigx{{\mathbf{X}}}
\begin{itemize}
 
 \item a mean function $m(\mathbf{x}) = \mathbb{E}[f(\mathbf{x})]$,   
 \item and a covariance function $k(\mathbf{x},\mathbf{x}')= 
 								 \mathbb{E}[f(\mathbf{x}) -m(\mathbf{x})
 								 			f(\mathbf{x'}) -m(\mathbf{x'})]$.
\end{itemize}
Although not necessary, it is often assumed that the mean function $m(\mathbf{x})$ of the GPR is zero. Therefore the design of a GPR is focused on the selection of an appropriate covariance function $k(\mathbf{x},\mathbf{x}')$ measuring the similarity between data points. This covariance function is equivalent to a similarity measure between two data points, giving small values for points that are close to each other and large values otherwise. In our case we define the covariance function as a squared exponential function of the form:

\begin{equation}
k(\bfx_i,\bfx_j) =  \sum_{k=1}^K  \exp \bigg[\frac{-(x^k_i-x^k_j)^2}{2l_k^2} \bigg] + \sigma_n^2 \delta(\bfx_i,\bfx_j), \label{eq:kernel}
\end{equation}

\noindent
where  $\sigma_n^2$ is the noise variance, $l^k$  is the length scale of the $k$-th feature, and $\delta$ is the   Kronecker delta function. We can think of the length scale vector $\mathbf{l}\in\real^K$ as a parameter controlling how close should two data points $x_i$ and $x_j$ should be in order to influence each other. In general, the smaller an element $l^k$ is, the more dependent $y$ is to the feature element $x_k$.

We model the joint distribution of the training and test outputs as:

\begin{equation}
\begin{bmatrix}
\mathbf{y} \\
\mathbf{y}'
\end{bmatrix}
 \sim
\mathcal{N}\bigg(0,\begin{bmatrix}
K(\bigx,\bigx)&K(\bigx,\bigx')\\
K(\bigx',\bigx)&K(\bigx',\bigx')
\end{bmatrix} \bigg).
\label{eq:Gaussian_process}
\end{equation}

The elements of the joint distribution in Eq.(\ref{eq:Gaussian_process}) can be summarized as follows:
\begin{itemize}
	\item an intra-covariance matrix of the training set $K(\bigx,\bigx)\in \real^{mxm}$ , 
	\item  an intra-covariance matrix of the testing set $K(\bigx',\bigx')\in \real^{nxn}$,
	\item  an inter-covariance matrix between the training and testing set $K(\bigx,\bigx')\in \real^{mxn}$,
	\item  a training labels  vector $\mathbf{y}\in \real^{m}$, and
	\item  a testing labels vector $\mathbf{y}'\in \real^{n}$, 
\end{itemize}
where $m$ corresponds to the number of training samples and $n$ to the number of testing samples.
The matrices $K(X,X)$ have the form:

\begin{equation}
K = \begin{bmatrix}
k(\bfx_{1},\bfx_{1} )  & k(\bfx_{1},\bfx_{2} ) & \dots  & k(\bfx_{1},\bfx_{n} ) \\
k(\bfx_{2},\bfx_{1} ) & k(\bfx_{2},\bfx_{2} )  & \dots  & k(\bfx_{2},\bfx_{n} ) \\
\vdots & \vdots &  \ddots & \vdots \\
k(\bfx_{m},\bfx_{1} ) & k(\bfx_{m},\bfx_{2} )  & \dots  & k(\bfx_{m},\bfx_{n} )
\end{bmatrix},
\end{equation}  
where each element $ k(\bfx_{1},\bfx_{2}) $ corresponds to a measure of the similarity between two feature vectors $x_i$ and $x_j$.  We are interested in predicting the values for the test labels $\mathbf{y}'$, which have the following conditional distribution:
\begin{dmath}
\mathbf{y}'|\mathbf{y},\bigx,\bigx' \sim \mathcal{N}(K(\bigx', \bigx), K(\bigx, \bigx)^{-1} \mathbf{y}, K(\bigx', \bigx')-K(\bigx', \bigx)K(\bigx, \bigx)^{-1}K(\bigx, \bigx')).
\end{dmath}  

Using this conditional distribution we can derive the predictive equations of a GPR:
\begin{equation}
\hat{\mathbf{y}'}=\mathbb{E}[\hat{\mathbf{y}'}|\bigx,\hat{\mathbf{y}},\bigx'] = K(\bigx',\bigx)K(\bigx,\bigx)^{-1}\mathbf{y} 
\end{equation}

\begin{equation}
\cov = K(\bigx',\bigx') - K(\bigx',\bigx)K(\bigx,\bigx)^{-1}K(\bigx,\bigx'), \label{eq:uncertainty}
\end{equation}  
  which correspond to the predicted labels and to the estimated covariance, respectively. The estimated covariance can also be thought as a measure of uncertainty for the predicted values. This uncertainty estimate is usually used in the GPR framework to measure the degree of confidence of a predicted value $\hat{y}'$ by measuring the similarity of a new observation with respect to the previous observations in the training set. 

When training a GPR model, the parameters $\theta=\{\mathbf{l},\sigma_n\}$ have to be tuned in order to fit the training data. This is done by maximizing the  marginal likelihood of the model given by:
\def\bfy{{\mathbf{y}}}
\begin{dmath}\label{eq:marginal_likelihood}  
\log(p(\mathbf{\bfy}|\mathbf{X},\theta))  = -\frac{1}{2}\bfy^T K(\bigx,\bigx) \bfy - \frac{1}{2} \log(K(\bigx,\bigx)) - \frac{n}{2}\log(2\pi). 
\end{dmath} 
By finding the parameters $\theta$ that maximize the marginal likelihood, we can obtain a GPR model that best fits the training data.

\subsubsection{Age-Weighted Uncertainty}
  
The uncertainty measurement of the GPR $\cov$ is solely defined with respect to the feature vectors $x$. In a common regression scenario this is a natural approach since the \emph{real} values of the labels are unknown. However, in the case of an age estimation framework, we do possess the real values of the labels, which correspond to the chronological age of the patient.

We can introduce the age information into the GPR framework by creating  age-weighted similarity matrices $K_w(\bigx,\bigx',\mathbf{y},\mathbf{y}')$. Similar to the GPR covariance matrices we can construct three different similarity matrices:

\begin{itemize}
	\item a weighted intra-similarity matrix for the training samples $K_w(\bigx,\bigx,\mathbf{y},\mathbf{y})\in \real^{mxm}$ , 
	\item  a weighted intra-similarity matrix for the testing samples  $K_w(\bigx',\bigx',\mathbf{y},\mathbf{y}') \in \real^{nxn}$, and  
	\item  a weighted inter-similarity matrix between the training  and test samples $K_w(\bigx,\bigx',\mathbf{y},\mathbf{y}')\in \real^{mxn}$.
\end{itemize}

These similarity matrices are constructed in the same manner as the covariance matrices presented in section \ref{sec:gpr}. The only difference consists in a modification of the kernel to take into account differences in age. This is achieved by creating an age weighted similarity kernel of the form:

 \begin{equation}
k_w(\bfx_i,\bfx_j,\bfy_i,\bfy_j) = s(\bfy_i,\bfy_j)  k(\bfx_i,\bfx_j),  \label{eq:weighted_kernel}
\end{equation} 
where $ k(\bfx_i,\bfx_j) $ corresponds to the kernel defined in Eq.(\ref{eq:kernel}) and $s(\bfy_i,\bfy_j)$ corresponds to an age similarity term defined as:
\begin{equation}
s(\bfy_i,\bfy_j) = \exp \bigg[\frac{-(\bfy_i-\bfy_j)^2}{2l_y^2} \bigg] + \sigma_y^2 \delta(\bfy_i,\bfy_j). \label{eq:age_similarity}
\end{equation}
where $l_y$ corresponds to the age length scale, which is a parameter controlling the effect of the age weighting. By using this updated kernel $k_w$, we  obtain a \emph{weighted uncertainty} term $\mathbf{cov}_w(y)$ which takes into account the age of the subjects to define similarities between subjects. This weighted uncertainty is obtained similar to the regular uncertainty presented in Eq. (\ref{eq:uncertainty}):

\begin{equation}
\textbf{cov}_w(\mathbf{y}') = K_w(\bigx',\bigx') - K_w(\bigx',\bigx)K_w(\bigx,\bigx)^{-1}K_w(\bigx,\bigx'). \label{eq:weighted_uncertainty}
\end{equation}

\subsection{Prediction Error, Uncertainty and Age-Weighted Uncertainty}

In this work, we compare three age regression based metrics in order to measure their usefulness as a biomarker to distinguish between healthy controls and subjects with different neuropathologies. These metrics are the commonly used prediction error $\epsilon=\hat{y}-y$  \citep{Franke2010}, the GPR uncertainty $\mathbf{cov}(y)$, and the GPR age-weighted uncertainty $\mathbf{cov}_w(y)$. As discussed in the introduction, the prediction error has previously been used to assess differences between healthy and non-healthy populations. The prediction error is the difference between the predicted and chronological age, as shown in figure \ref{fig:prediction_error}. A higher prediction error is assumed to indicate  an accelerated aging process \citep{Franke2010, Gaser2013}.    

The computation of the GPR uncertainty $\cov$ was presented in section \ref{sec:gpr}.  It can be thought of  as a metric on how close a testing point is to all the training points in the feature space, illustrated in figure \ref{fig:prediction_uncertainty}. The scatter plot represents a set of subjects in a 2-dimensional space composed of the volume of two different structures. By training a GPR on a set of training points (represented by circles), we  obtain a measure of uncertainty $\cov$ for every point in the 2D-space. This covariance matrix is represented by the shading of the grid, where darker regions correspond to regions where the predictor has higher confidence on its prediction. When performing prediction on previously unseen points (Test Subject 1 and Test Subject 2), we can obtain both a predicted age $\hat{y}$ and its confidence $\cov$. In figure \ref{fig:prediction_uncertainty}, we observe that even though both test subjects get similar predicted values, the confidence of the prediction for subject two is higher due to its proximity to the training set.

The third metric, the age-weighted uncertainty $\covw$ expands upon the notion of uncertainty by taking into account the subject's age. Measuring $\covw$ is equivalent to adding a further dimension to the distance measured by the normal GPR uncertainty. The reasoning behind this is to give higher similarity to individuals which have similar morphological features to healthy individuals of similar ages. For example, we see in figure \ref{fig:weighted_uncertainty} that a healthy testing point (blue) is close to training points with similar features and age; the testing point would therefore have a high $\covw$ value. On the other hand, the Testing Point corresponding to a diseased subject (red point) would have a low $\covw$ value because even though there are individuals in the training set with similar feature values, they correspond to subjects with a different age range. Both proposed metrics are closely related. In fact, $\cov$ is equivalent to $\covw$ for the special case when $l_y=\infty$.

\subsection{Aging and Disease Assessment } \label{sec:aging}
As discussed in the introduction, several studies have demonstrated that aging is a complex process, which affects different brain structures and regions at different rates of change. It has also been reported that deleterious changes caused by neurodegenerative disease follow a pattern that resembles an accelerated aging process. In order to evaluate how aging and neurodegenerative disease affect different brain regions, we performed an analysis of the volumetric features obtained from our training set. To facilitate this analysis we restricted our analysis to images obtained from the ADNI database. To assess which individual structures are affected either by aging, disease or both factors, a series of simple linear fixed effects model were fitted to our data. In each one of these models, the dependent variable corresponds to the volume of a different brain structure, and the independent variables correspond to age, sex, diagnosis (0 = healthy, 1 = MCI/AD) and an interaction term between diagnosis and age.

\section{Results} \label{sec:results}

\subsection{Assessing the Effect of Aging and Disease on Brain Development} \label{sec:progress}

\begin{figure*}[h!]
	\begin{subfigure}{.5\textwidth}
		\centering\includegraphics[width=\textwidth]{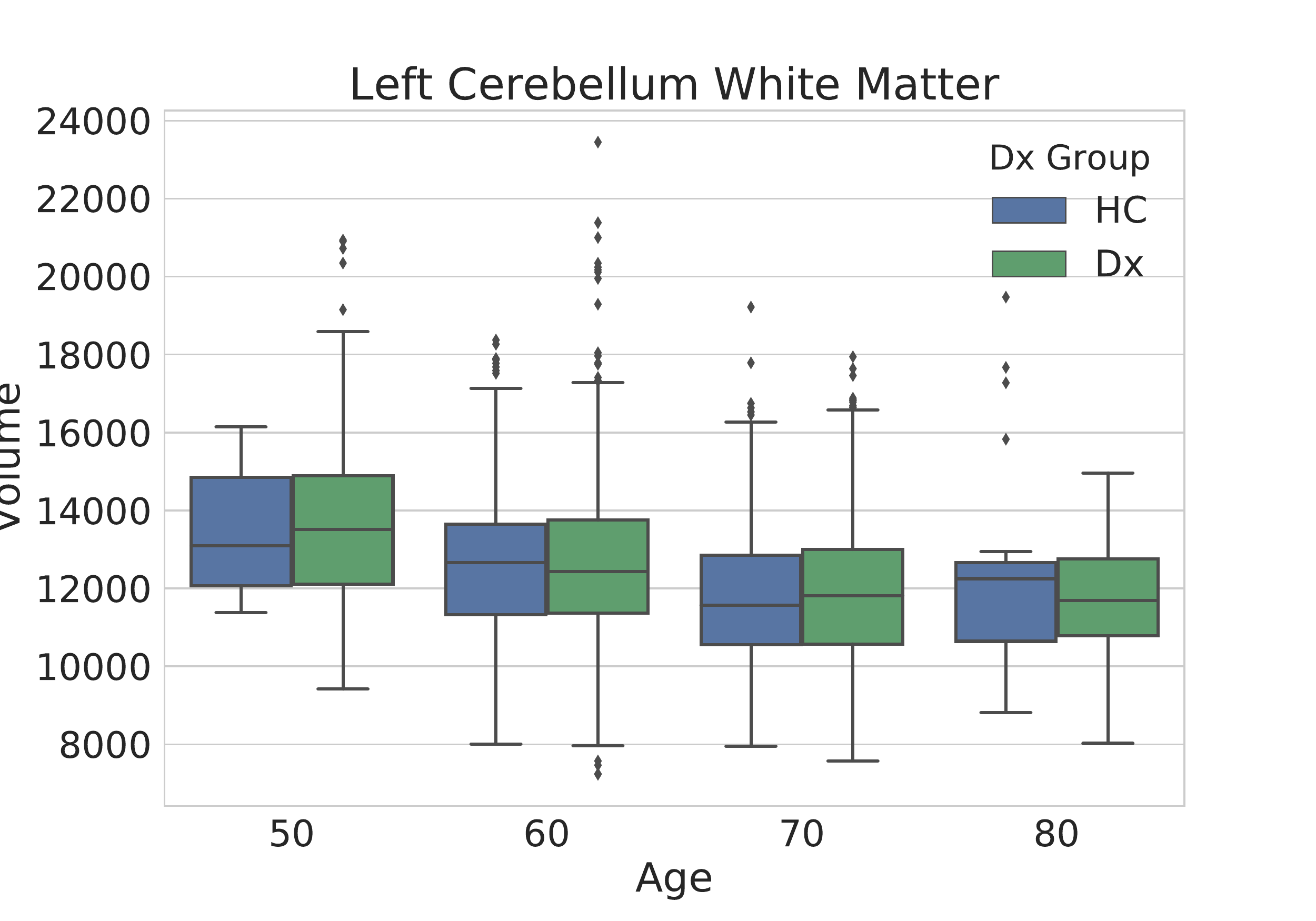}
	\end{subfigure}
	\begin{subfigure}{.5\textwidth}
		\centering\includegraphics[width=\textwidth]{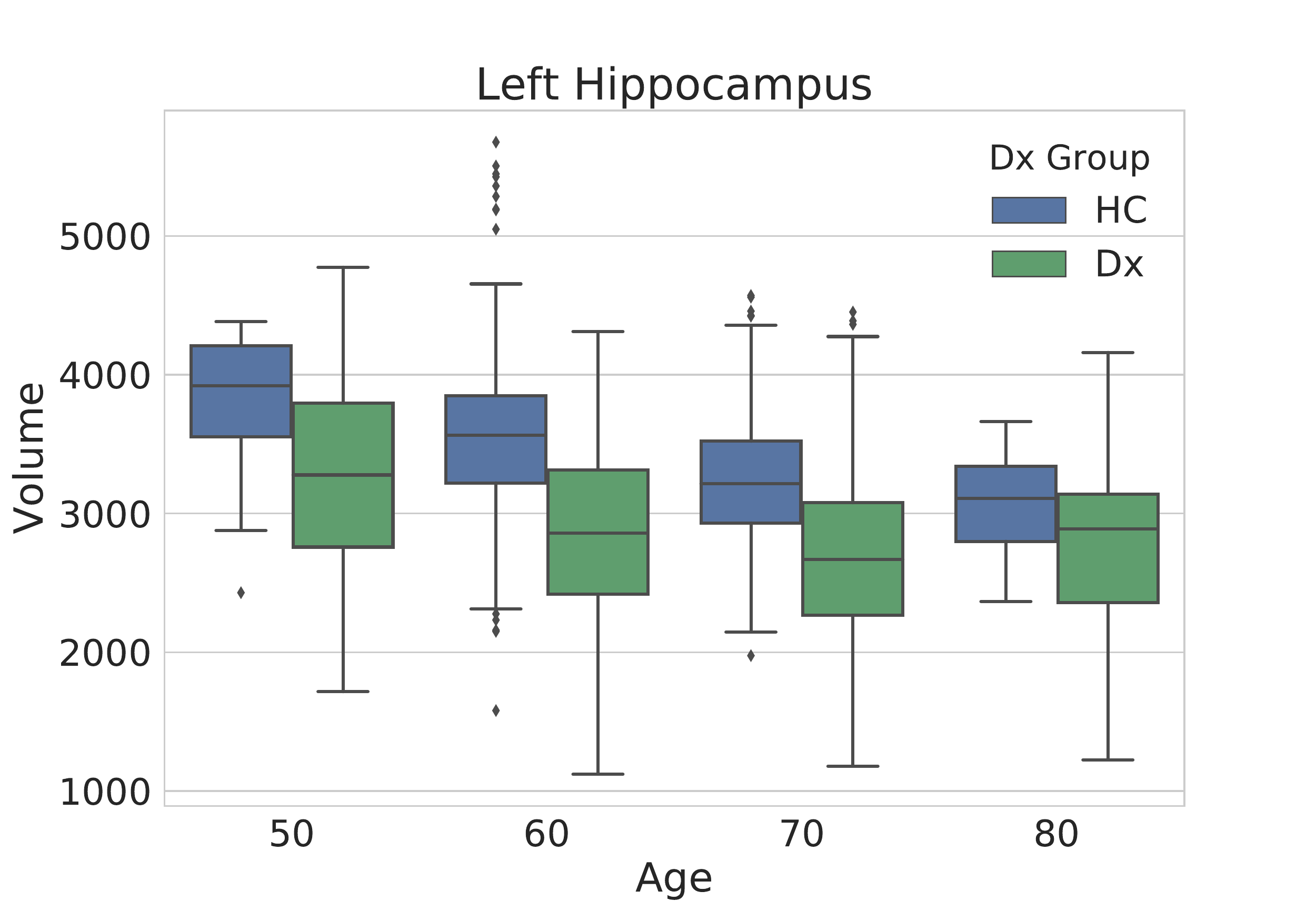}
	\end{subfigure}
	\begin{subfigure}{.3\textwidth}
		\centering\includegraphics[width=\textwidth]{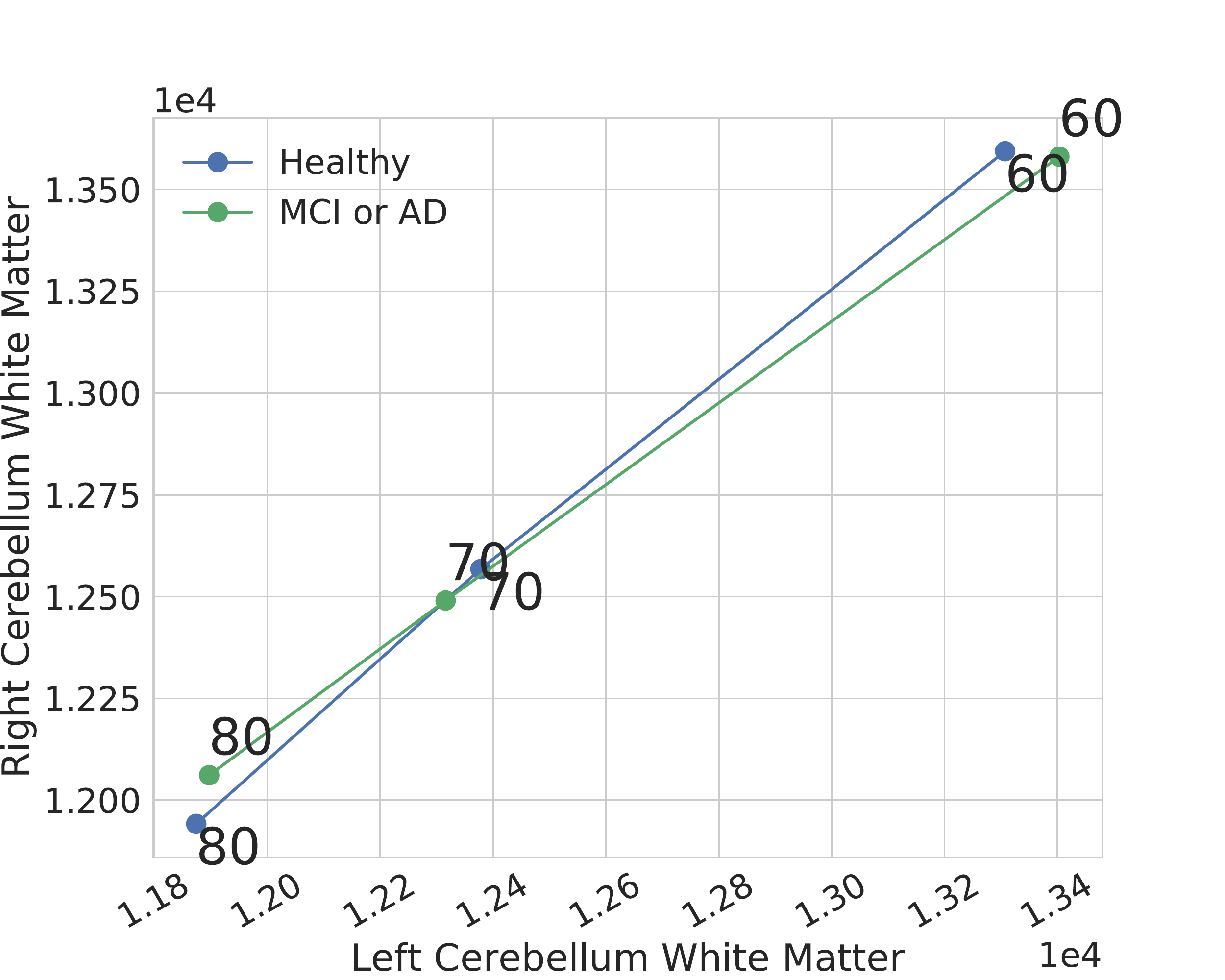}
	\end{subfigure}
	\begin{subfigure}{.3\textwidth}
		\centering\includegraphics[width=\textwidth]{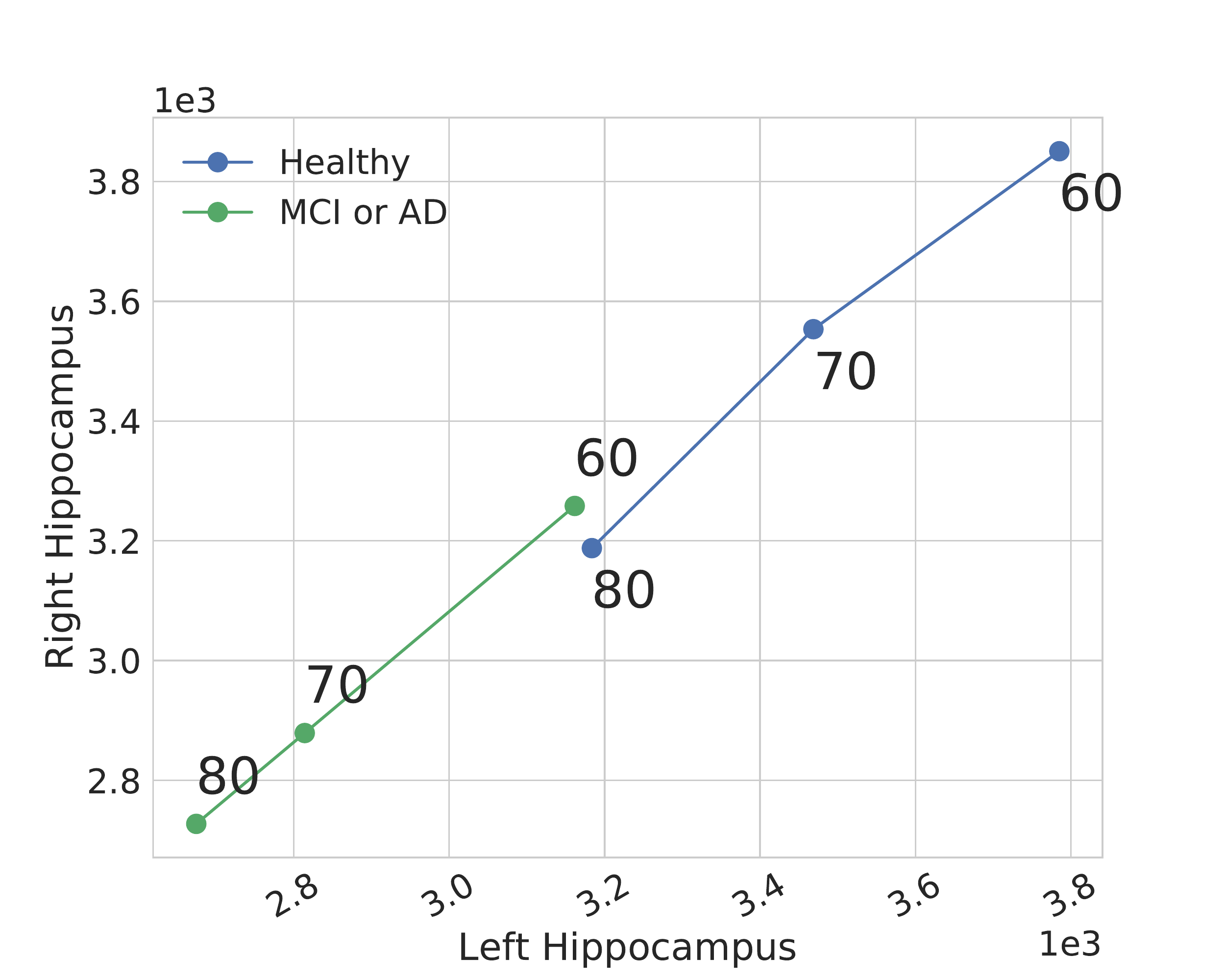}
	\end{subfigure}
	\begin{subfigure}{.3\textwidth}
		\centering\includegraphics[width=\textwidth]{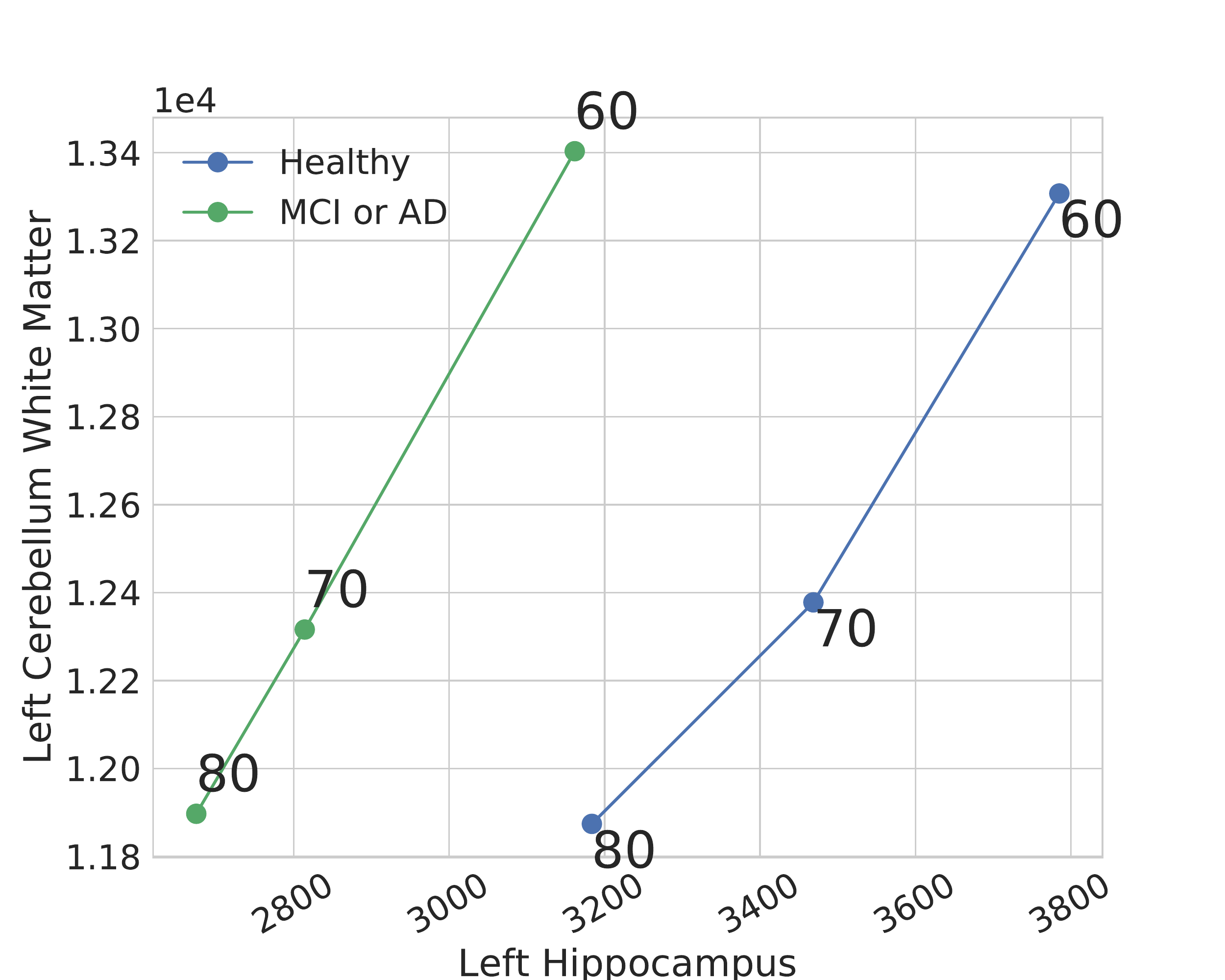}
	\end{subfigure}
	
	\caption{Top: Box plots showing changes in left cerebellum white matter volume and hippocampus volume for individuals between 50 and 80 years old. Bottom: Plots showing age progression of feature pairs for three different cases;  left: structures that are not affected by disease; center: structures affected by disease that show an accelerated aging pattern; right: one structure affected by disease (left hippocampus) and one structure with no significant disease effect (left cerebellum white matter). }
\label{fig:progress}
\end{figure*}

In this section we present the results obtained after fitting the aging and disease model described in section \ref{sec:aging}. In table \ref{tab:fit}, we present the regression coefficients for age, diagnosis and the age/diagnosis interaction term as well as their corresponding p-values for the linear fixed effects model. The table is sorted by descending p-value for the diagnostic coefficient, which means that the structures at the top of the table are those which present more significant volume alterations caused by disease. Age and volume variables were normalized in order to make the coefficients of different structures comparable. In the case of bilateral structures we only show the values for the left hemisphere in order to simplify our analysis. Plots showing the progress of the hippocampus and cerebellum white matter across different ages for both healthy and individuals with MCI/AD are also presented to illustrate our results in figure \ref{fig:progress}. The  hippocampus was selected since it was the structure which had the most evident effects of age and disease. On the other hand, the cerebellum white matter was selected as a structure which showed significant effects of aging but is apparently not largely affected by Alzheimer's disease.

There are a couple of relevant observations that can be extracted from table \ref{tab:fit}. First, the regression coefficients for age and diagnosis always have the same sign for structures with significant associations, and there exist significant interactions between aging and diagnosis for most of the analyzed structures. This supports the hypothesis that disease and aging are overlapping processes that affect the brain structures in the same direction. However, we can also observe in table \ref{tab:fit} that there exist some structures that although largely affected by aging do not present significant disease effects (\emph{i.e.} cerebellum white matter).  This can also be observed on the box plots in the top of figure \ref{fig:progress}, where a clear difference between the HC and Dx groups is evident for all age ranges in the case of the hippocampus, whereas for the cerebellum white matter no significant differences exist between both groups.

To further illustrate the point that aging and disease are processes that affect different regions of the brain at different rates,  we show the progress of pairs of features for both the HC and Dx groups (bottom of figure \ref{fig:progress}). By looking at the central plot, where the volume of left and right hippocampus is shown, we can understand the reasoning behind the accelerated aging hypothesis of previous age estimation works. Indeed, by looking only at these features, we would be tempted to conclude that the brain of a healthy 80 year old is essentially similar to that of a diseased 60 year old. However, this observation contrasts with the left plot, where the left and right cerebellum white matter volumes are shown. By looking at these features alone, we would draw a different conclusion, since it would appear that there are no differences between the brains of healthy and diseased subjects of the same age. By looking at the left hippocampus and left cerebellum white matter simultaneously (right in figure \ref{fig:progress}), we can observe that disease produce changes in the brain that are essentially different to those of accelerated aging, causing the overall appearance of the brain of an average 60 year old diagnosed with AD to be different to a healthy individual of any age.  These observations support our hypothesis that morphological changes associated to AD and MCI are complex and that a model of accelerated aging across the whole brain may be too simplistic to model the specific effects of disease and aging at specific brain structures. 

\begin{sidewaystable}[]
	\centering
	\caption{Coefficients and p-values corresponding to the linear models fitted to predict volume of individual structures. Structures are sorted by descending p-value for diagnostic. }
	\label{my-label}
	\begin{tabular}{lllllll}
		Structure& Age Coefficient & Dx Coefficient & Age $\times$ Dx Coefficient & Age p-value & Dx p-value & Age $\times$ Dx p-value \\
		\hline
		\rowcolor{lightgray} 
		Left.Hippocampus             & -0.30           & -0.53          & 0.03               &\num{ 9.12E-107 }  &\num{4.04E-291} & \num{2.87E-02}       \\
		Left.Amygdala                & -0.21           & -0.41          & 0.07               &\num{ 4.13E-50  }  &\num{1.53E-169} & \num{1.67E-05}       \\
				\rowcolor{lightgray} 
		Left.Inf.Lat.Vent            & 0.27            & 0.38           & -0.07              &\num{ 1.77E-70  }  &\num{1.50E-141} & \num{3.00E-05}       \\
		Left.Lateral.Ventricle       & 0.22            & 0.24           & -0.09              &\num{ 8.38E-46  }  &\num{1.84E-57 } & \num{1.32E-07}       \\
		\rowcolor{lightgray} 
		CSF                          & 0.16            & 0.22           & -0.11              &\num{ 1.07E-23  }  &\num{1.85E-46 } & \num{5.95E-11}       \\
		Left.Accumbens.area          & -0.30           & -0.22          & 0.07               &\num{ 2.75E-75  }  &\num{1.48E-43 } & \num{7.89E-05}       \\
		\rowcolor{lightgray} 
		3rd.Ventricle               & 0.27            & 0.19           & -0.13              &\num{ 1.04E-70  }  &\num{1.53E-36 } & \num{3.49E-15}       \\
		Left.choroid.plexus          & 0.14            & 0.17           & -0.04              &\num{ 2.04E-18  }  &\num{1.59E-27 } & \num{2.16E-02}       \\
		\rowcolor{lightgray} 
		Left.Putamen                 & -0.15           & -0.14          & 0.02               &\num{ 9.84E-21  }  &\num{2.15E-17 } & \num{0.30}       \\
		Left.VentralDC               & -0.27           & -0.11          & -0.03              &\num{ 7.09E-71  }  &\num{3.38E-14 } & \num{3.00E-02}       \\
		\rowcolor{lightgray} 
		Left.Thalamus.Proper         & -0.32           & -0.11          & -0.01              &\num{ 6.28E-97  }  &\num{2.65E-13 } & \num{0.48}       \\
		Left.Cerebellum.Cortex       & -0.26           & -0.07          & -0.03              &\num{ 1.98E-65  }  &\num{5.01E-06 } & \num{9.02E-02}       \\
		\rowcolor{lightgray} 
		Brain.Stem                   & -0.23           & -0.06          & -0.02              &\num{ 1.15E-48  }  &\num{1.60E-05 } & \num{0.13}       \\
		Left.Caudate                 & 0.07            & 0.04           & 0.03               &\num{ 7.89E-06  }  &\num{1.93E-02 } & \num{0.12}       \\
		\rowcolor{lightgray} 
		Left.Cerebellum.White.Matter & -0.30           & -0.03          & 0.00               &\num{ 1.60E-74  }  &\num{0.10 } & \num{0.86}       \\
		Left.Pallidum                & -0.06           & -0.02          & -0.04              &\num{ 4.84E-04  }  &\num{0.23 } & \num{1.31E-02}       \\
		\rowcolor{lightgray} 
		4th.Ventricle               & 0.09            & 0.00           & -0.08              &\num{ 2.38E-07  }  &\num{0.83 } & \num{1.13E-05}       \\
		Left.vessel                  & 0.12            & 0.00           & -0.02              &\num{ 1.05E-12  }  &\num{0.93} & \num{0.37}      
	\end{tabular}
\label{tab:fit}
\end{sidewaystable}

\subsection{Training of the Age Prediction Model}

\begin{figure*}[h!]
	
	\begin{subfigure}{.5\textwidth}
		\centering\includegraphics[width=\textwidth]{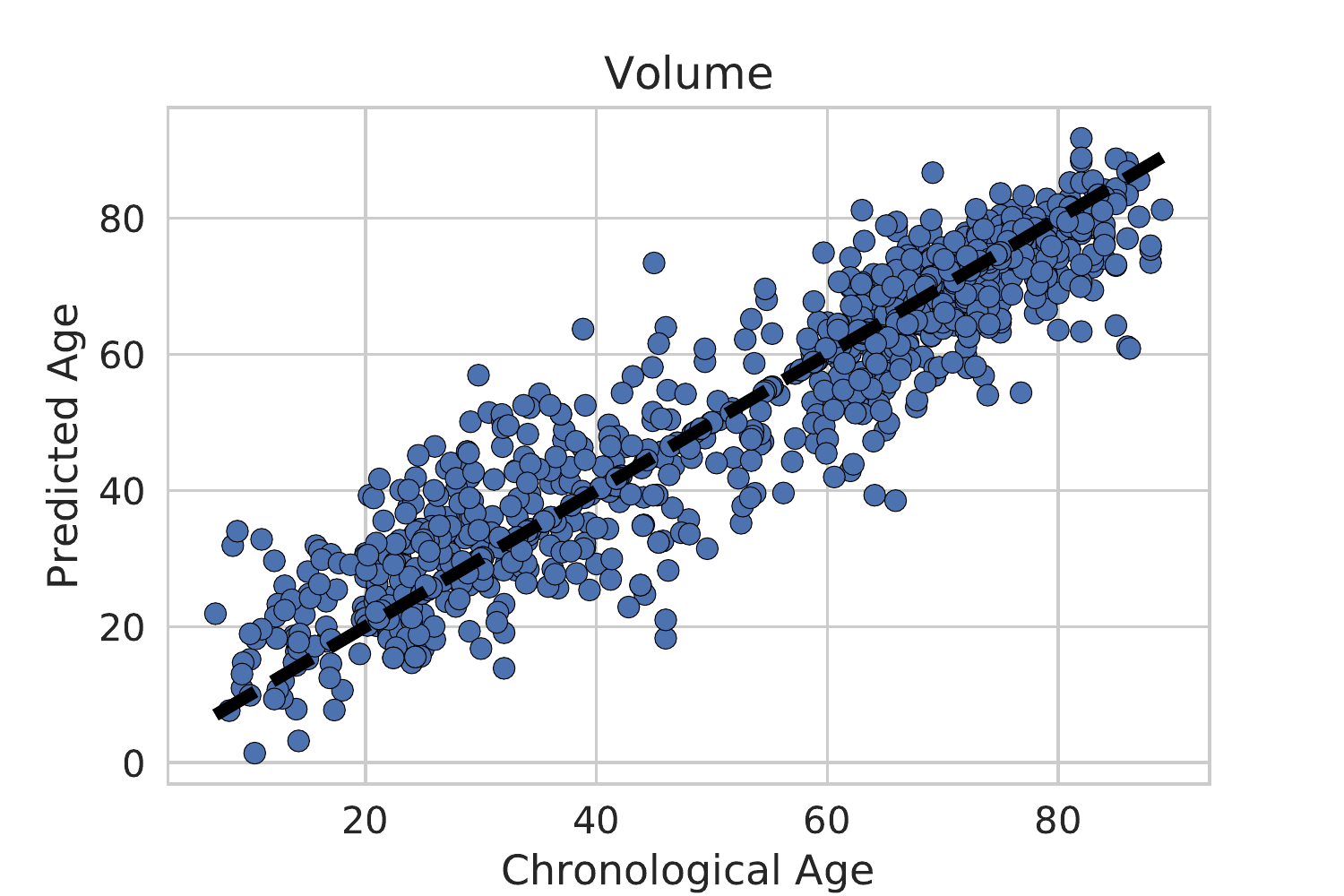}
	\end{subfigure}
	\begin{subfigure}{.5\textwidth}
		\centering\includegraphics[width=\textwidth]{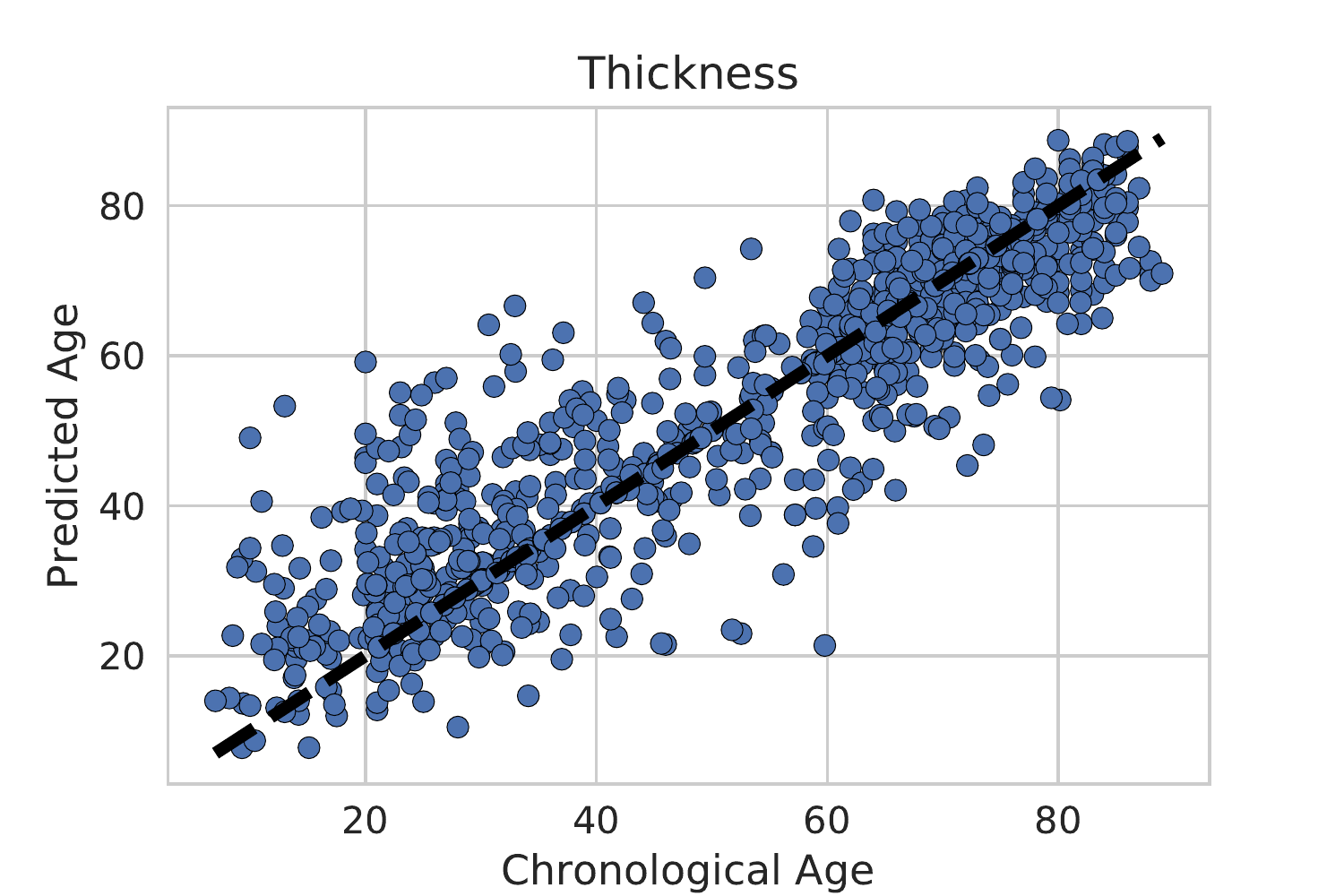}
	\end{subfigure}
	\begin{subfigure}{.5\textwidth}
		\centering\includegraphics[width=\textwidth]{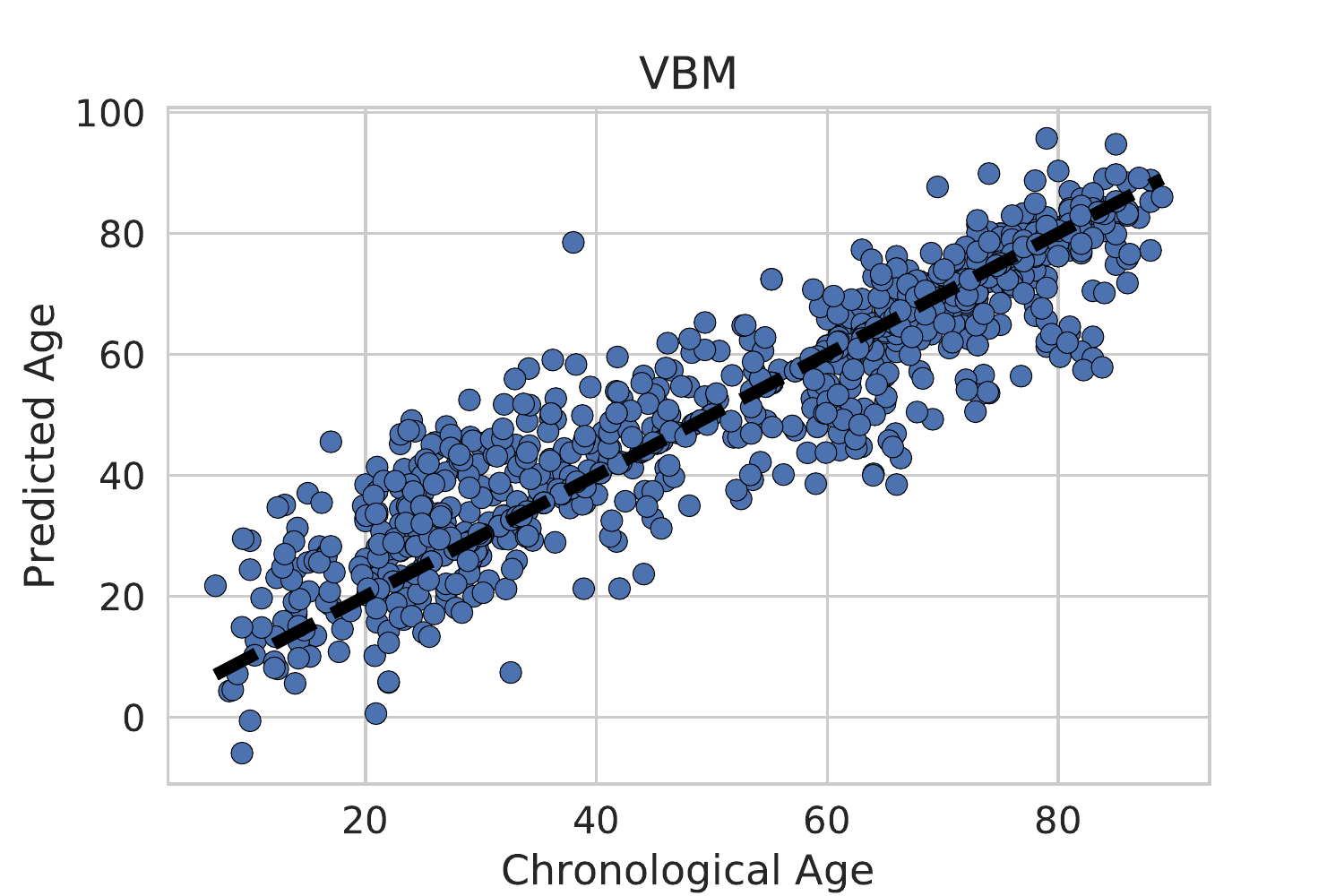}
	\end{subfigure}
	\begin{subfigure}{.5\textwidth}
		\centering\includegraphics[width=\textwidth]{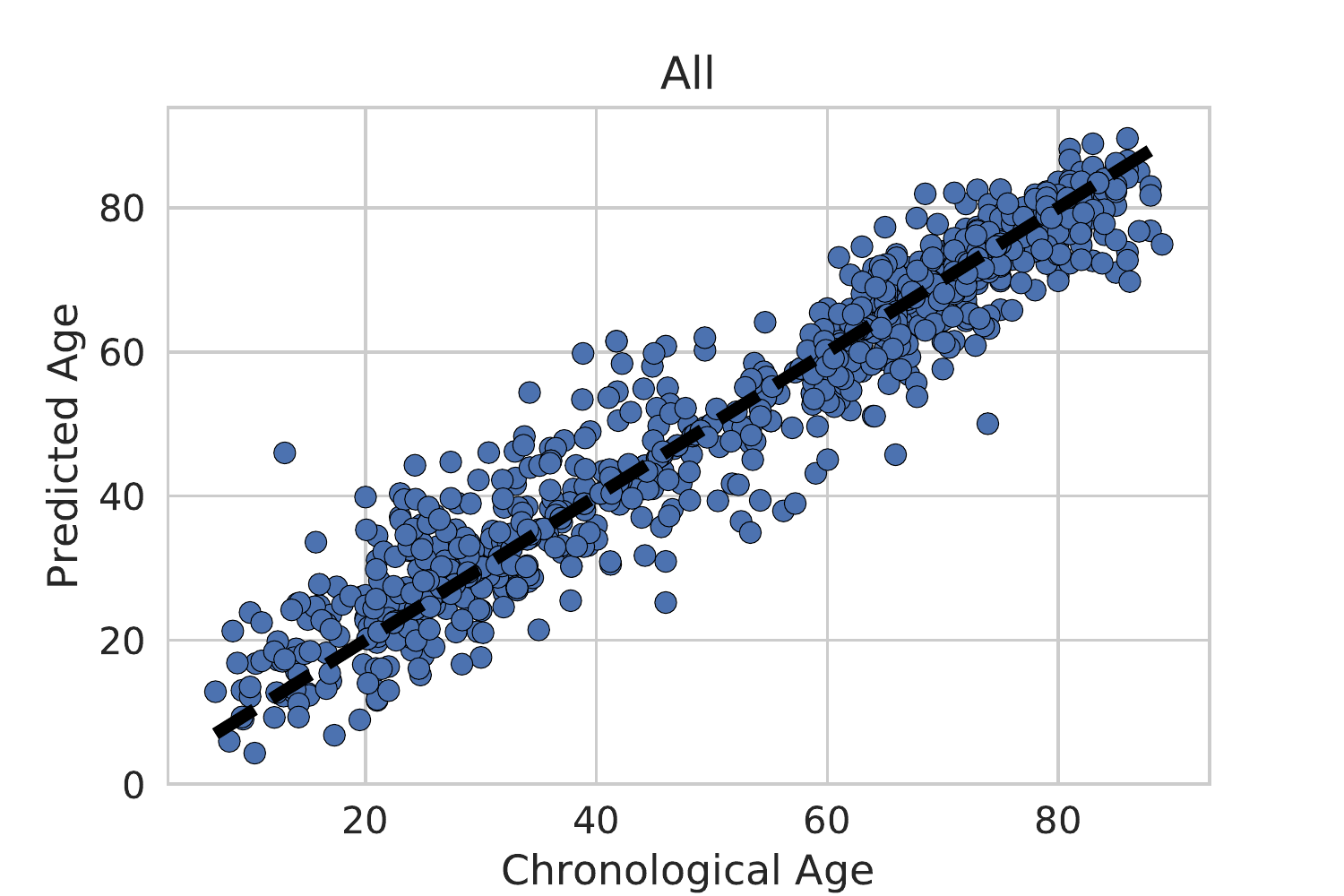}
	\end{subfigure}
	\caption{Scatter plots showing the prediction results of the age prediction models trained using different feature sets. }
	\label{fig:scatter}
\end{figure*}

Using the training datasets summarized in table \ref{table:datasets-training}, we train 4 different GPR models, each one with a different set of features as described in section \ref{sec:features}. Each one of the GPR models is trained to estimate the age of healthy subjects based on either volume, thickness, VBM features or a combination of all features. Our models were implemented using python together with the scikit-learn toolbox. In table \ref{table:results-training}, we show the Mean Absolute Error (MAE) and $R^2$ score for the training set, using a 5-fold cross validation. Our model presents similar MAE and  $R^2$ when compared to previous work on age estimation \citep{Valizadeh2017, Cole2016}. Similar to previously reported results \citep{Valizadeh2017}, we observed higher $R^2$ score and lower MAE for the model trained using an ensemble of all available features. The chronological and predicted age for each subject in the training set are presented in the scatter plots in figure \ref{fig:scatter}.

\begin{table*}[ht]
	\centering
	\begin{tabular}{ l c  c    }
		Feature Set    & MAE  & $R^2$    \\ 
		\rowcolor{lightgray} 
		Volume  & 5.52 & 0.87      \\ 
		Thickness & 6.50 & 0.80       \\
		\rowcolor{lightgray}
		VBM & 5.65 & 0.86         \\
		All & 3.86 & 0.93        \\
	\end{tabular}
	\caption{Mean Absolute Error (MAE) and $R^2$ score of the age prediction models trained with different sets of features. Measurements are obtained using a 5-fold cross validation on the training set.}
	\label{table:results-training}
\end{table*}

\subsection{Evaluation of Gassian Process Uncertainty as a Measure of Brain Abnormality } 

\begin{figure*}[h!]
	\begin{subfigure}{.33\textwidth}
		\centering\includegraphics[width=\textwidth]{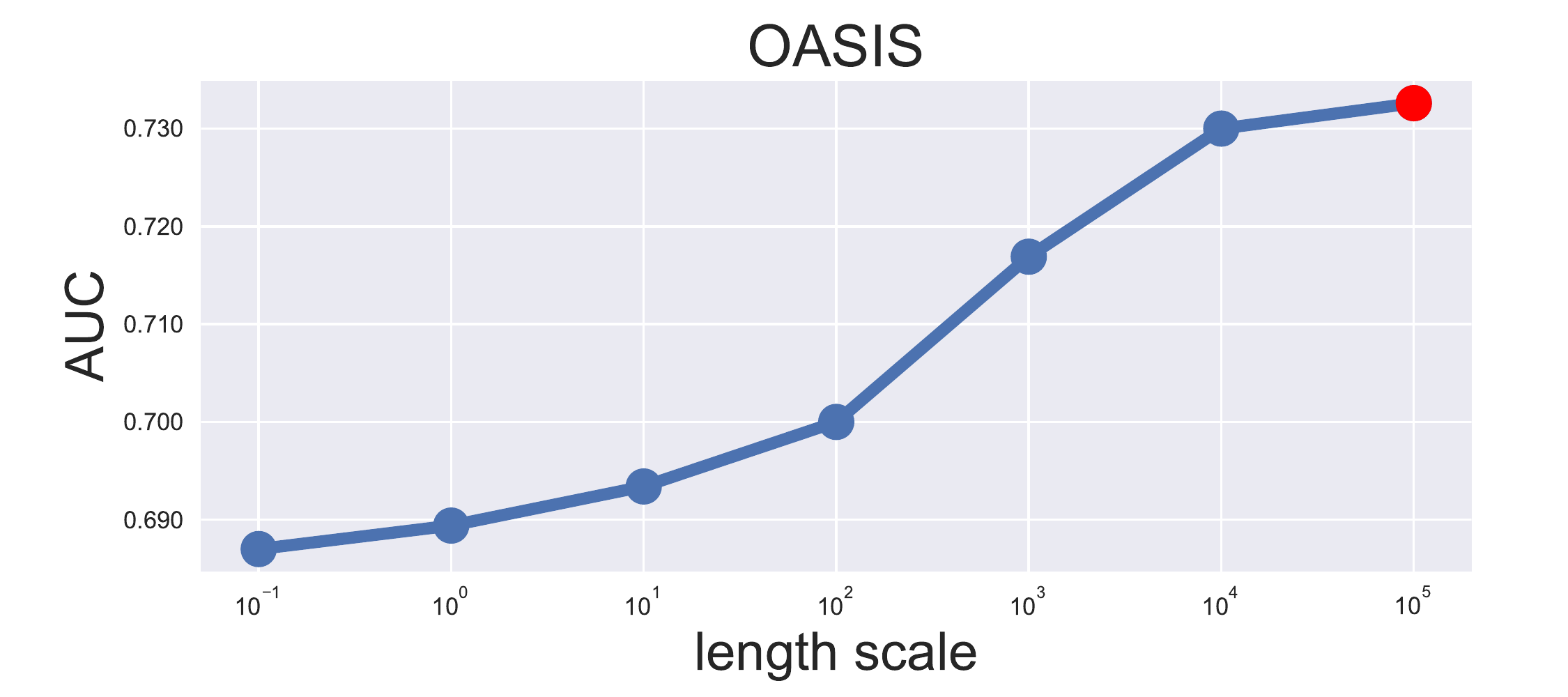}
	\end{subfigure}
	\begin{subfigure}{.33\textwidth}
		\centering\includegraphics[width=\textwidth]{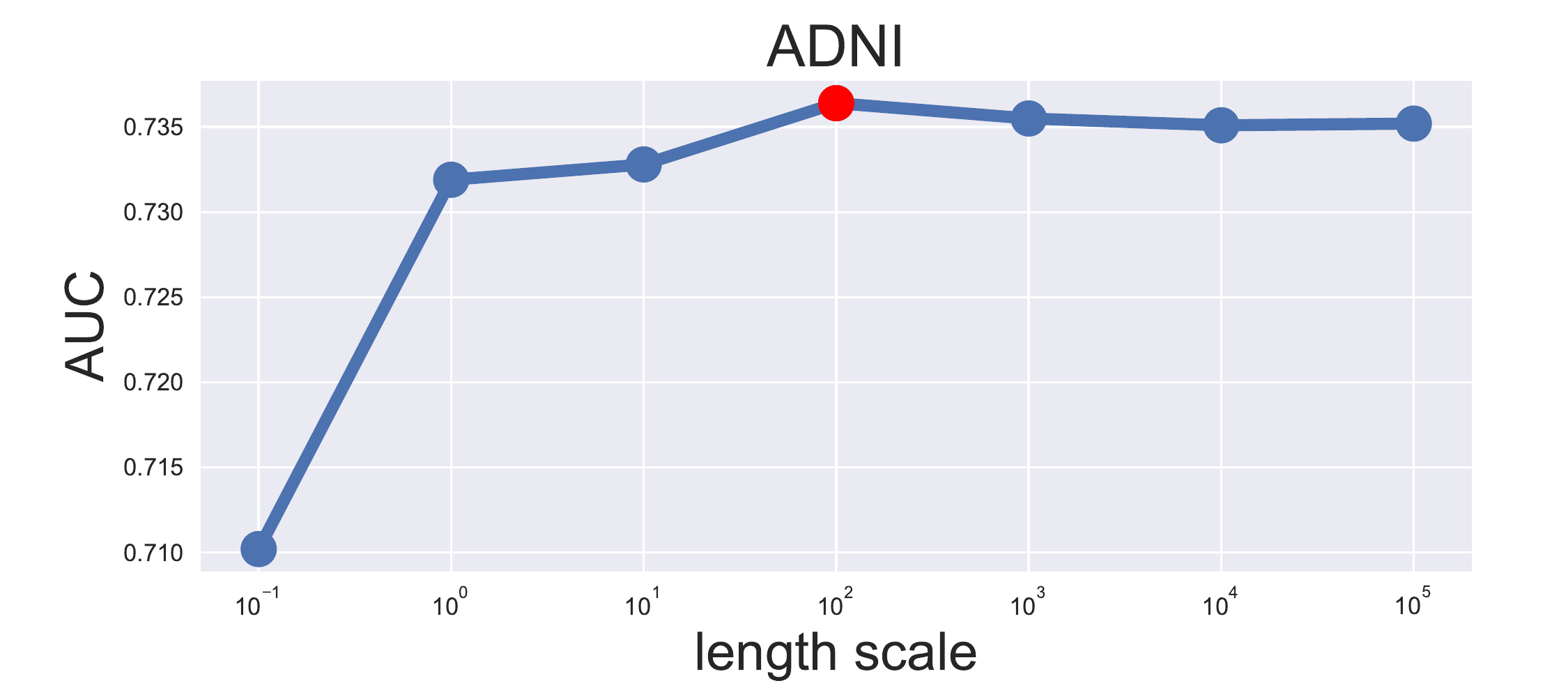}
	\end{subfigure}
	\begin{subfigure}{.33\textwidth}
		\centering\includegraphics[width=\textwidth]{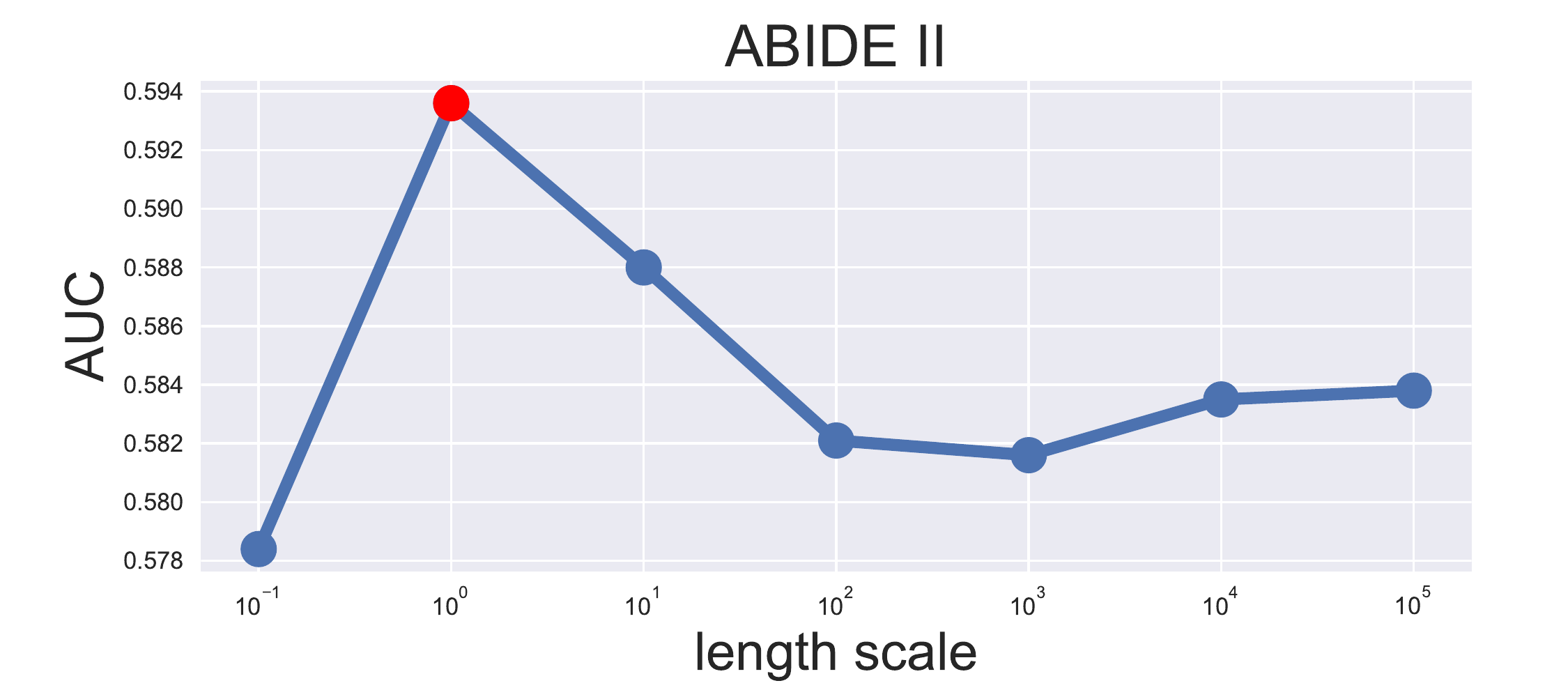}
	\end{subfigure}
	\caption{AUC values obtained by using the $\covw$ metric for different age length scales $l_y$. According to this curves, a different length scale was selected for each experiment (ADNI: $l_y=$\num{1e2}, OASIS: $l_y=$\num{1E5}, ABIDE: $l_y=$1). The selected length scales are highlighted in red in each plot. }
	\label{fig:length_scale}
\end{figure*}

In this section, we present results of our experiments comparing the use of prediction error $\epsilon$, uncertainty $\cov$ and age-weighted uncertainty $\covw$ as a biomarker for differentiating between healthy controls and patients with MCI, AD or autism. We performed three different experiments: the first two experiments are targeted at finding differences between HC, MCI and AD groups in both the ADNI and OASIS databases; the third experiment is performed on the ABIDE II database where differences between autism and healthy groups are evaluated. Note that as summarized in tables \ref{table:datasets-training} and \ref{table:datasets-testing}, the datasets used for testing are different to those used for training.  For all experiments we assessed differences between groups both by performing  non-parametric Wilcoxon rank-sum tests \citep{mann1947} and by measuring the classification performance in a per subject basis by generating Receiver Operating Characteristic (ROC) curves with their corresponding Area Under the Curve (AUC) values.  For all experiments, results are shown for four different sets of features: volume, thickness and VBM, as well as for the combination of all three feature sets. Our proposed GPR uncertainty based metrics are compared to the prediction error $\epsilon$, obtained in a similar fashion as previous work on age estimation \citep{Franke2010}. An appropriate age length scale parameter $l_y$ for the $\covw$ metric was set independently for each experiment by performing evaluations at different scales an keeping the best performing results (See figure \ref{fig:length_scale}).

\begin{table}[ht]
	\centering
	\resizebox{\columnwidth}{!}
	{
		\begin{tabular}{lccc|ccc}
			\multicolumn{1}{c}{} &\multicolumn{3}{c}{HC-MCI} & \multicolumn{3}{c}{MCI-AD}  \tabularnewline
			
			& $\epsilon$ & $\cov$ & $\covw$ & $\epsilon$ & $\cov$ & $\covw$  \tabularnewline
			\hline
			\rowcolor{lightgray} 
			Volume  & \num{6.82E-19} & \num{2.07E-29} &\num{1.85E-31 } &\num{3.36E-60 } &\num{ 4.54E-69 }& \num{2.05E-74}  \tabularnewline
			Thickness& \num{2.18E-18} &\num{3.53E-27} &\num{2.92E-24} &\num{3.98E-38} &\num{6.03E-105} &\num{1.45E-101} \tabularnewline
			\rowcolor{lightgray} 
			VBM  & \num{3.10E-07} &\num{6.12E-30} &\num{1.20E-34} &\num{5.61E-22} &\num{ 2.92E-77} &\num{1.17E-76}  \tabularnewline
			All  & \num{5.63E-05} &\num{5.16E-36} &\num{5.10E-37} &\num{9.46E-23} &\num{ 2.76E-103} &\num{1.02E-103} \tabularnewline			
		\end{tabular} 
	} \caption{p-values corresponding  to the statistical tests performed on the experiments comparing the HC, MCI and AD groups on the ADNI dataset.}
	\label{table:pvalues-adni}
\end{table}
\begin{table}[ht]
	\centering
	\resizebox{0.7\columnwidth}{!}
	{
		\begin{tabular}{lccc|ccc}
			\multicolumn{1}{c}{} &\multicolumn{3}{c}{HC-MCI} & \multicolumn{3}{c}{MCI-AD}  \tabularnewline
			& $\epsilon$ & $\cov$ & $\covw$ & $\epsilon$ & $\cov$ & $\covw$  \tabularnewline
			\hline
			\rowcolor{lightgray} 
			Volume  & 0.67 & 0.71 & \textbf{0.72} & 0.73 &  0.76 & \textbf{0.77}\tabularnewline
			Thickness  & 0.66 & \textbf{0.71} & \textbf{0.71} & 0.69 &  \textbf{0.80} &\textbf{0.80} \tabularnewline
			\rowcolor{lightgray} 
			VBM  & 0.60 & 0.71 & \textbf{0.72} & 0.64 &  \textbf{0.77} &\textbf{0.77} \tabularnewline
			All  & 0.59 & \textbf{0.74} & \textbf{0.74} & 0.64 &  \textbf{0.81} &\textbf{0.81} \tabularnewline
			
		\end{tabular} 
	} \caption{Area Under the Curve (AUC) values corresponding  to the statistical tests performed on the experiments comparing the HC, MCI and AD groups on the ADNI dataset.}
	\label{table:auc-adni}
\end{table}
\begin{figure*}[h!]
	\begin{subfigure}{.245\textwidth}
		\centering\includegraphics[width=\textwidth]{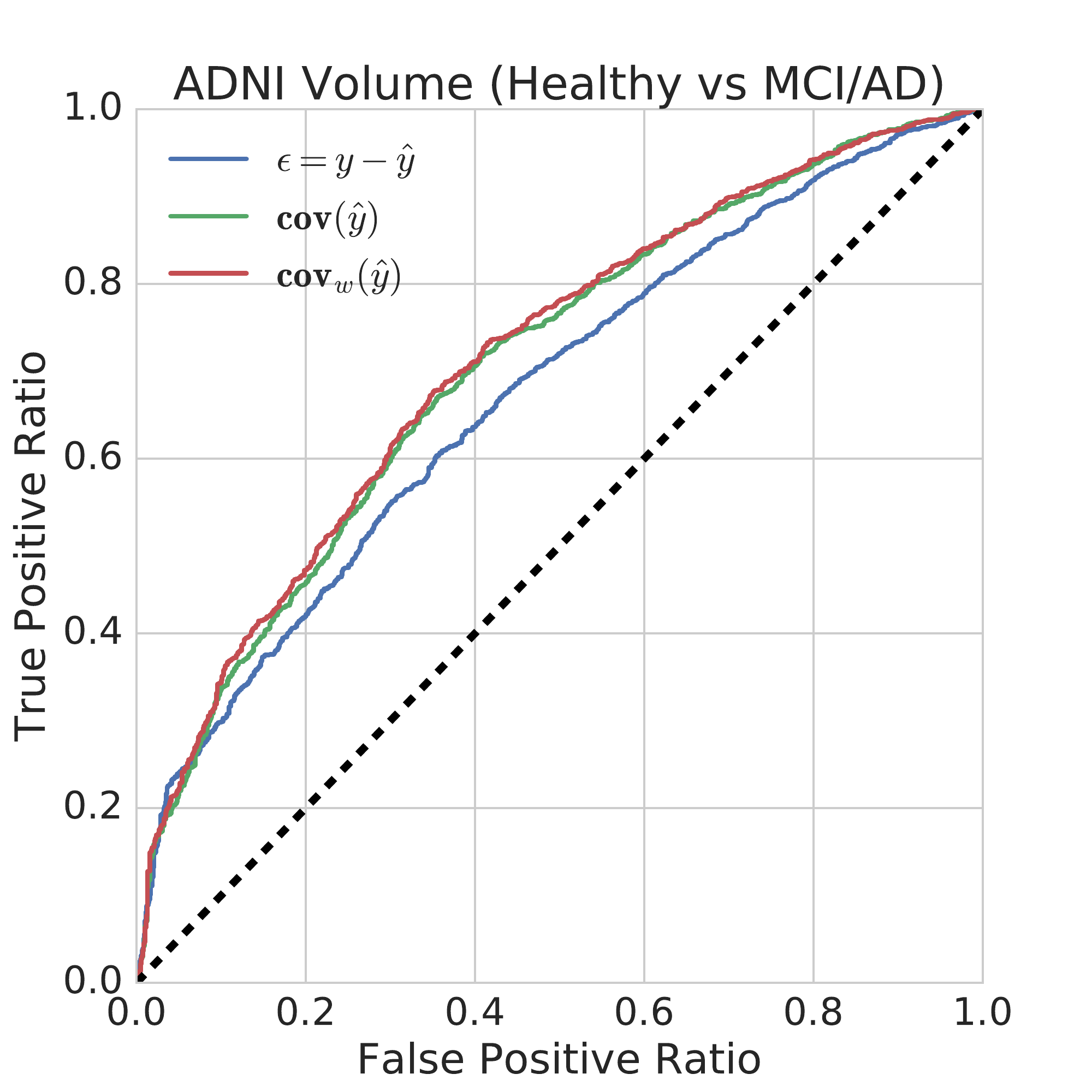}
	\end{subfigure}
	\begin{subfigure}{.245\textwidth}
		\centering\includegraphics[width=\textwidth]{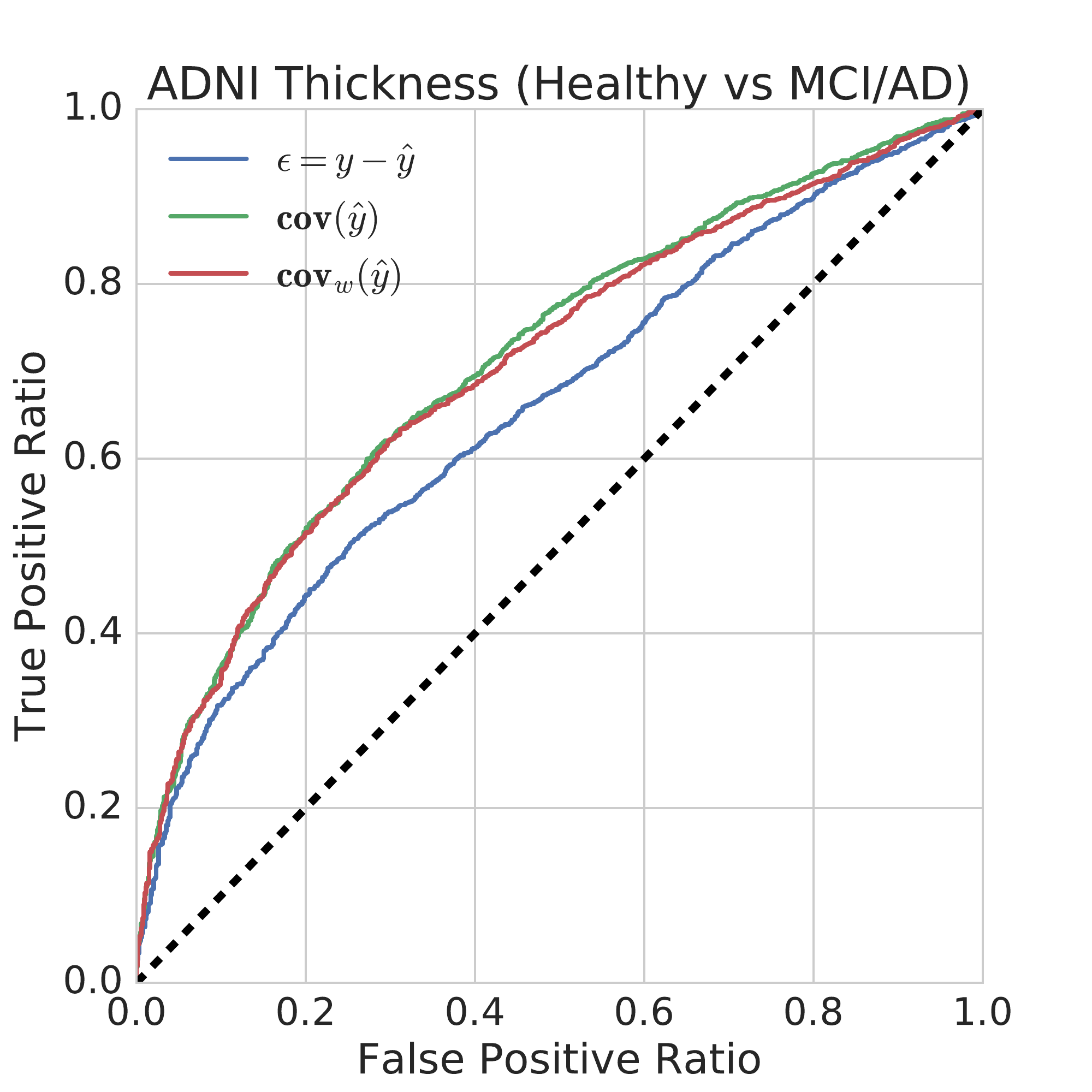}
	\end{subfigure}
	\begin{subfigure}{.245\textwidth}
		\centering\includegraphics[width=\textwidth]{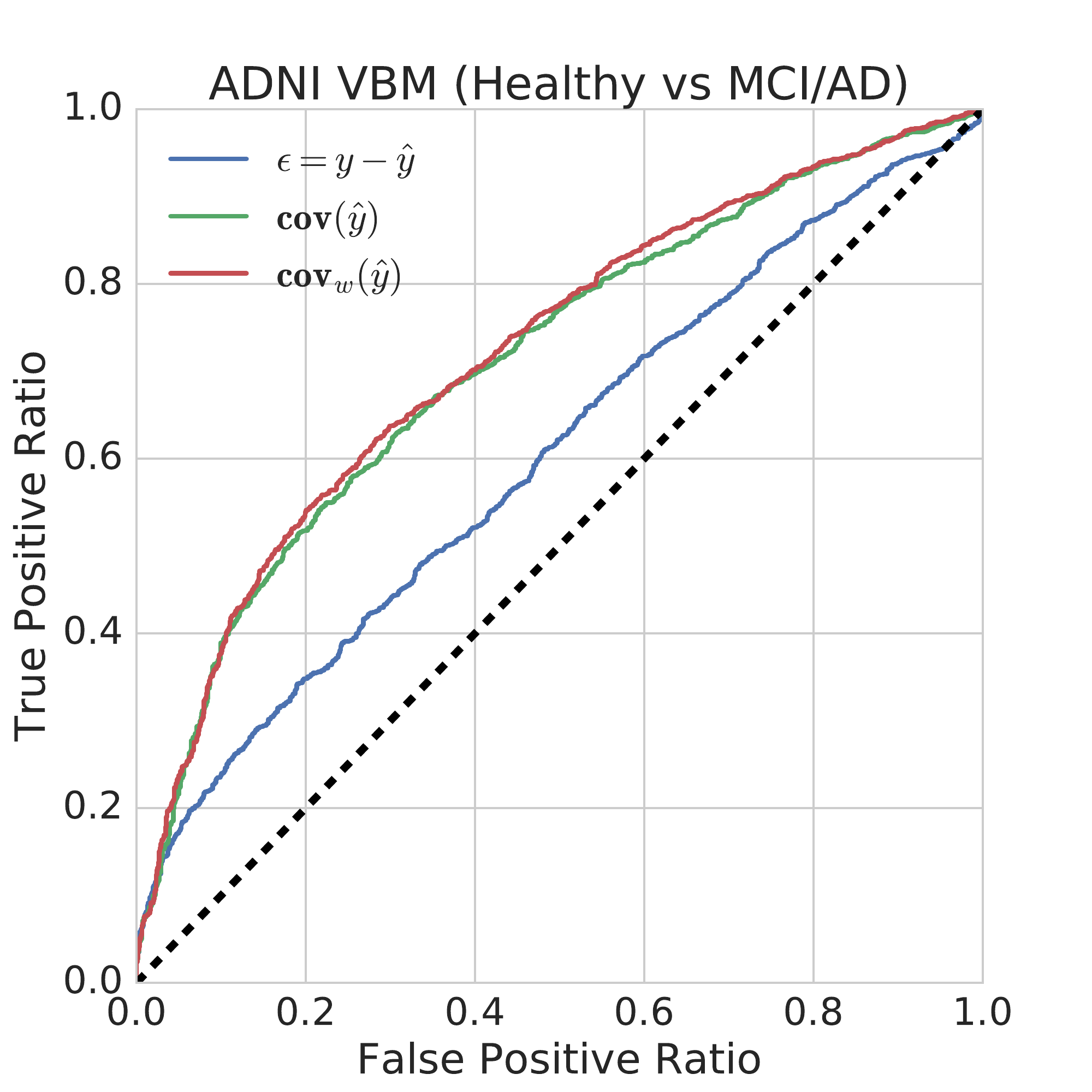}
	\end{subfigure}
	\begin{subfigure}{.245\textwidth}
		\centering\includegraphics[width=\textwidth]{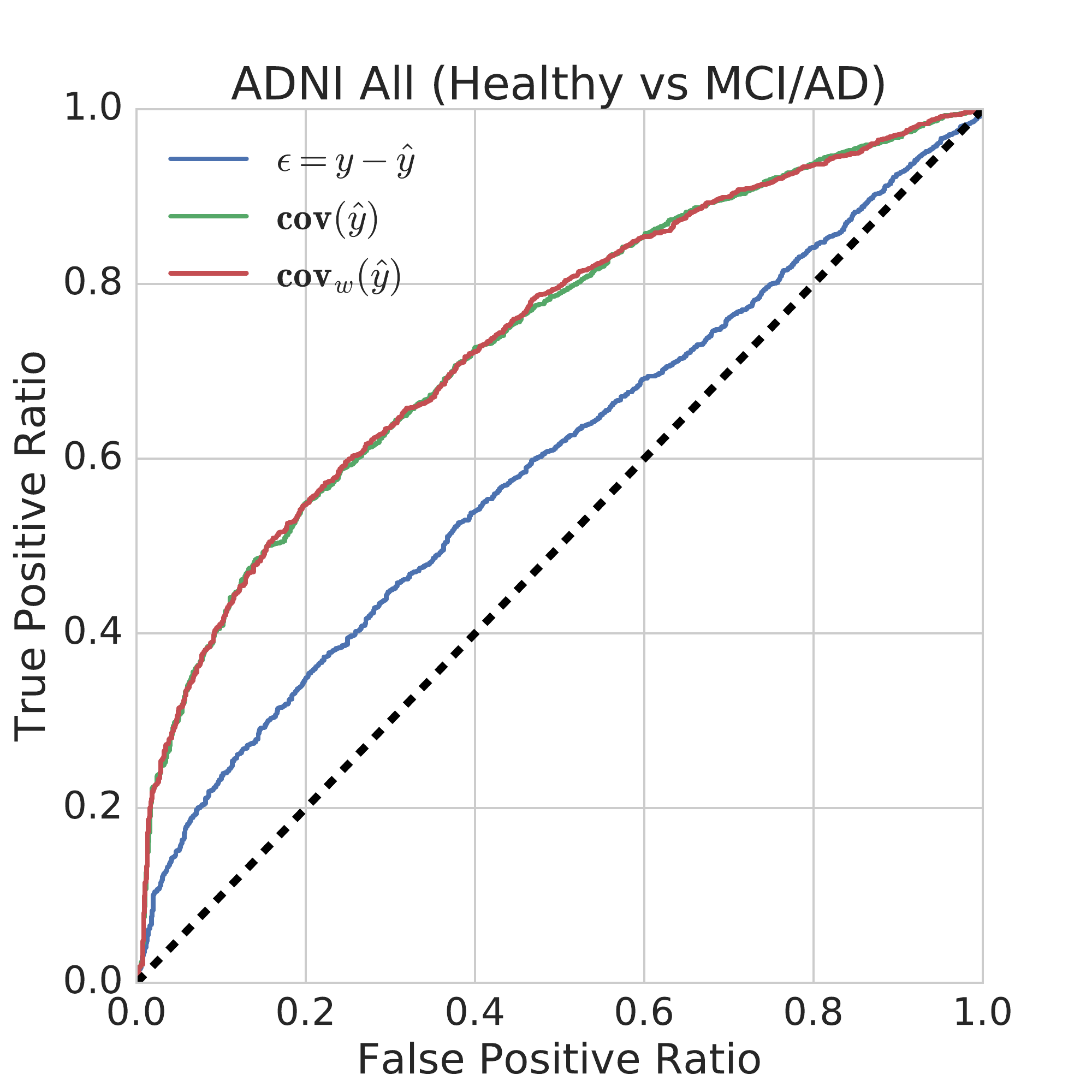}
	\end{subfigure}
	\begin{subfigure}{.245\textwidth}
		\centering\includegraphics[width=\textwidth]{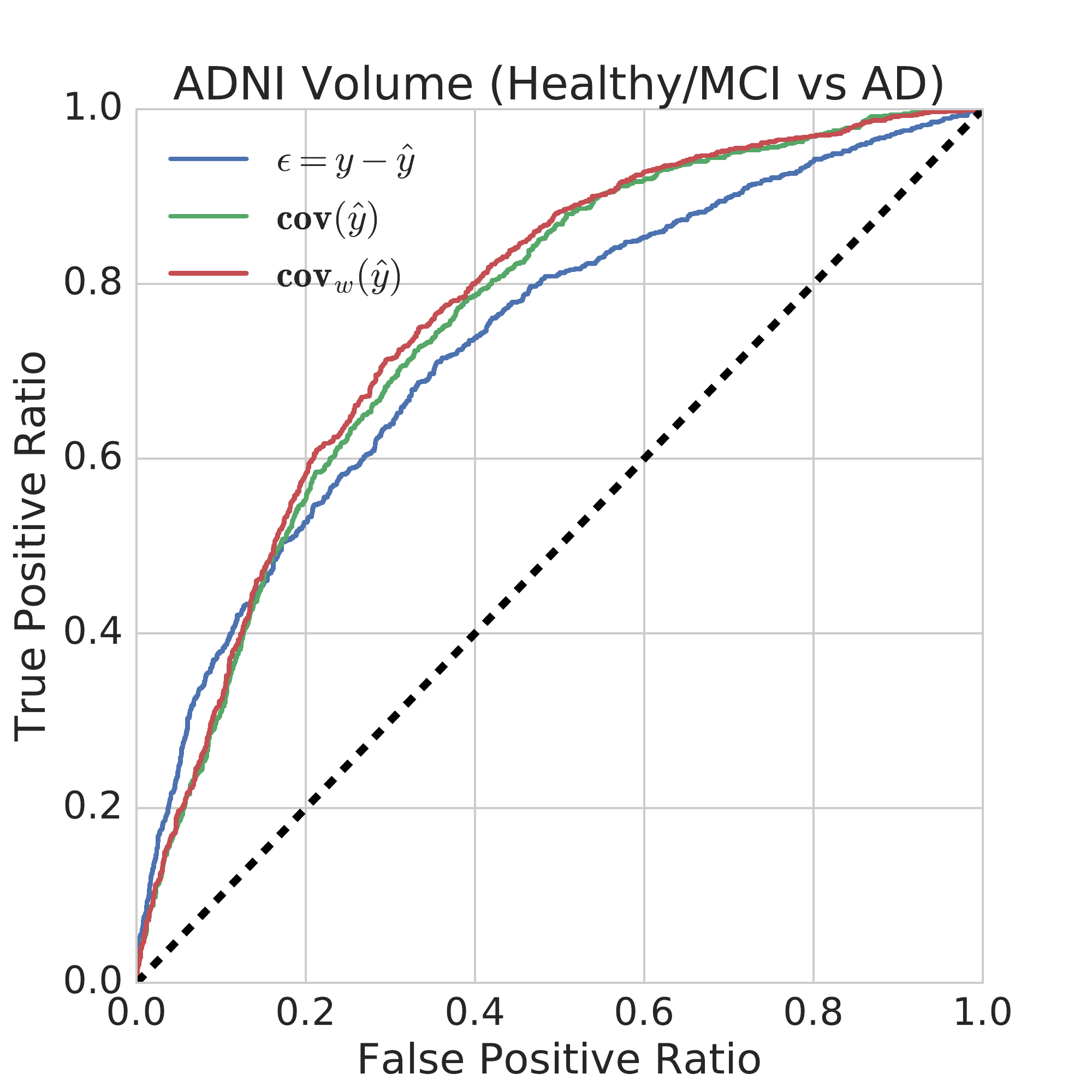}
	\end{subfigure}
	\begin{subfigure}{.245\textwidth}
		\centering\includegraphics[width=\textwidth]{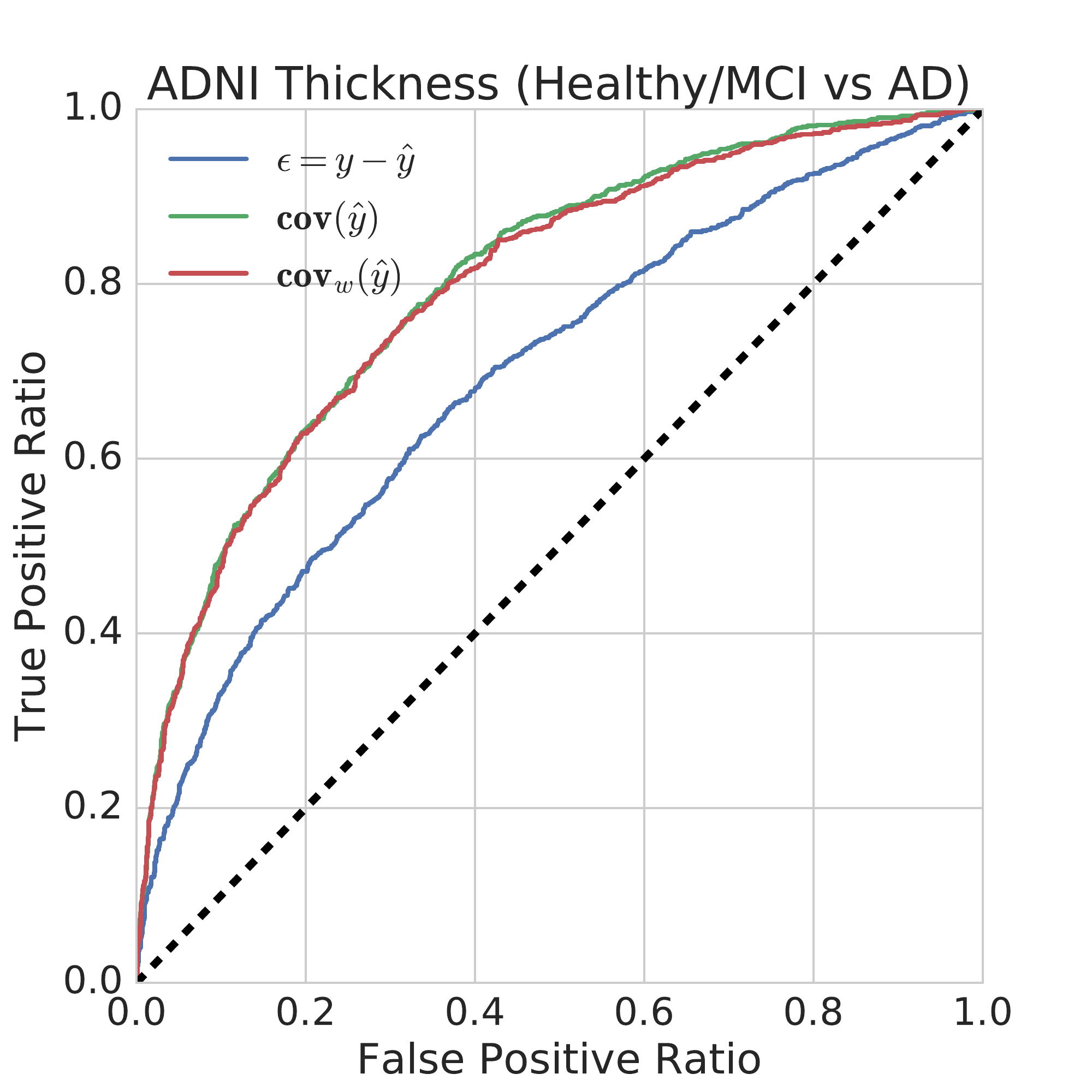}
	\end{subfigure}
	\begin{subfigure}{.245\textwidth}
		\centering\includegraphics[width=\textwidth]{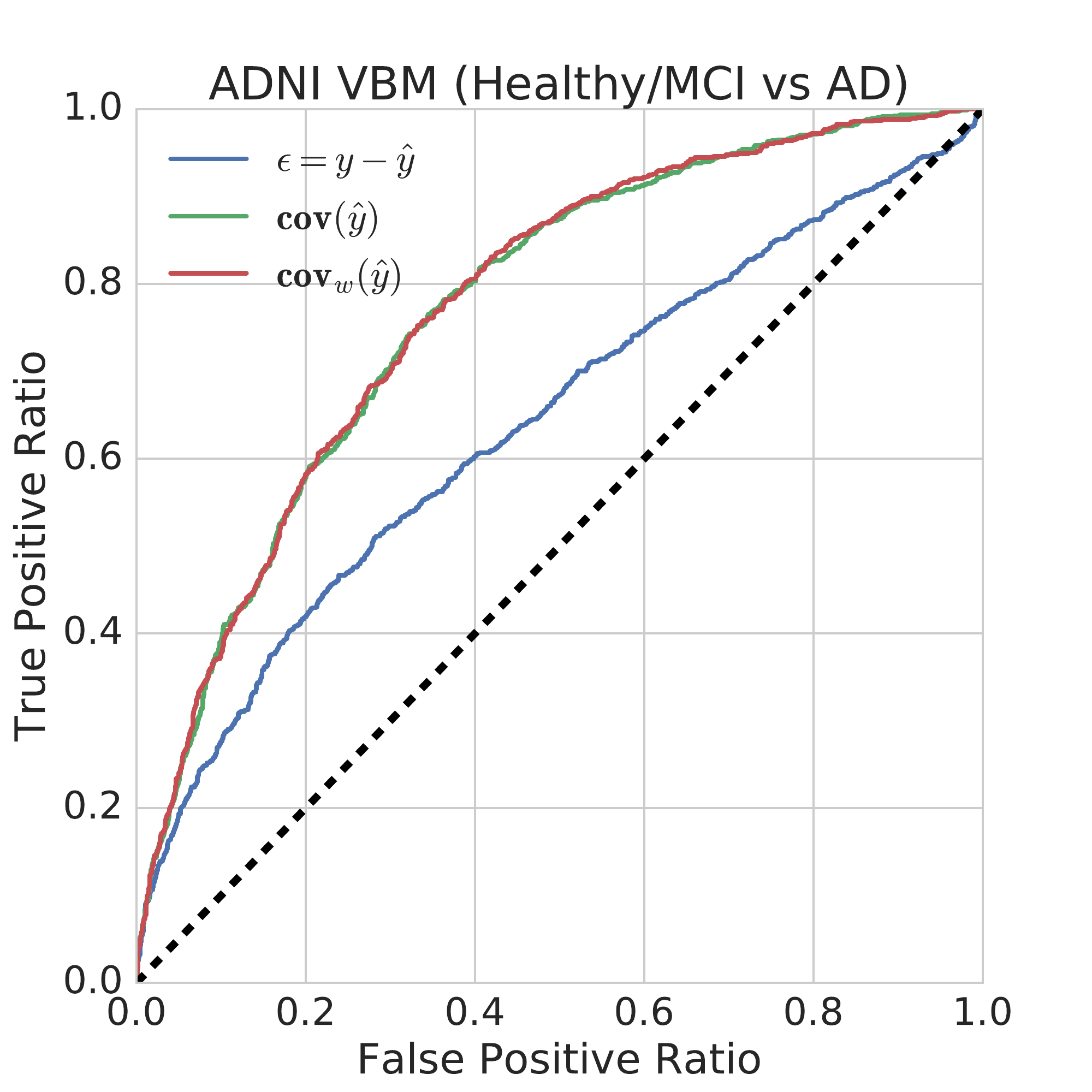}
	\end{subfigure}
	\begin{subfigure}{.245\textwidth}
		\centering\includegraphics[width=\textwidth]{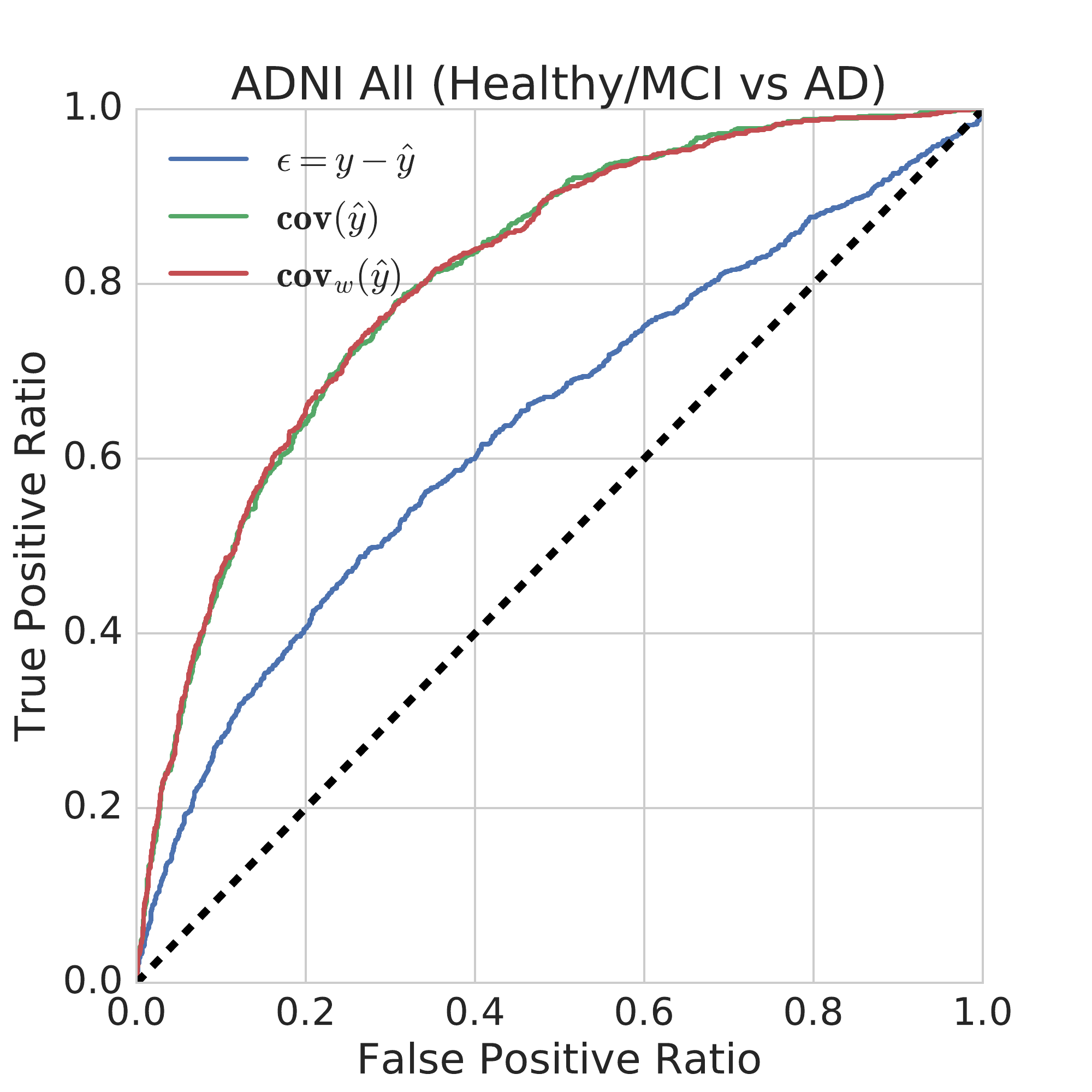}
	\end{subfigure}
	\caption{Receiver Operating Characteristic (ROC) curves for the prediction of the presence of MCI/AD (Top) or the presence of AD (Bottom) evaluated on the ADNI dataset. Columns correspond to the different evaluated features.}
	\label{fig:ROC_ADNI}
\end{figure*}

\begin{figure*}[h!]
	\begin{subfigure}{.245\textwidth}
		\centering\includegraphics[width=\textwidth]{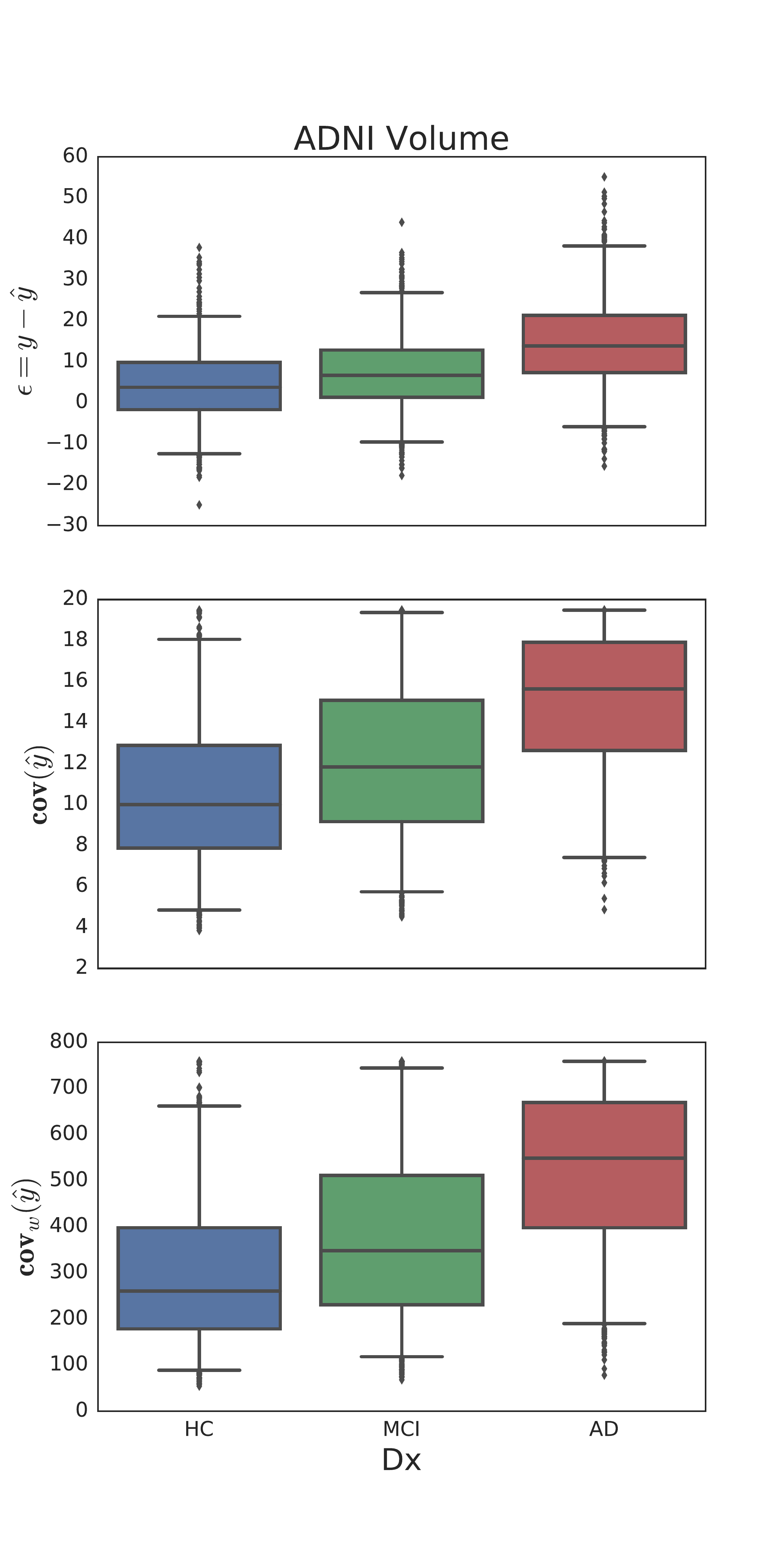}
	\end{subfigure}
	\begin{subfigure}{.245\textwidth}
		\centering\includegraphics[width=\textwidth]{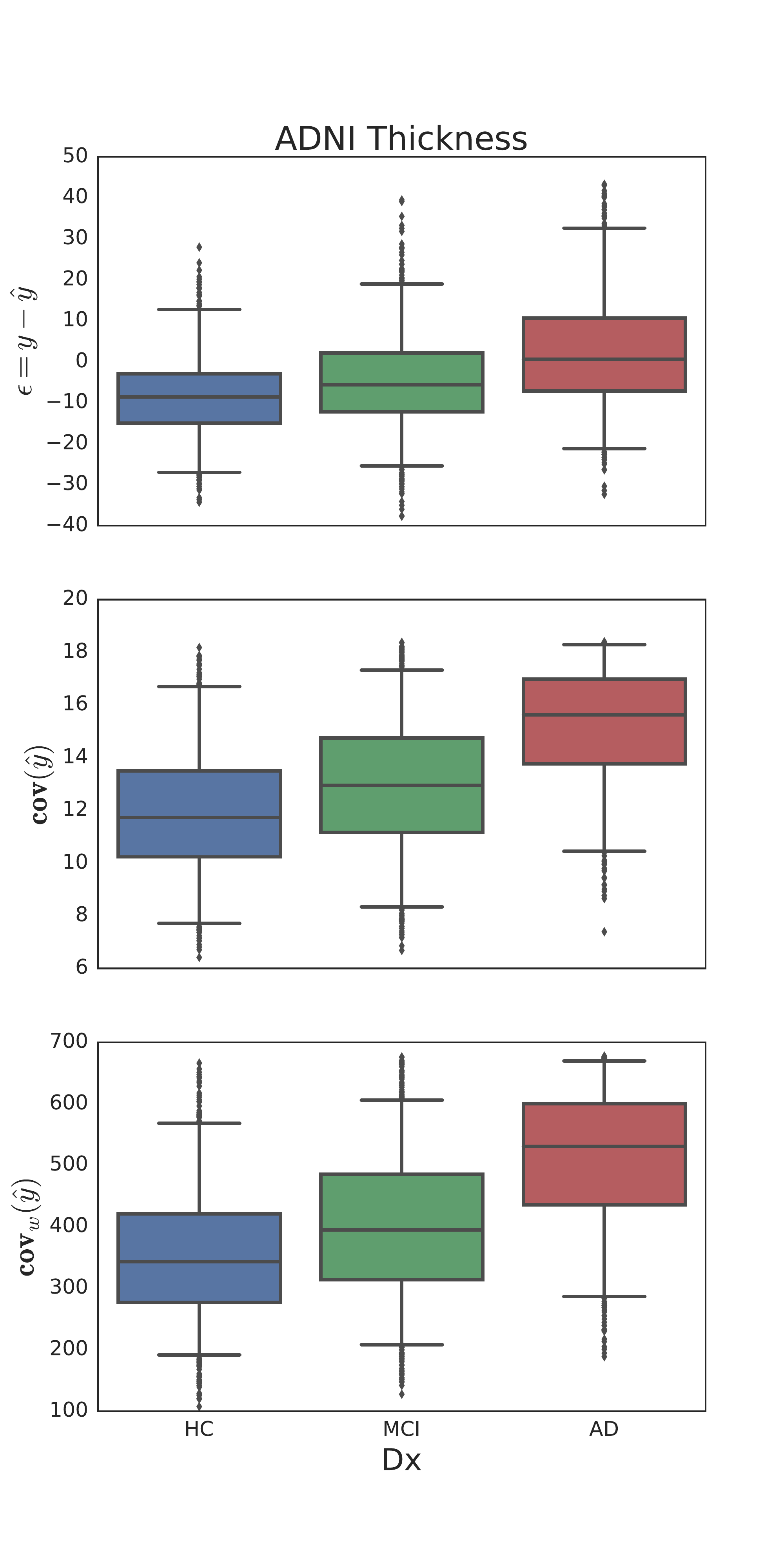}
	\end{subfigure}
	\begin{subfigure}{.245\textwidth}
		\centering\includegraphics[width=\textwidth]{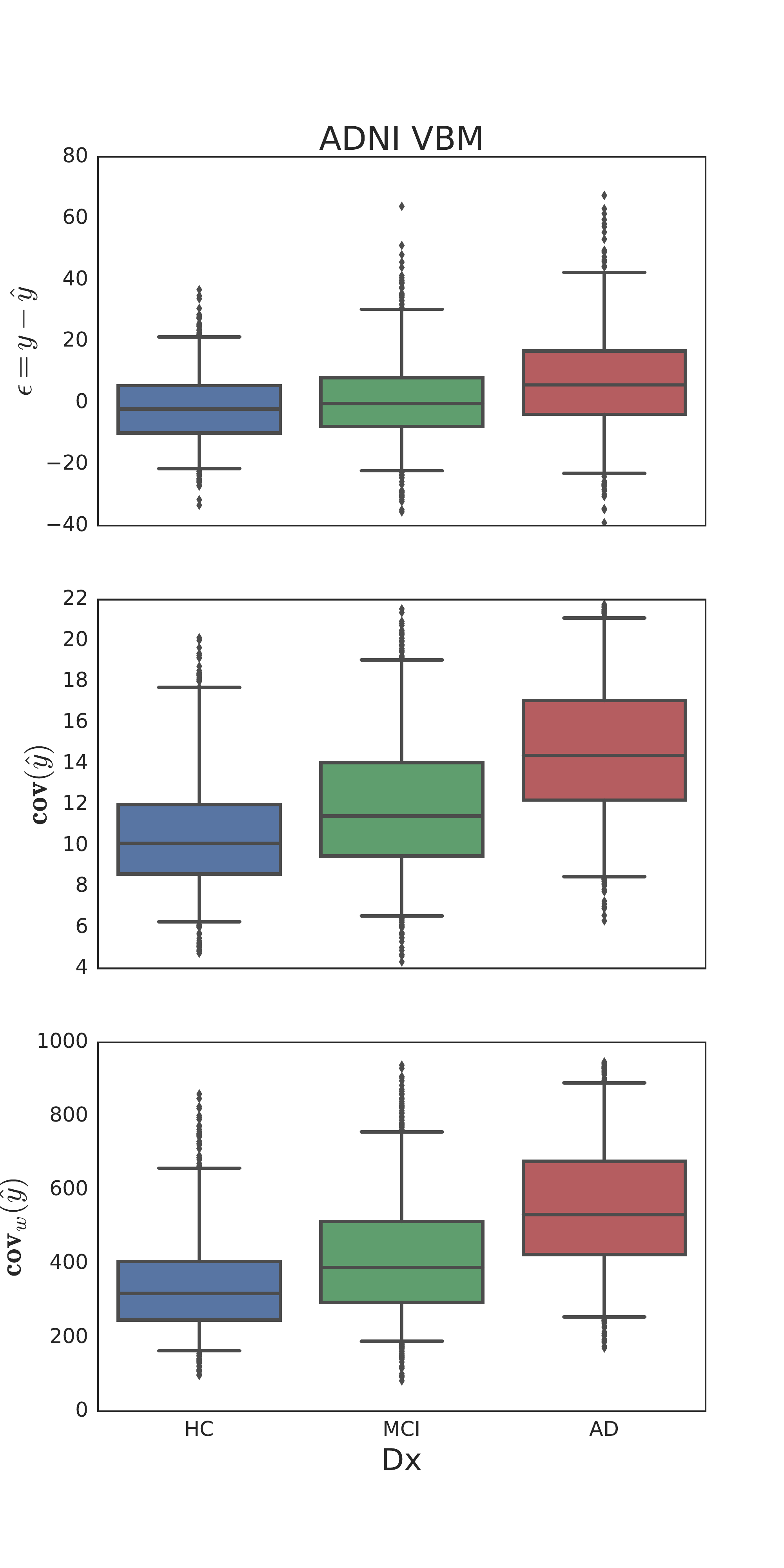}
	\end{subfigure}
	\begin{subfigure}{.245\textwidth}
		\centering\includegraphics[width=\textwidth]{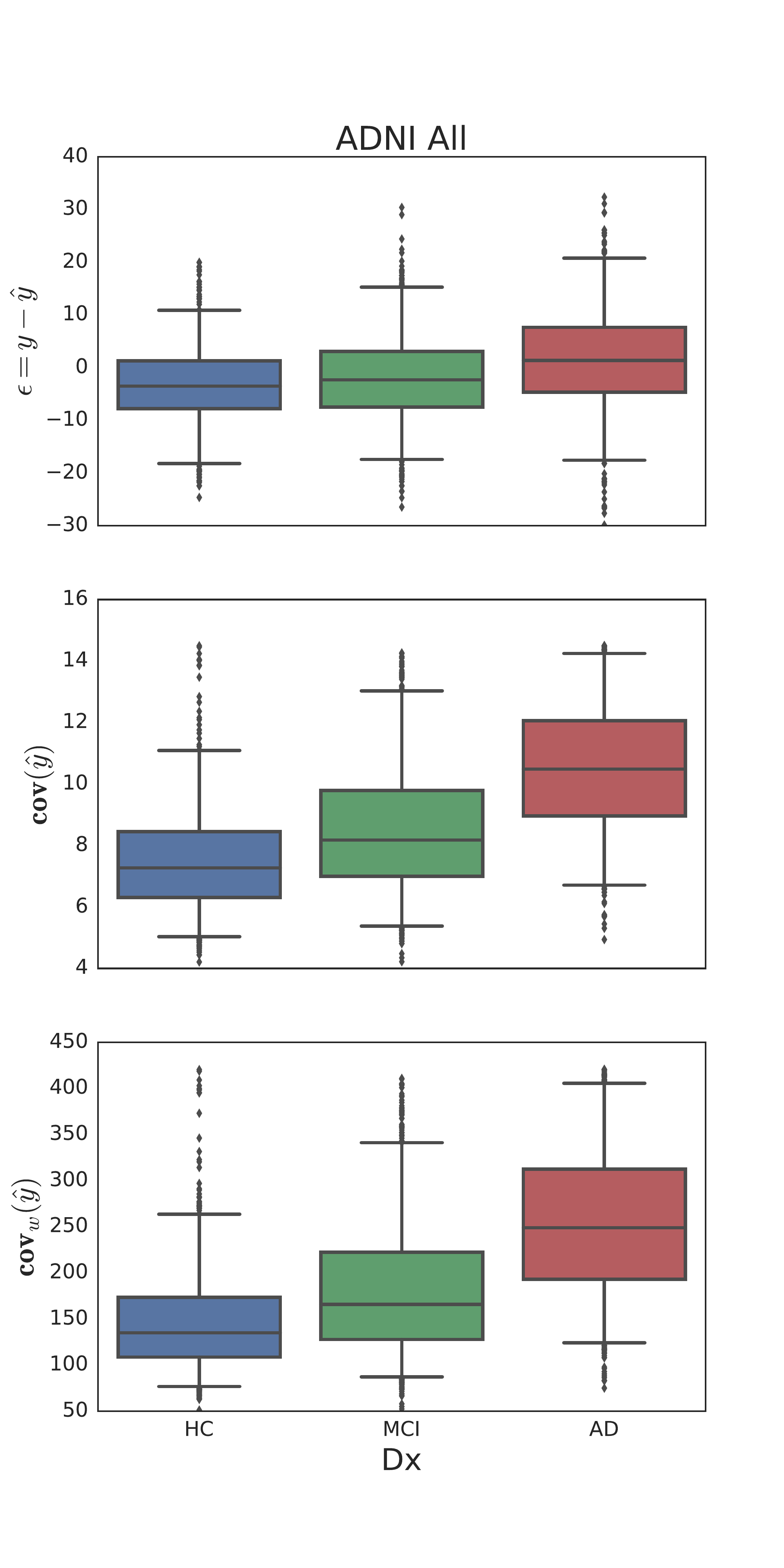}
	\end{subfigure}
	
	\caption{Box plots showing prediction results for $\epsilon$ (top), $\cov$ (middle) and $\covw$ (bottom)  for HC, MCI and AD groups on the ADNI dataset. Columns correspond to the different evaluated features. }
	\label{fig:all_box_ADNI}
\end{figure*}

\subsubsection{Experiment 1: ADNI Dataset}  
 For our first experiment we measure the separation between HC, MCI and AD groups for images obtained from the ADNI database. Due to the very large dataset size of this testing scenario, all the p-values reported in table \ref{table:pvalues-adni} are statistically significant (p-value \textless \ \num{6e-5}). The reported AUC values in table \ref{table:auc-adni} and the ROC curves in figure \ref{fig:ROC_ADNI} show consistently a better performance of the uncertainty based metrics $\cov$ and $\covw$ with respect to $\epsilon$.  In general $\cov$ and $\covw$ presented similar performance, but adding the age term resulted in larger AUC values for the experiments on volume and VBM features. Box plots for each feature set and each diagnostic group are also shown in figure \ref{fig:all_box_ADNI}. The results using uncertainty based measures $\cov$ and $\covw$ were strongly correlated (R =0.95). In contrast, correlations of 0.35 and 0.37 were obtained between $\epsilon$-$\cov$ and $\epsilon$-$\covw$, respectively.

\subsubsection{Experiment 2: OASIS Dataset}  
\begin{figure*}[h!]
	\begin{subfigure}{.245\textwidth}
		\centering\includegraphics[width=\textwidth]{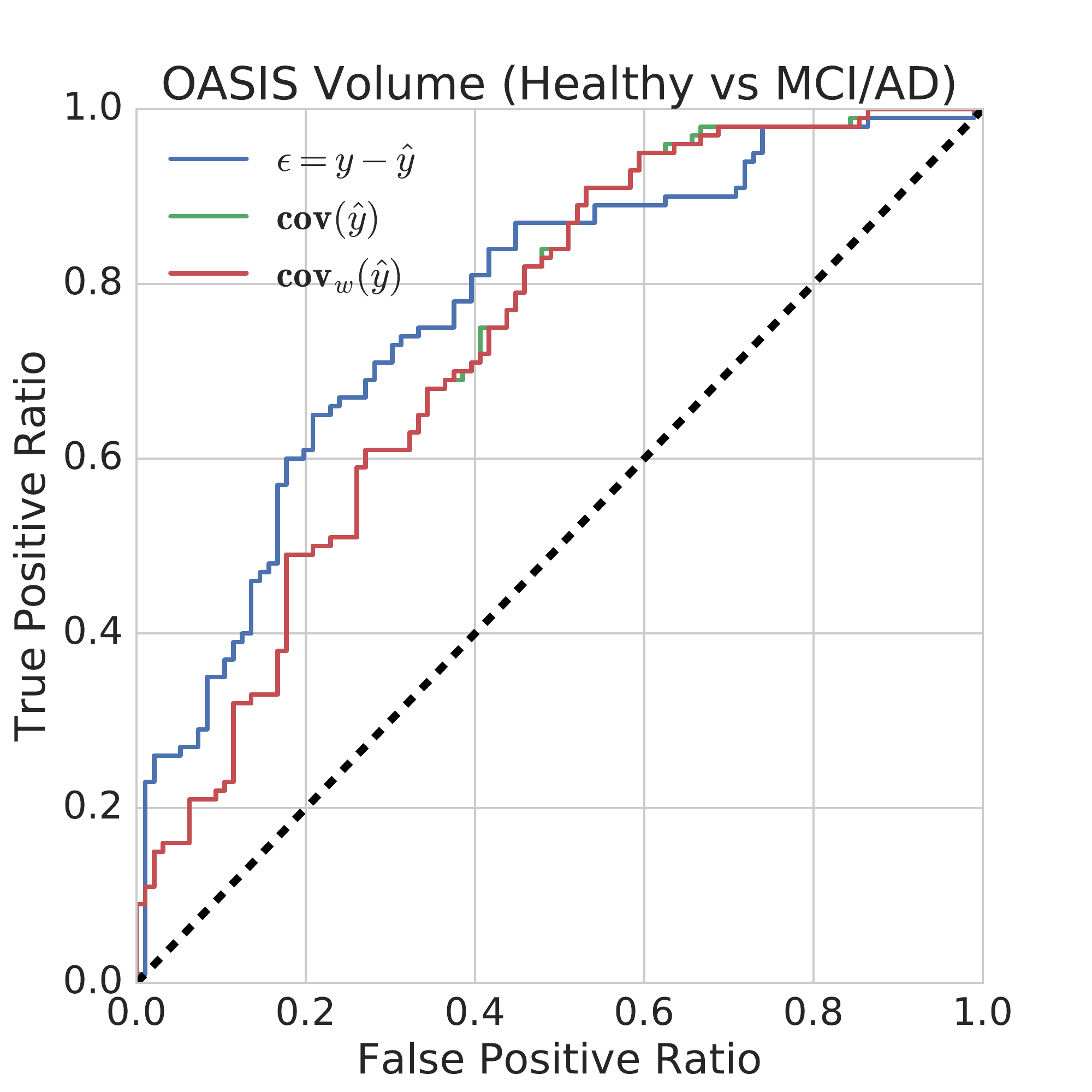}
	\end{subfigure}
	\begin{subfigure}{.245\textwidth}
		\centering\includegraphics[width=\textwidth]{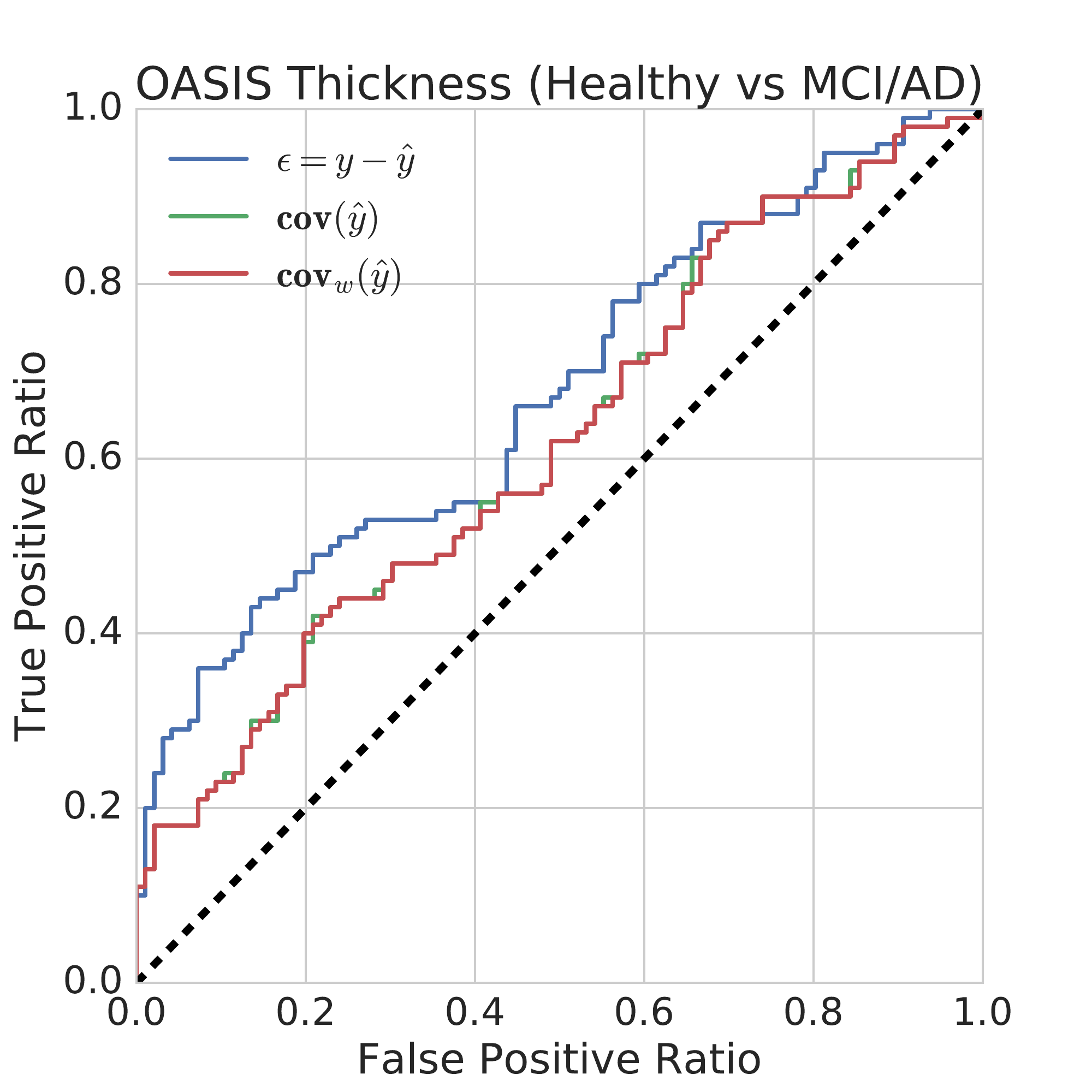}
	\end{subfigure}
	\begin{subfigure}{.245\textwidth}
		\centering\includegraphics[width=\textwidth]{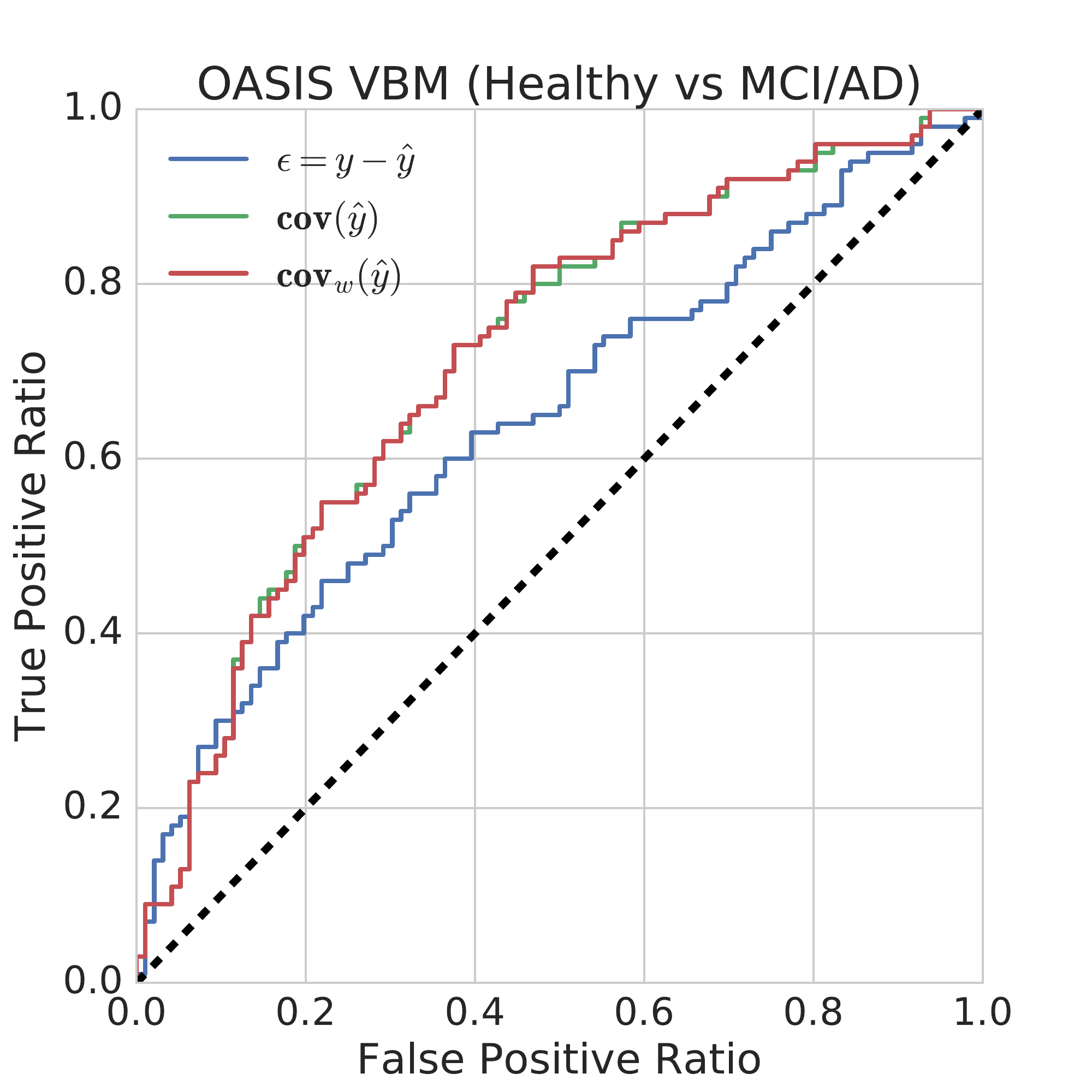}
	\end{subfigure}
	\begin{subfigure}{.245\textwidth}
		\centering\includegraphics[width=\textwidth]{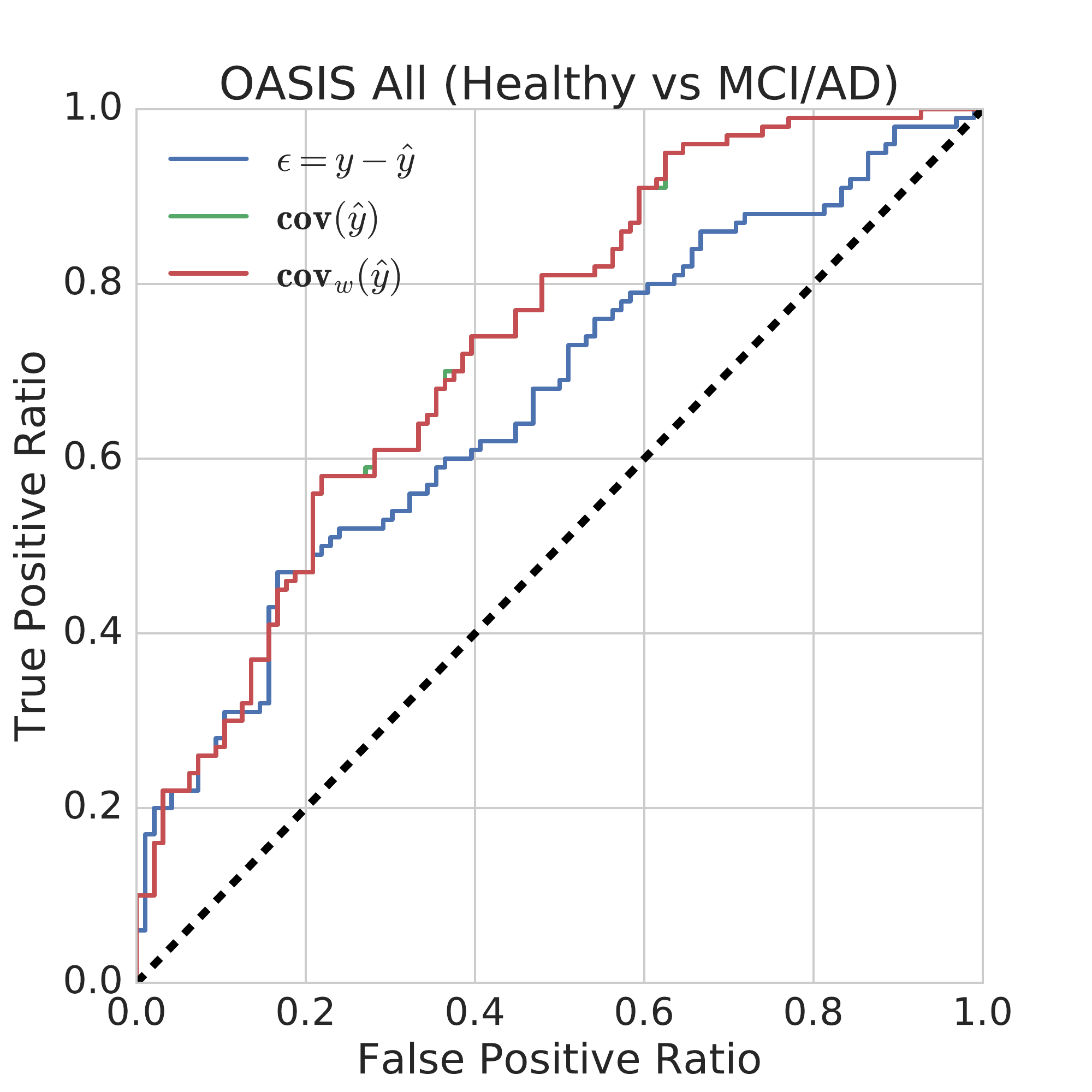}
	\end{subfigure}
	\begin{subfigure}{.245\textwidth}
		\centering\includegraphics[width=\textwidth]{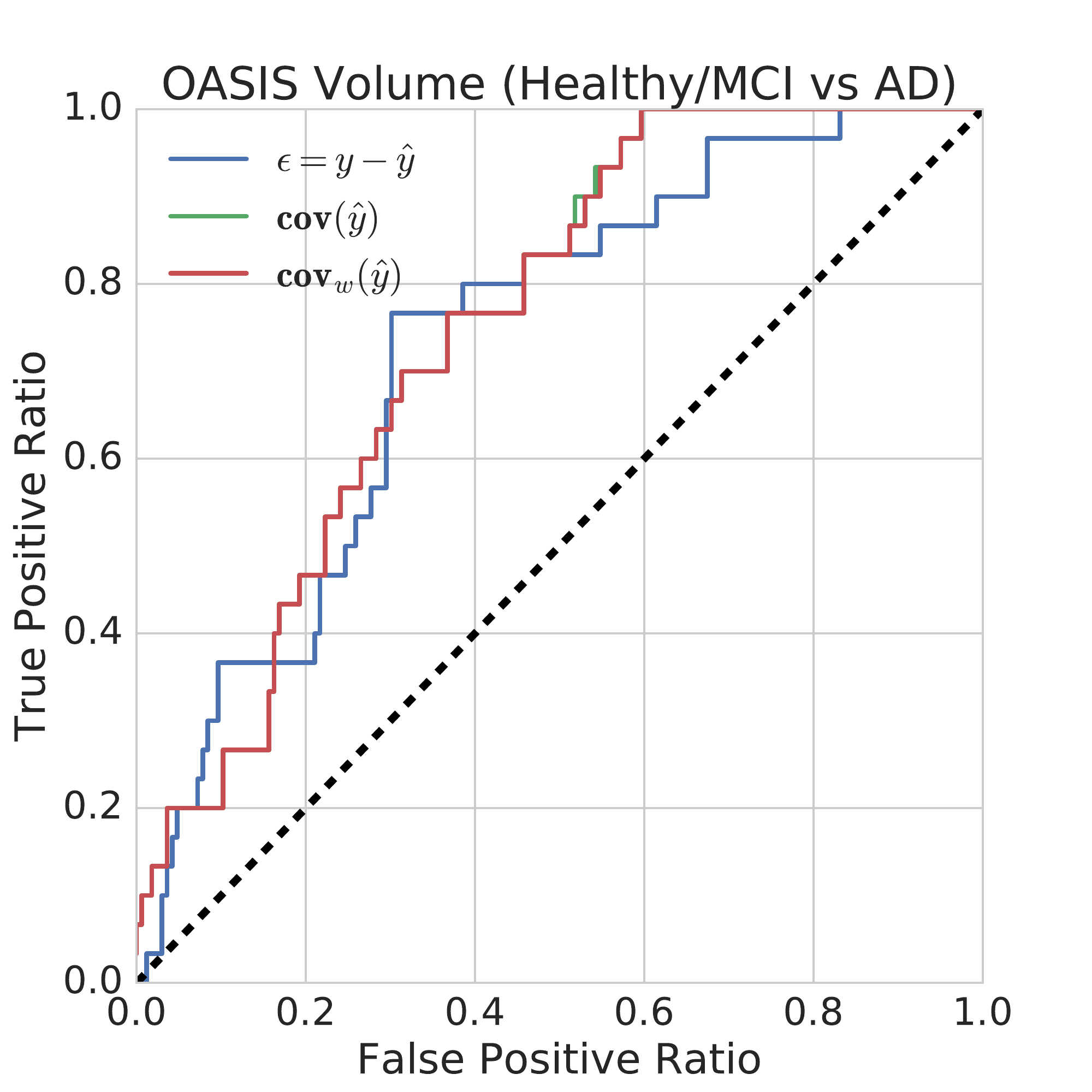}
	\end{subfigure}
	\begin{subfigure}{.245\textwidth}
		\centering\includegraphics[width=\textwidth]{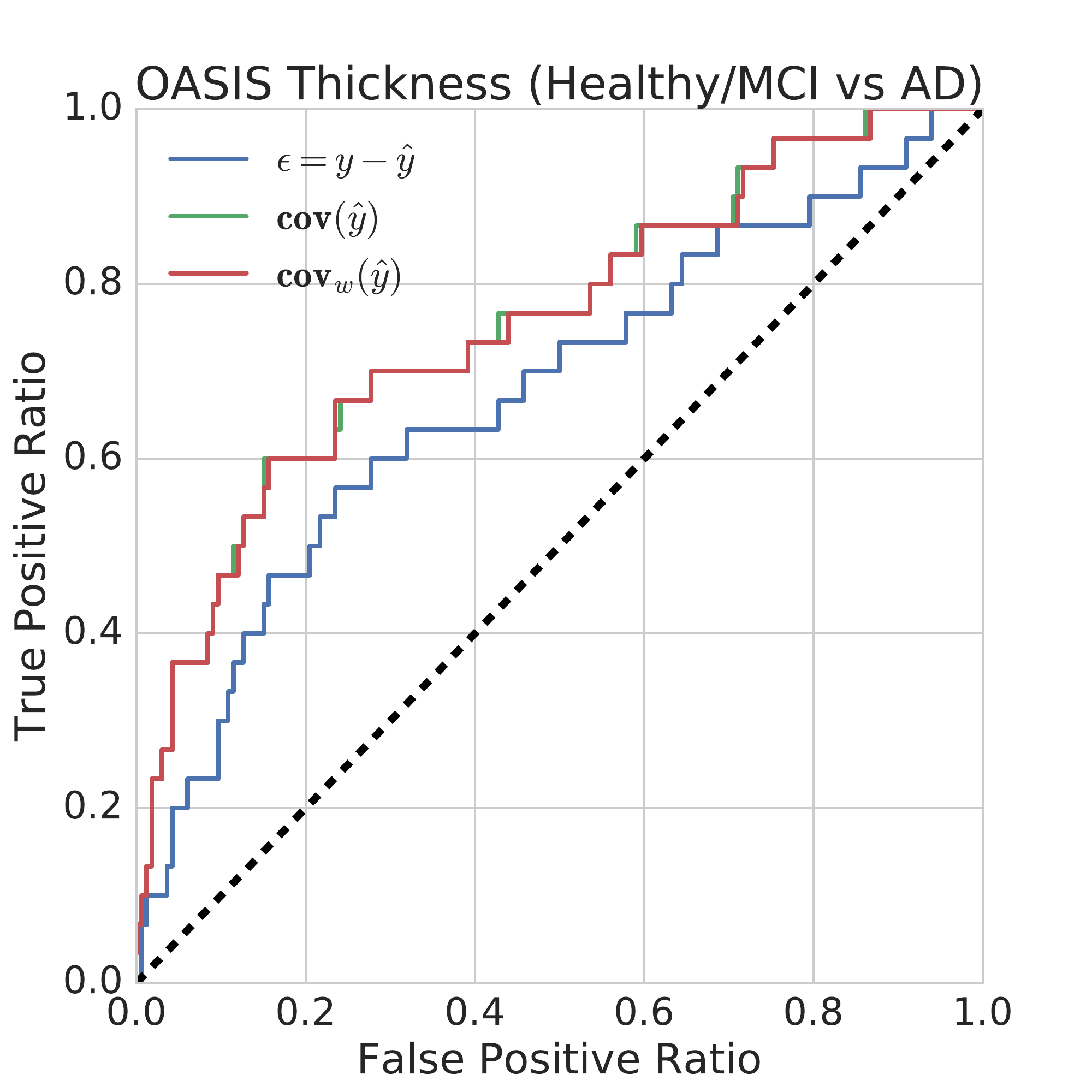}
	\end{subfigure}
	\begin{subfigure}{.245\textwidth}
		\centering\includegraphics[width=\textwidth]{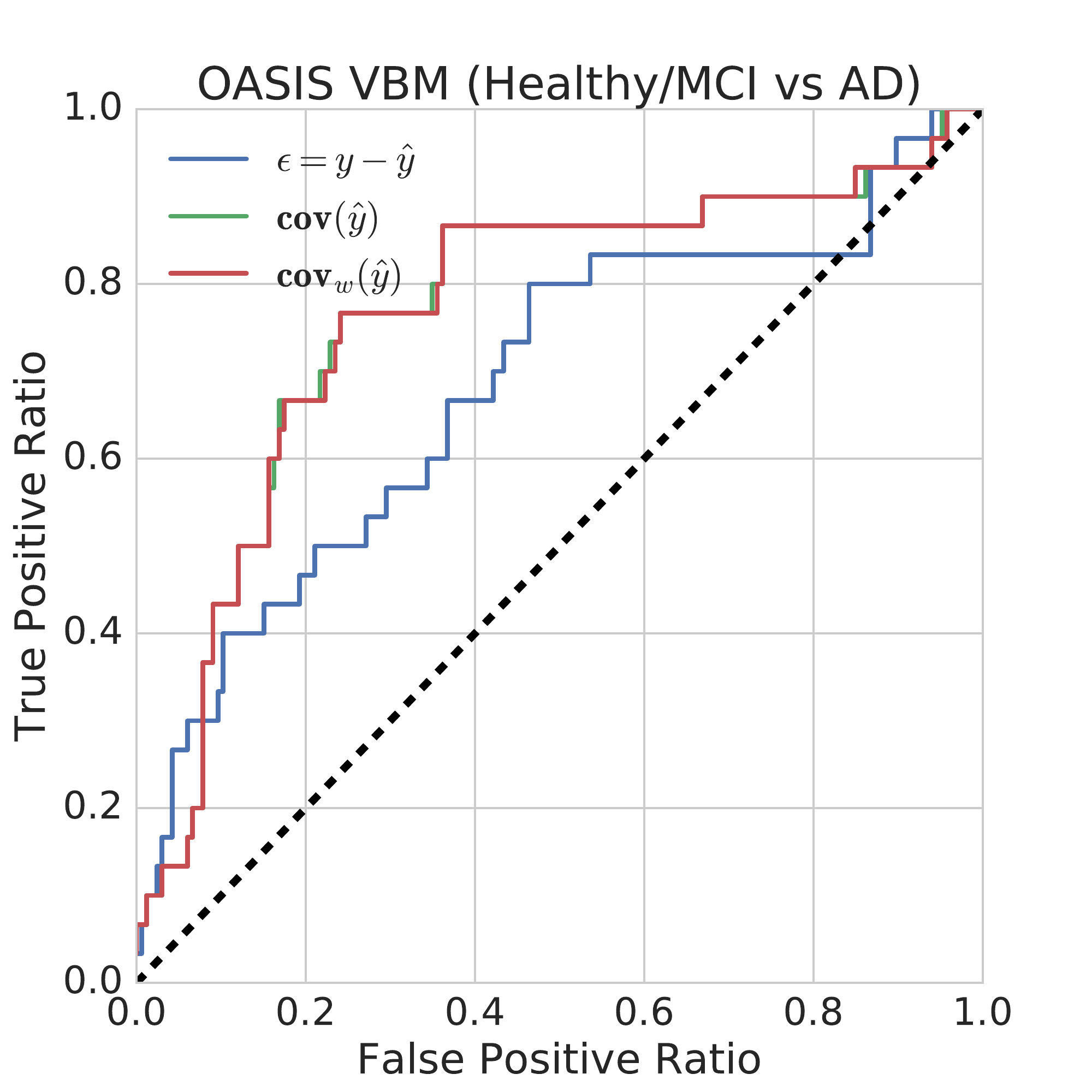}
	\end{subfigure}
	\begin{subfigure}{.245\textwidth}
		\centering\includegraphics[width=\textwidth]{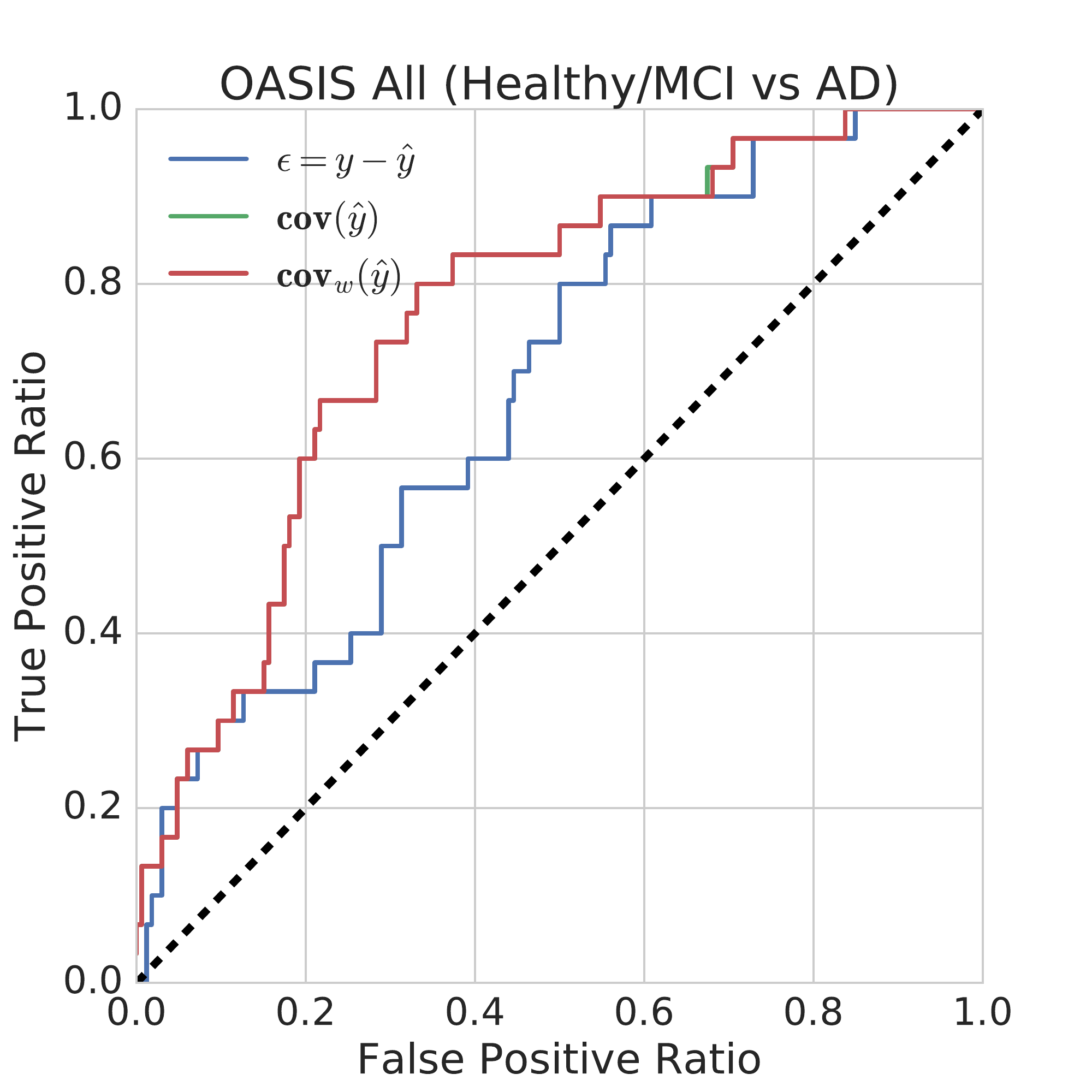}
	\end{subfigure}
	\caption{Receiver Operating Characteristic (ROC) curves for the prediction of the presence of MCI/AD (Top) or the presence of AD (Bottom) evaluated on the OASIS dataset. Columns correspond to the different evaluated features.}
	\label{fig:roc_oasis}
\end{figure*}

\begin{figure*}[h!]
	\begin{subfigure}{.245\textwidth}
		\centering\includegraphics[width=\textwidth]{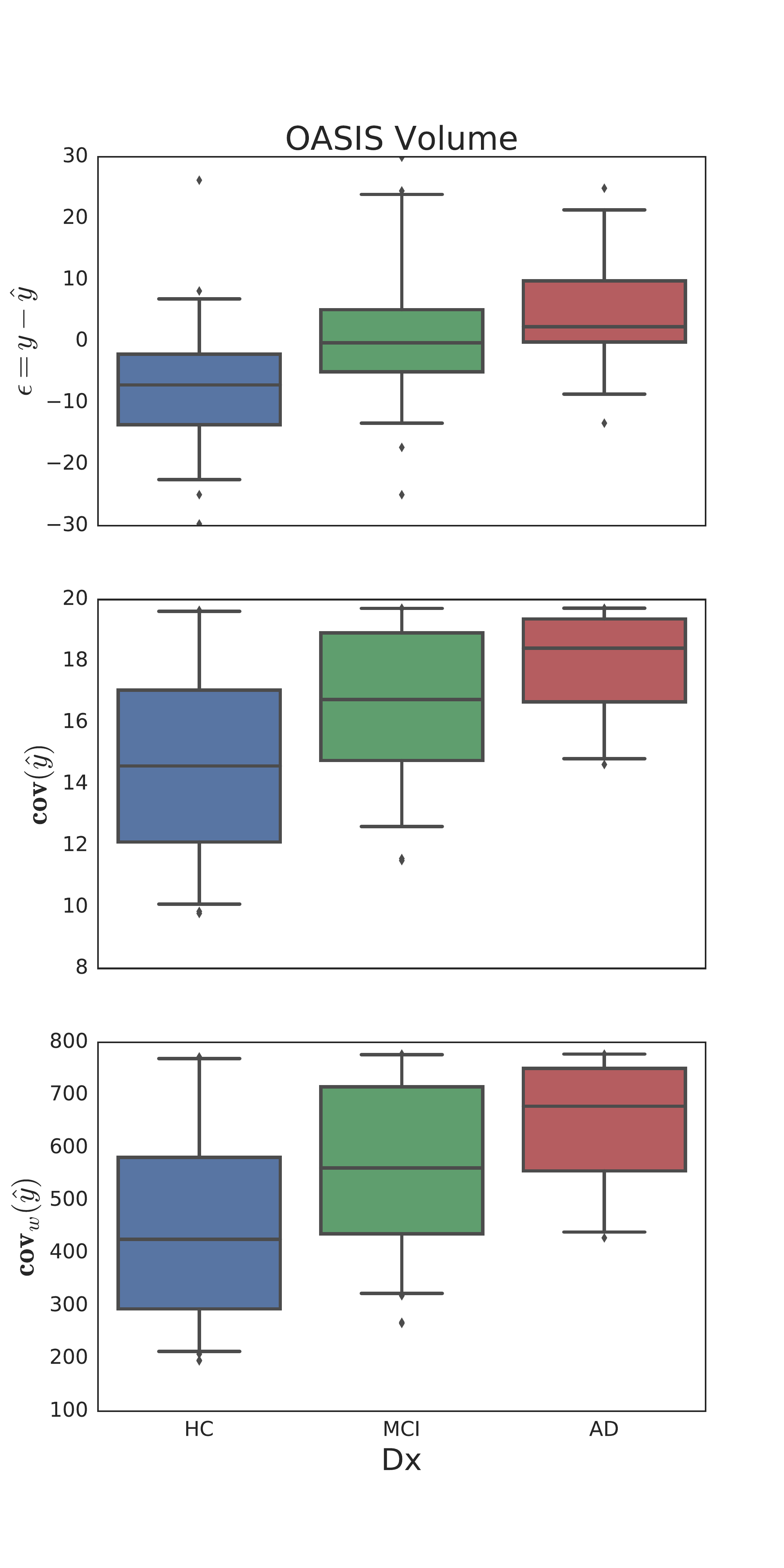}
	\end{subfigure}
	\begin{subfigure}{.245\textwidth}
		\centering\includegraphics[width=\textwidth]{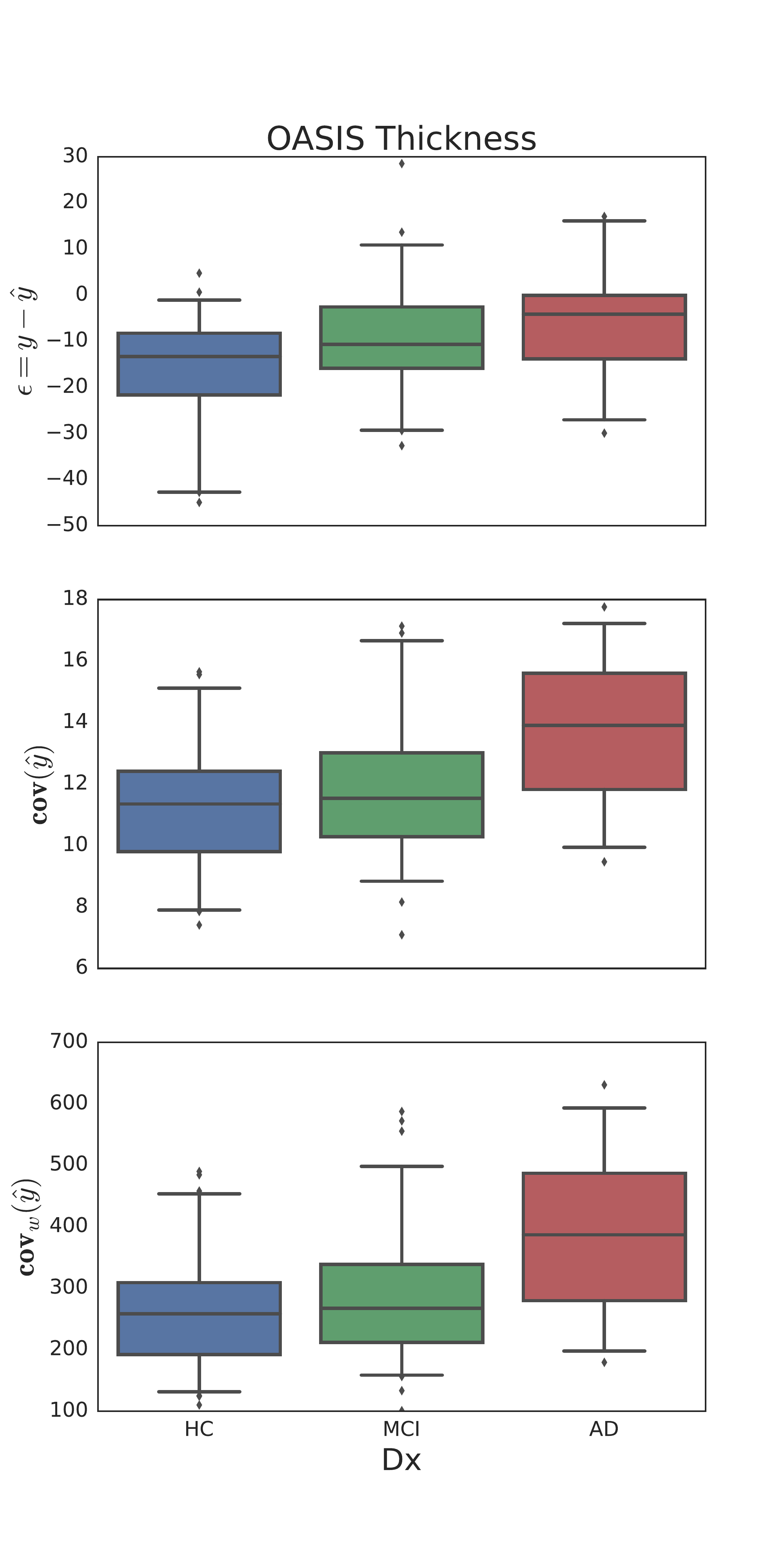}
	\end{subfigure}
	\begin{subfigure}{.245\textwidth}
		\centering\includegraphics[width=\textwidth]{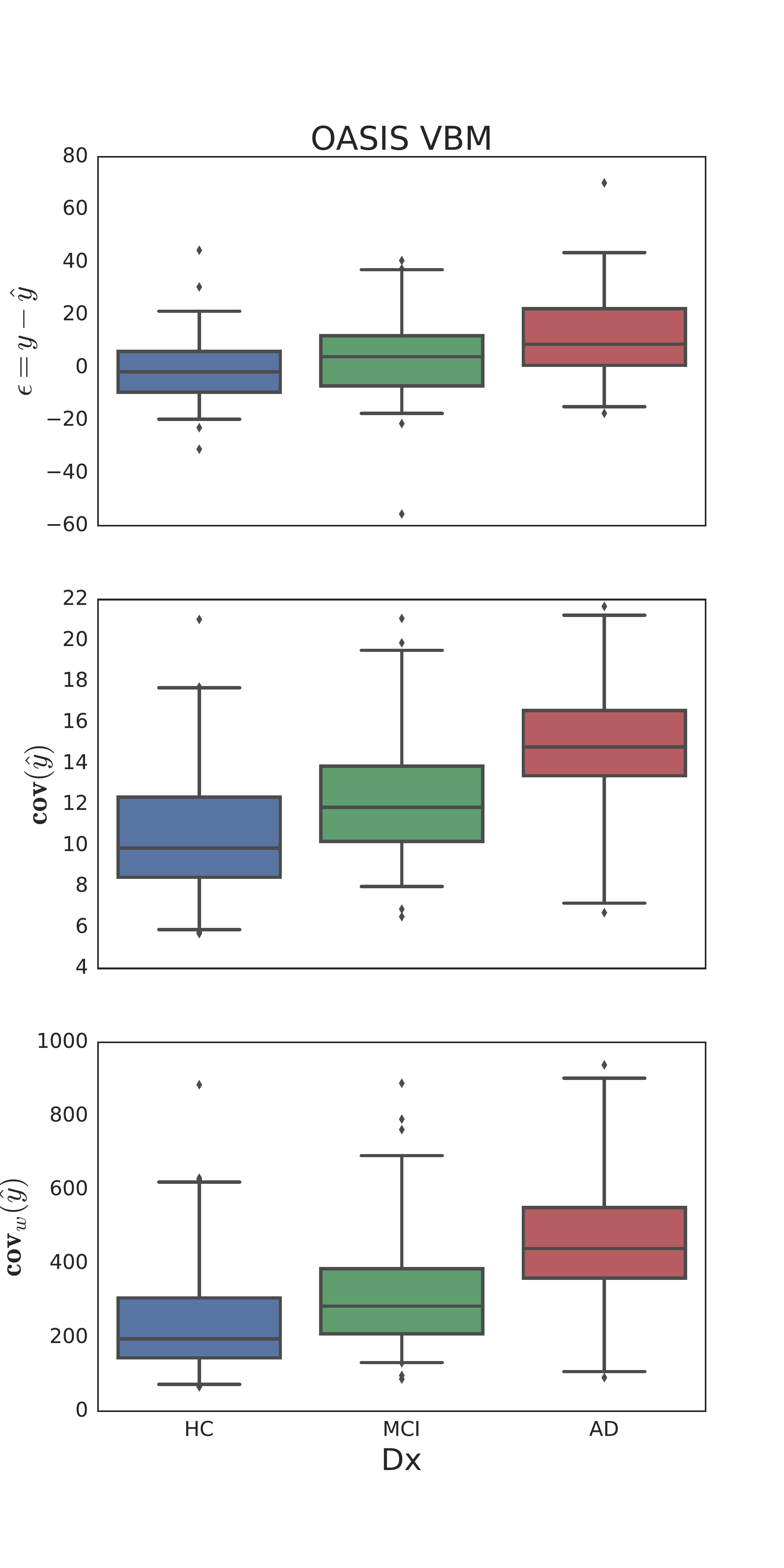}
	\end{subfigure}
	\begin{subfigure}{.245\textwidth}
		\centering\includegraphics[width=\textwidth]{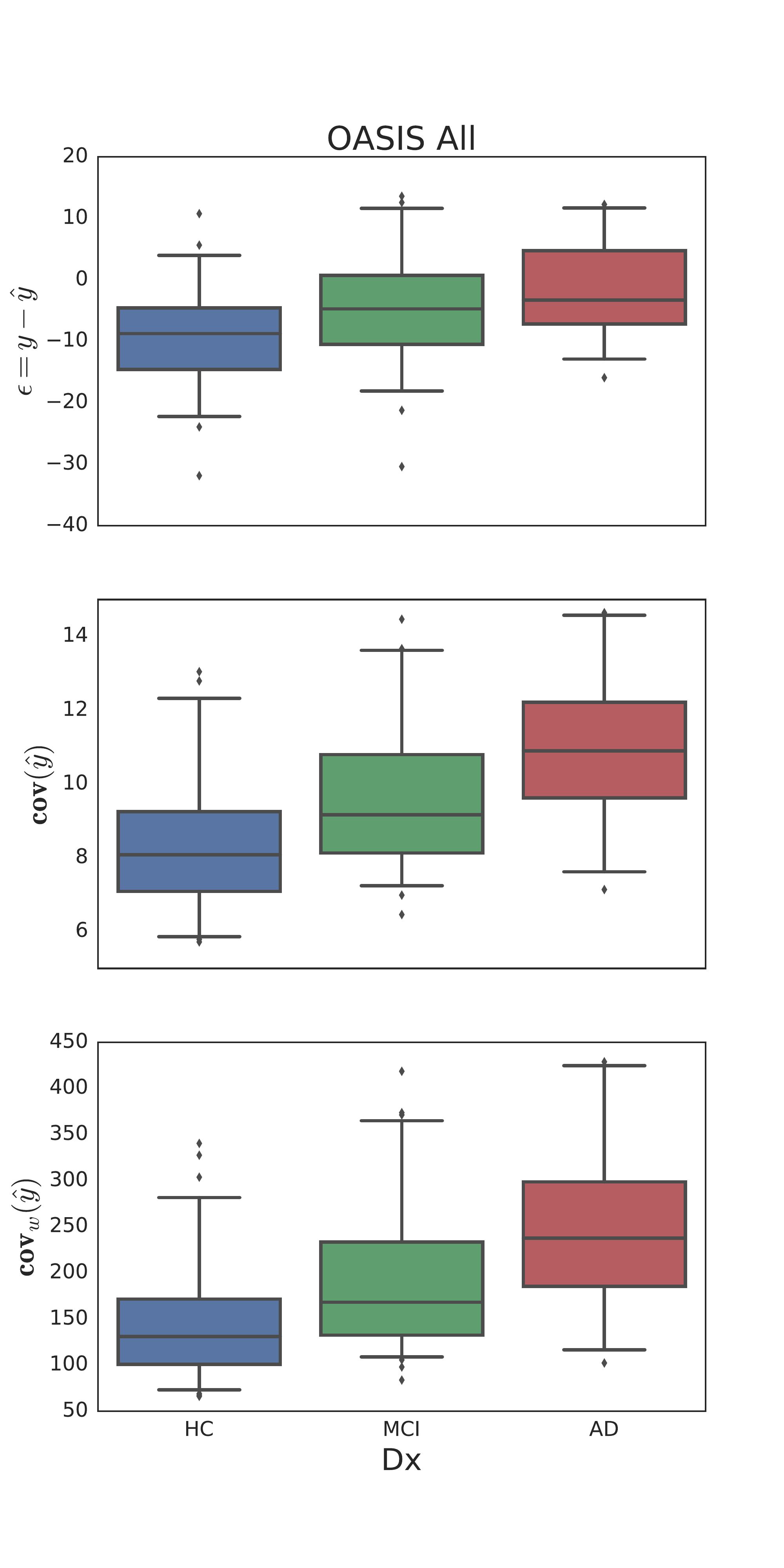}
	\end{subfigure}
	\caption{Box plots showing prediction results for $\epsilon$ (top), $\cov$ (middle) and $\covw$ (bottom)  for HC, MCI and AD groups on the OASIS dataset. Columns correspond to the different evaluated features.}
	\label{fig:all_box_oasis}
\end{figure*}

Our second experiment is similar to experiment 1, but our evaluation is performed on images obtained from the OASIS database. In order to ensure similar age ranges for the HC, MCI and AD groups, all individuals under 60 years  were removed from the testing dataset. Tables \ref{table:pvalues-oasis} and \ref{table:auc-oasis} summarize the numerical results of the comparisons between HC-MCI and MCI-AD groups. Similar to previous results \citep{Franke2010}, we observed that prediction error $\epsilon$ is a useful biomarker in this particular dataset. According to the results in tables \ref{table:pvalues-oasis} and \ref{table:auc-oasis}, $\epsilon$ presented larger AUC values and smaller p-values for the models trained using volume and thickness features when discriminating between HC and MCI groups. However for all the rest of the evaluations on the OASIS database, $\cov$ and $\covw$ presented the best performance amongst the evaluated metrics. Correlation coefficients between the metrics were 0.99 for $\cov$-$\covw$, 0.38 for $\epsilon$-$\cov$  and 0.38 for $\epsilon$-$\covw$. Notice that in this case the results for $\cov$ and $\covw$ are strongly correlated due to the very large value assigned to the age length scale parameter $l_y$. The ROC curves in figure \ref{fig:roc_oasis} and the box plots in figure \ref{fig:all_box_oasis} confirm these observations.  
\begin{table}[h]
	\centering
	\resizebox{0.7\columnwidth}{!}
	{
		\begin{tabular}{lccc|ccc}
			\multicolumn{1}{c}{} &\multicolumn{3}{c}{HC-MCI} & \multicolumn{3}{c}{MCI-AD}  \tabularnewline
			& $\epsilon$ & $\cov$ & $\covw$ & $\epsilon$ & $\cov$ & $\covw$ 
			\tabularnewline
			Volume  & \textless\zz{0.0001} &\textless\zz{0.0001} &\textless\zz{ 0.0001 } &\zz{0.0611 } &\zz{ 0.0065 } &\zz{ 0.0067 } \tabularnewline
			Thickness& \zz{0.0005 } &\zz{ 0.1186 } &\zz{ 0.1123 } &\zz{0.0707 } &\zz{ 0.0002 } &\zz{ 0.0002 } \tabularnewline
			VBM  & \zz{0.0075 } &\textless\zz{0.0001} &\textless\zz{0.0001} &\zz{0.0225 } &\zz{ 0.0003 } &\zz{ 0.0003 }  \tabularnewline
			All  & \zz{0.0015 } &\textless\zz{0.0001} &\textless\zz{ 0.0001 } &\zz{0.0707 } &\zz{ 0.0020 } &\zz{ 0.0020 } \tabularnewline			
		\end{tabular} 
	} \caption{p-values corresponding  to the statistical tests performed on the experiments comparing the HC, MCI and AD groups on the OASIS dataset. The highlighted values correspond to p values with significance levels under 0.05 (light background), 0.01 (middle background) and 0.001 (dark background).}
	\label{table:pvalues-oasis}
\end{table}

\begin{table}[h]
	\centering

	{
		\begin{tabular}{lccc|ccc}
			\multicolumn{1}{c}{} &\multicolumn{3}{c}{HC-MCI} & \multicolumn{3}{c}{MCI-AD}  \tabularnewline
			& $\epsilon$ & $\cov$ & $\covw$ & $\epsilon$ & $\cov$ & $\covw$  \tabularnewline
			\hline
			\rowcolor{lightgray} 
			Volume  & \textbf{0.77} & 0.73 & 0.73 & 0.73 &  \textbf{0.75} &\textbf{0.75} \tabularnewline
			Thickness  & \textbf{0.68} & 0.62 & 0.62 & 0.68 &  \textbf{0.76} &\textbf{0.76} \tabularnewline
			\rowcolor{lightgray} 
			VBM  & 0.64 & \textbf{0.72} & \textbf{0.72} & 0.68 &  \textbf{0.77} &\textbf{0.77} \tabularnewline
			All  & 0.67 & \textbf{0.74} & \textbf{0.74} & 0.68 &  \textbf{0.77} &\textbf{0.77} \tabularnewline
			
		\end{tabular} 
	} \caption{Area Under the Curve (AUC) values corresponding  to the statistical tests performed on the experiments comparing the HC, MCI and AD groups on the OASIS dataset.}
	\label{table:auc-oasis}
\end{table}

\subsubsection{Experiment 3: ABIDE II Dataset}  

\begin{figure*}[h!]
	\begin{subfigure}{.245\textwidth}
		\centering\includegraphics[width=\textwidth]{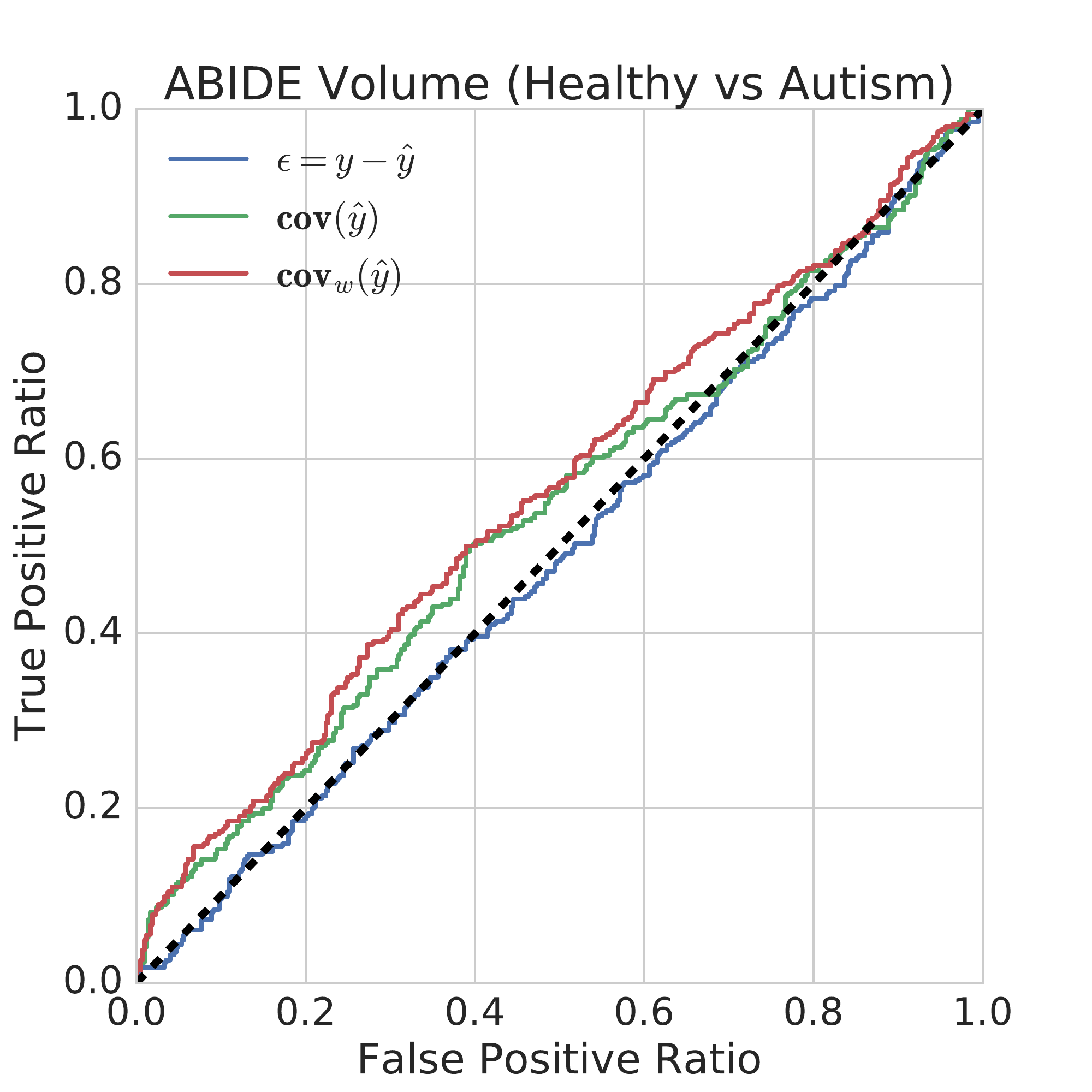}
	\end{subfigure}
	\begin{subfigure}{.245\textwidth}
		\centering\includegraphics[width=\textwidth]{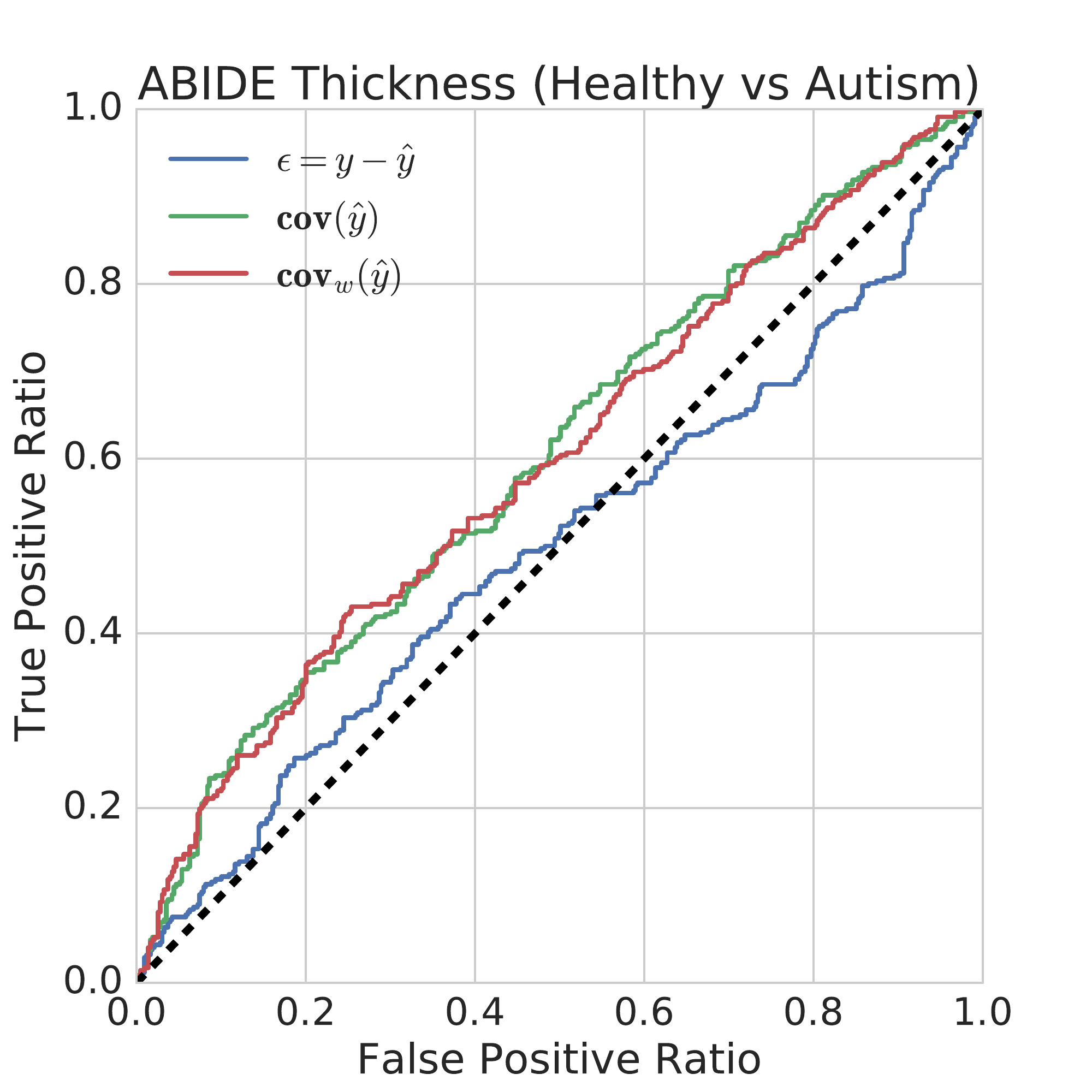}
	\end{subfigure}
	\begin{subfigure}{.245\textwidth}
		\centering\includegraphics[width=\textwidth]{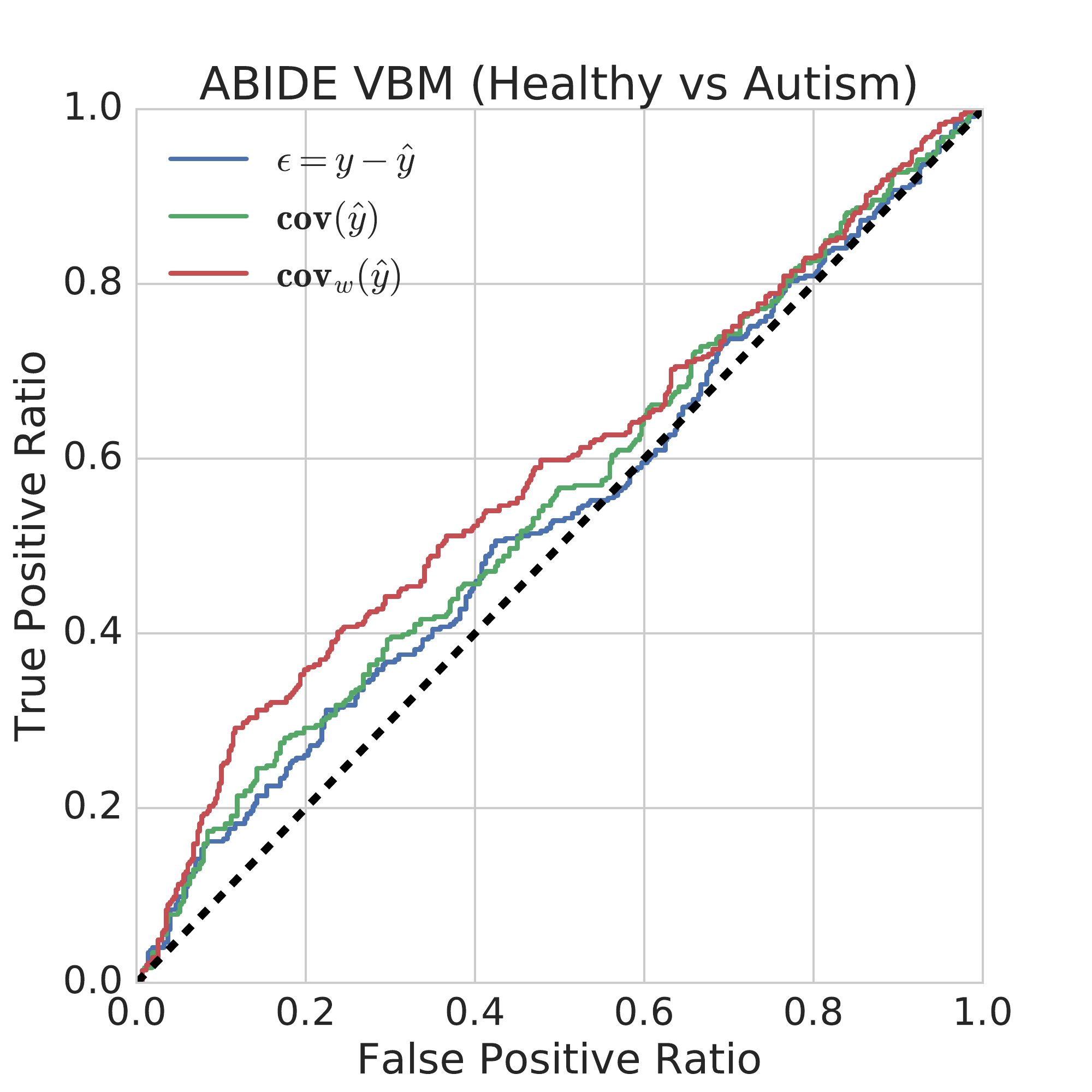}
	\end{subfigure}
	\begin{subfigure}{.245\textwidth}
		\centering\includegraphics[width=\textwidth]{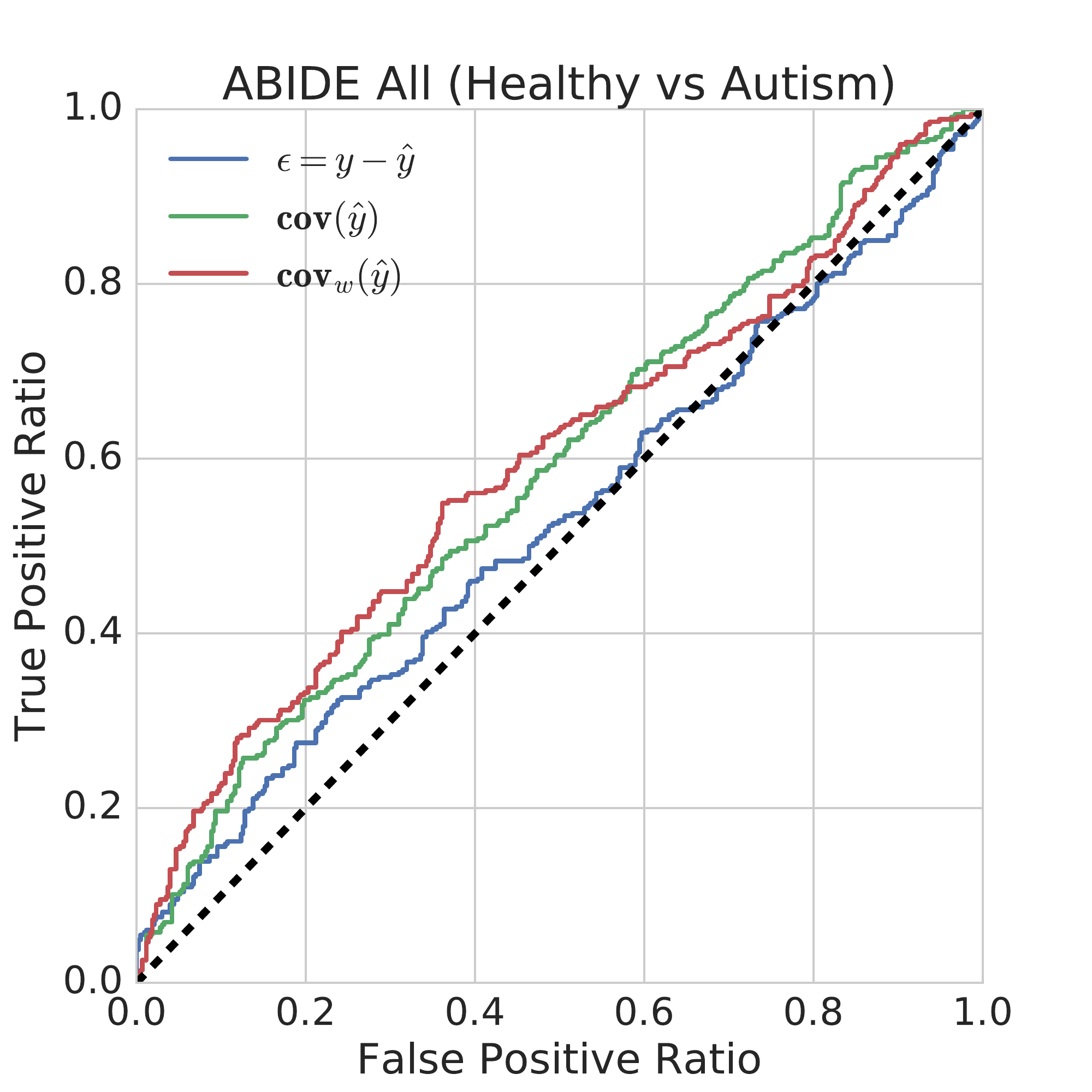}
	\end{subfigure}
	\caption{Receiver Operating Characteristic (ROC) curves for the prediction of the presence of Autism. Columns correspond to the different evaluated features.}
	\label{fig:roc_abide}
\end{figure*}

\begin{figure*}[h!]
	\begin{subfigure}{.245\textwidth}
		\centering\includegraphics[width=\textwidth]{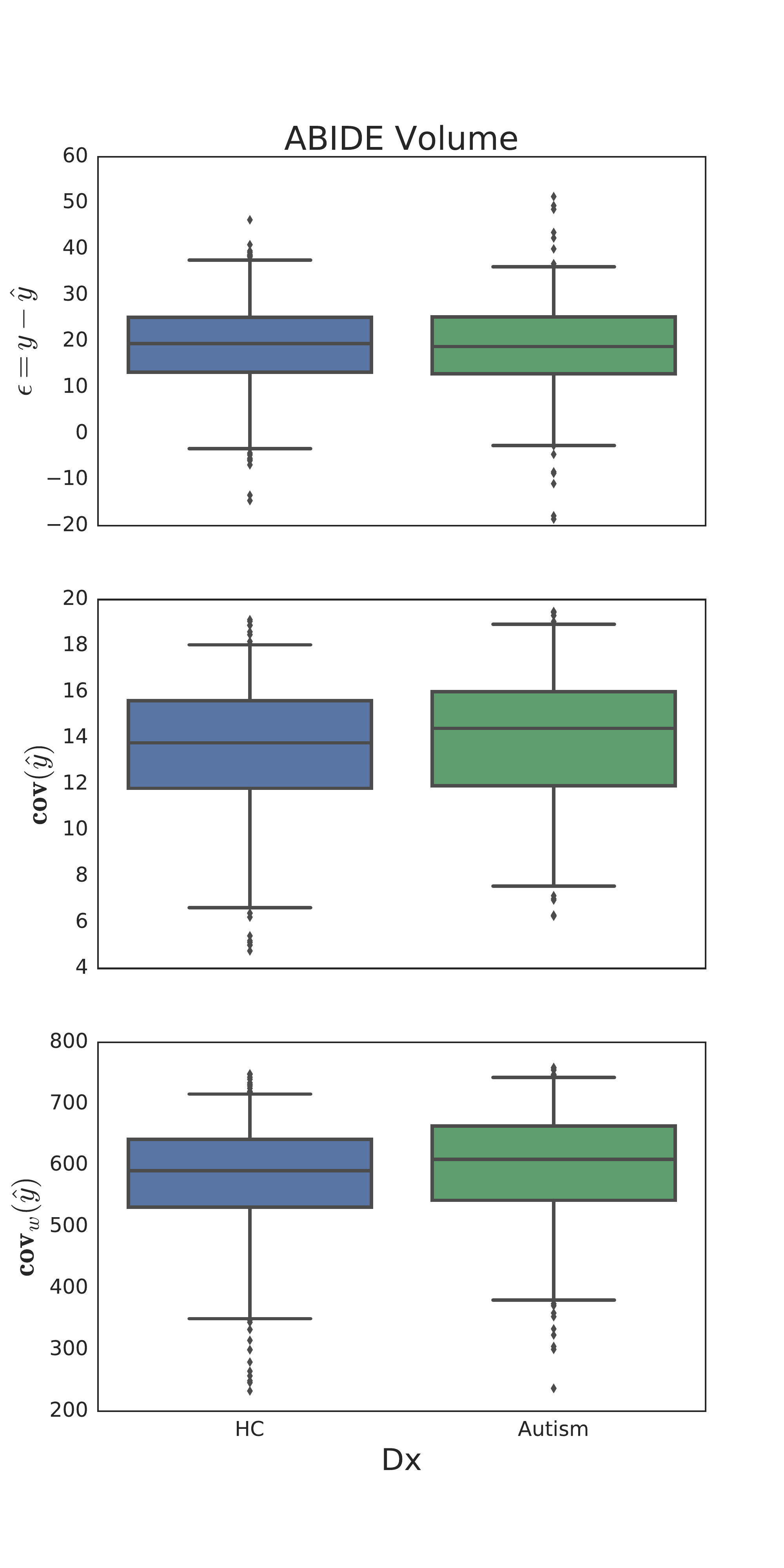}
	\end{subfigure}
	\begin{subfigure}{.245\textwidth}
		\centering\includegraphics[width=\textwidth]{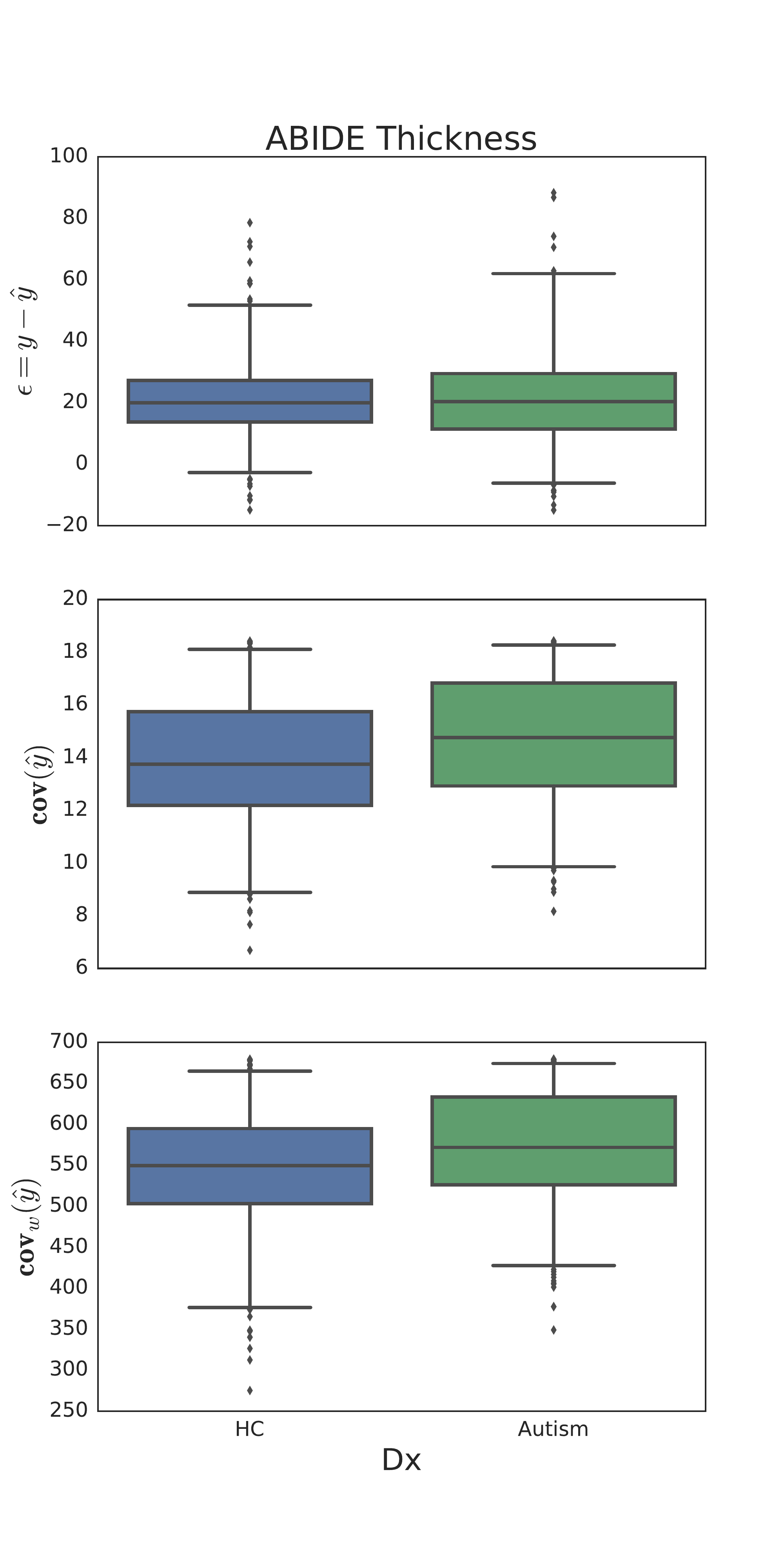}
	\end{subfigure}
	\begin{subfigure}{.245\textwidth}
		\centering\includegraphics[width=\textwidth]{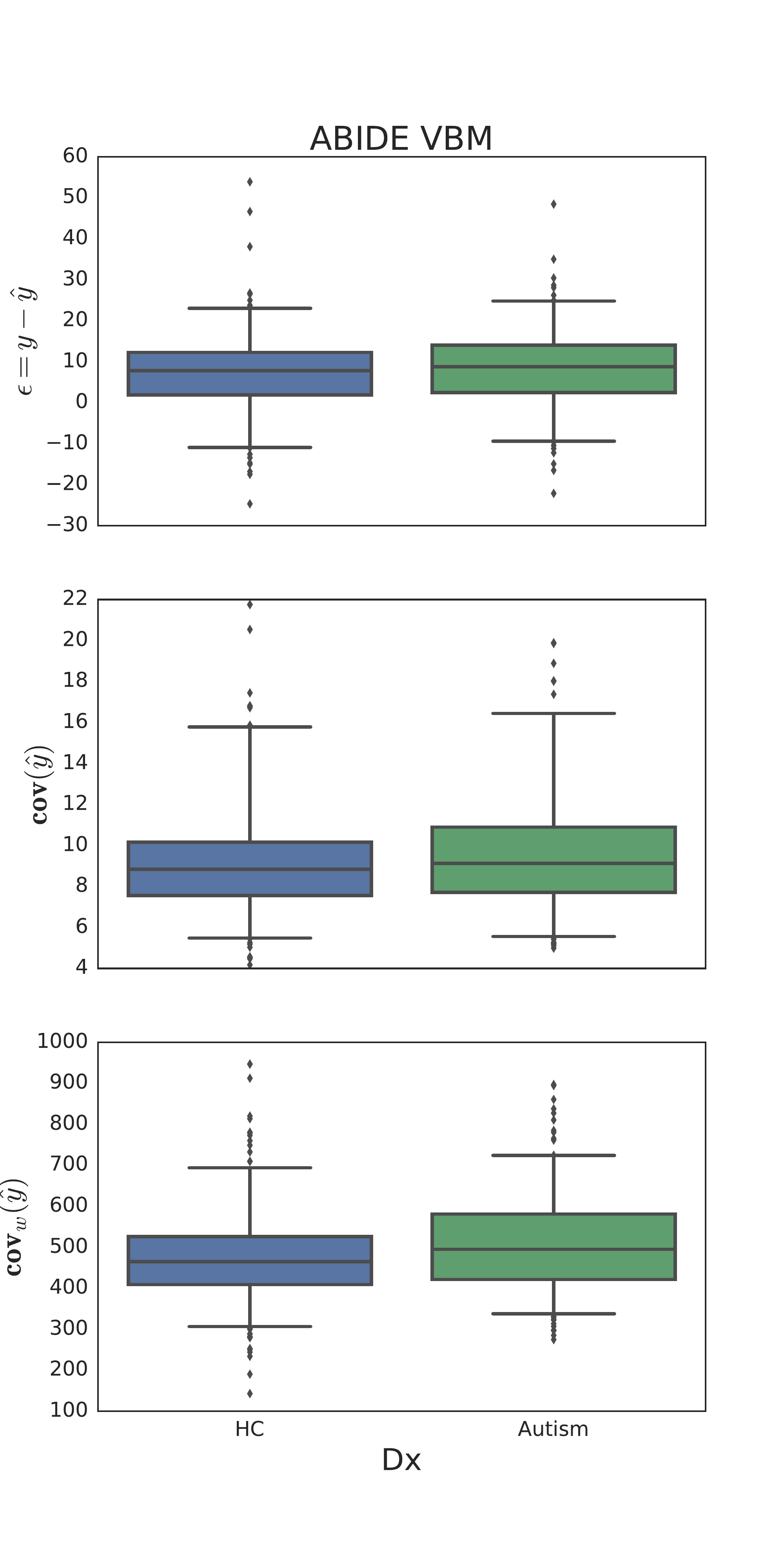}
	\end{subfigure}
	\begin{subfigure}{.245\textwidth}
		\centering\includegraphics[width=\textwidth]{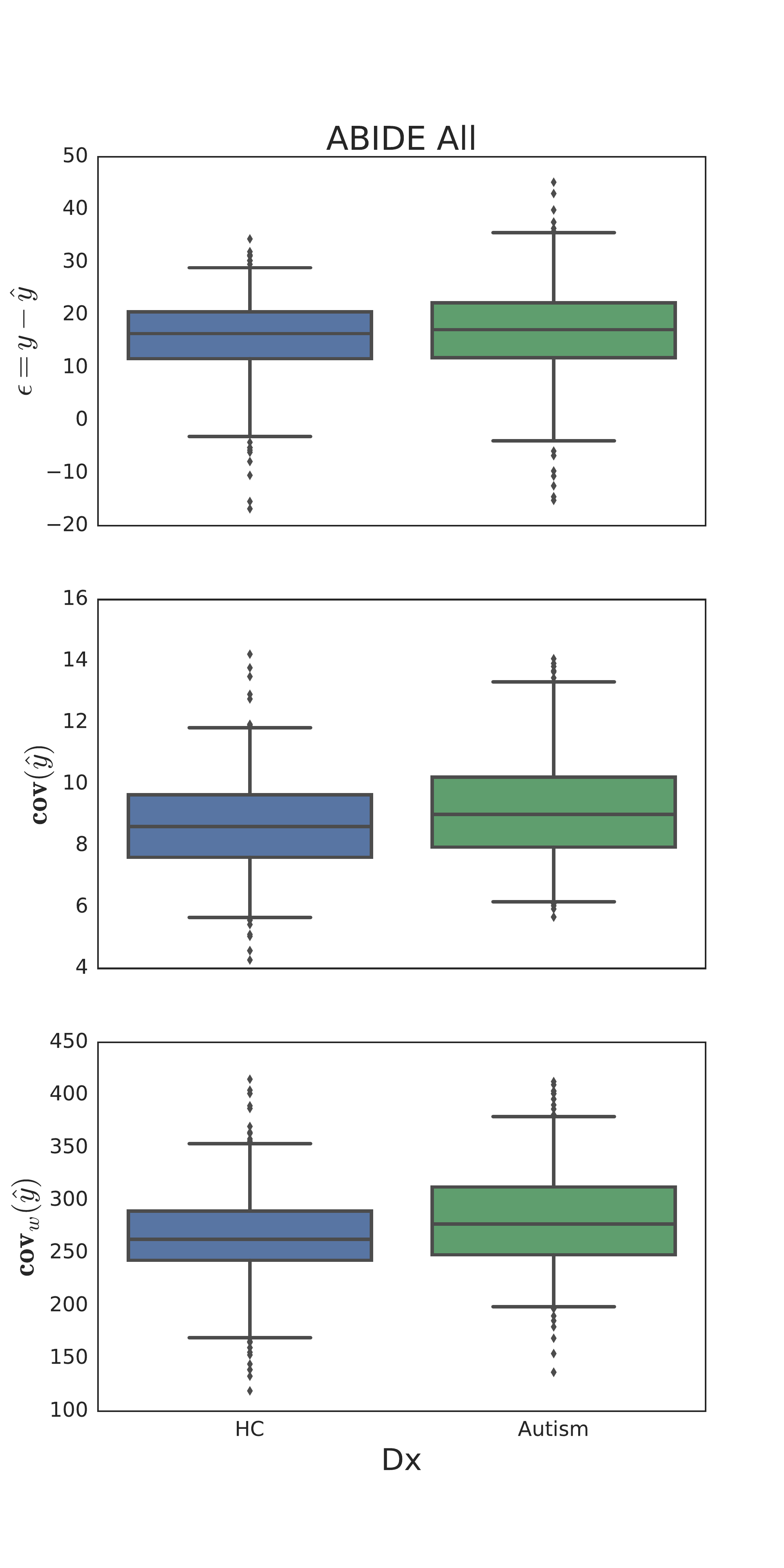}
	\end{subfigure}
	\caption{Box plots showing prediction results for $\epsilon$ (top), $\cov$ (middle) and $\covw$ (bottom)  for HC and Autism groups on the ABIDE II dataset. Columns correspond to the different evaluated features.}
	\label{fig:all_box_abide}
\end{figure*}

For our third experiment, we evaluate the age prediction on the ABIDE II dataset that contains subjects with autism. To the best of our knowledge, no previous age prediction based approach has been used for studying autism. However, previous studies have suggested abnormal brain development in patients diagnosed with autism \cite{courchesne2001unusual}. By observing figures \ref{fig:all_box_abide} and \ref{fig:roc_abide} it is clear that differences between HC and MCI groups are considerably less noticeable than in the previous two experiments. In fact, by analyzing the AUC results in table \ref{table:auc-abide} and the p-values in table \ref{table:pvalues-abide}, we can observe that no significant differences between groups were found using the standard prediction error approach using $\epsilon$ as a predictive variable. In contrast, both uncertainty based measurements showed significant differences between HC and autistic groups. Also different to the previous two experiments, the weighted uncertainty based measurement $\covw$ showed a better performance than the standard uncertainty $\cov$.  In this case correlation coefficients between the metrics were 0.64 for $\cov$-$\covw$, 0.30 for $\epsilon$-$\cov$ and 0.49 for $\epsilon$-$\covw$. The lower correlation between $\cov$-$\covw$  and the larger correlation between $\epsilon$-$\covw$ when compared to the two previous experiments are caused by the smaller value of $l_y$.

\begin{table}[h]
	\centering
	{
		\begin{tabular}{lccc}
			& $\epsilon$ & $\cov$ & $\covw$   \tabularnewline
			Volume  & \zz{0.3210} &\textless\zz{0.0317} &\textless\zz{0.0019}   \tabularnewline
			Thickness& \zz{0.4664} &\textless\zz{0.0001} &\textless\zz{0.0001}  \tabularnewline
			VBM  & \zz{0.0556} &\zz{0.0089} &\textless\zz{0.0001}   \tabularnewline
			All  & \zz{0.1060} &\textless\zz{0.0001} &\textless\zz{0.0001}  \tabularnewline			
		\end{tabular} 
	} \caption{p-values corresponding  to the statistical tests performed on the experiments comparing the HC, and Autism groups. The highlighted values correspond to p values with significance levels under 0.05 (light background), 0.01 (middle background) and 0.001 (dark background).}
	\label{table:pvalues-abide}
\end{table}
\begin{table}[h]
	\centering
	{
		\begin{tabular}{lccc}
			& $\epsilon$ & $\cov$ & $\covw$   \tabularnewline
			\hline
			\rowcolor{lightgray} 
			Volume  & 0.50 & 0.53 & \textbf{0.58}   \tabularnewline
			Thickness& 0.50 &\textbf{0.60} & \textbf{0.60}  \tabularnewline
			\rowcolor{lightgray} 
			VBM  & 0.53 &0.55 &\textbf{0.60}   \tabularnewline
			All  & 0.53 & 0.58 & \textbf{0.60}  \tabularnewline			
		\end{tabular} 
	} \caption{Area Under the Curve (AUC) values corresponding  to the statistical tests performed on the experiments comparing the HC, and Autism groups on the ABIDE II dataset.}
	\label{table:auc-abide}
\end{table}


\section{Discussion}
In this work, we have proposed to use uncertainty in GPR as a measure of neuropathology.  In contrast to previous work based on the prediction error, which assumes similar trajectories between aging and disease processes, the GPR uncertainty handles differences in morphology of diseased brains that do not necessarily lie on a healthy aging trajectory. If we consider predicted age as an aging biomarker, GPR uncertainty can be seen as a measure of uncertainty of the aging biomarker.
We have evaluated the ability of GPR uncertainty to discriminate subjects with pathology for two very different diseases: Alzheimer's disease where we worked with a cohort of advanced age individuals, and autism where we operated on a younger cohort. To the best of our knowledge, this is the first time that uncertainty in a GPR model is used as a measure of neuropathology and it is also the first application of an age regression model for autism. For both applications, we work with a single model that was trained on healthy subjects from a wide age range. This distinguishes our work from discriminative approaches, which require the inclusion of images of patients diagnosed with a particular disease in the training set.
	
In this work, we build age prediction models using three different types of brain features: VBM, volume and thickness as well as a combination of all of them. Based on our results, we have observed that none of the features outperformed the others across all our evaluations. However, combining all features resulted on a model with the lowest prediction error and consistently achieved the best results when performing separation between healthy and disease groups. This is in line with previous work \citep{Valizadeh2017, liem2017predicting}, where it has been observed that extended feature sets which give models  a larger variety of measurements to base the prediction on, result in more accurate age estimation models.  Although the main goal of this paper is not to present a state-of-the-art age prediction method, we have observed that our proposed GPR model has a high prediction accuracy, comparable to that of current age estimation approaches \citep{Valizadeh2017, Cole2016}.

We have also demonstrated the generalization ability of our method by training and testing our model in completely independent datasets. Our training dataset was built based on the IXI, ABIDE and AIBL databases while testing was performed on the OASIS, ADNI and ABIDE II databases. Using different datasets for training and testing complicates the age prediction problem, as undesired dataset biases can impact the result \citep{wachinger2016domain, gutierrez2017}. However, such experiments model a scenario that is more realistic, as the translation to the clinic requires the accurate deployment of our method on data that differs from the training set.  

Based on GPR uncertainty, we have introduced two metrics to assess the similarity of a test subject to a model of healthy aging: the uncertainty of the predictions of the GPR $\cov$ and an age-weighted uncertainty measurement $\covw$. We have shown in our experiments in section \ref{sec:results} that both measures find statistically significant differences between  HC, MCI and AD groups as well as between autism and HC groups. We have compared these results to the commonly used prediction error $\epsilon$, and we have shown that the proposed metrics yield a better separation between groups.  The age-weighted uncertainty measurement can be seen as an extension to the standard uncertainty measure, with the inclusion of a weighting parameter based on the chronological age of the test subject. The effect of this weighting is controlled by the age-length scale parameter $l_y$. We have analyzed the effect of $l_y$ in the performance of the age-weighted uncertainty measurement and we have observed that although the use of the age-weighting had a limited effect in the case of the MCI/AD experiments there was a clear improvement in the case of autism. We hypothesize that these differences in performance of $\covw$ can be attributed to the different age ranges of the testing cohorts, since age prediction models present smaller prediction errors when testing on younger cohorts compared to the prediction error presented on datasets consisting of older individuals \citep{Cole2017}.

For the experiment on autism,  our proposed uncertainty based metrics showed its ability to discriminate between autistic and healthy groups. We find these results particularly encouraging since the prediction error based approach showed to be insufficient to find differences for this particular disease. Given the analysis performed in section \ref{sec:progress} and the results of our experiments, we believe that the main reason of the better performance of our uncertainty based measures is that they do not model brain anomaly  as an accelerated aging process, but rather as deviations from healthy aging. As discussed before, the complex effects of aging and disease follow trajectories that affect different areas of the brain at different rates. The more relaxed assumptions, which our proposed uncertainty based measures are based on, are therefore better suited to account for the complex impact of aging and disease across the entire brain.

We have not performed direct quantitative comparisons between our uncertainty based measures and  discriminative approaches. The main reason behind this is that discriminative approaches require training images  not only from healthy individuals but also from patients. This means that separate models have to be trained for each specific disease. In contrast, age-prediction based models are only built on images from healthy individuals. This allows to have a flexible model which can be used for different diseases without any need to retrain or adjust the model. We demonstrated this in our experiments, where we used the same age prediction model to predict brain anomaly both on patients with Alzheimer's disease and patients with autism.


\section{Conclusions}

We introduced the prediction uncertainty in age estimation as a measure of neuropathology, based on a multivariate age prediction model based on Gaussian process regression. Our measure does not make a priori specific assumptions about the nature of the changes caused by disease, but rather models these changes as deviations from healthy aging. The method is therefore not limited to a specific pathology or age range, as demonstrated in our experiments on patients with Alzheimer's disease and patients with autism. Our method is also flexible to work with different sets of features, as we have illustrated in our experiments using volume, thickness, and VBM features. We have introduced an extension of the Gaussian process uncertainty measure for age estimation that also takes the chronological age into account, resulting in a weighted uncertainty measure, and we have demonstrated that the inclusion of this weighted measure can potentially be helpful for some applications. 
In comparison to the commonly used prediction error, the prediction uncertainty yielded an improved separation of diagnostic groups across all feature types and for different applications. It is also important to point out that in contrast to discriminative approaches, age prediction based models only require images of healthy individuals for training, which may allow for incorporating scans from large population-based studies in the future. The results presented in this paper encourage us to further explore the potential of uncertainty based measures and to apply our method to different diseases or conditions that might have complex effects in the anatomy of the brain. We are further interested in investigating the relationship between the prediction uncertainty and  cognitive and clinical characteristics, as well as, future health outcomes.

\section*{Acknowledgements}

This work was partly funded by SAP SE, the F\"orderprogramm f\"ur Forschung und Lehre, the Bavarian State Ministry of Education, Science and the Arts in the framework of the Centre Digitisation.Bavaria (ZD.B) and the Faculty of Medicine of the University Munich (F\"orderprogramm f\"ur Forschung \& Lehre). Data collection and sharing for this project was funded by the Alzheimer's DiseaseNeuroimaging Initiative (ADNI) (National Institutes of Health Grant U01 AG024904) and DOD ADNI (Department of Defense award number W81XWH-12-2-0012). ADNI is funded by the National Institute on Aging, the National Institute of Biomedical Imaging and
Bioengineering, and through generous contributions from the following: AbbVie, Alzheimer's Association; Alzheimer's Drug Discovery Foundation; Araclon Biotech; BioClinica, Inc.; Biogen; Bristol-Myers Squibb Company; CereSpir, Inc.; Cogstate; Eisai Inc.; Elan Pharmaceuticals, Inc.; Eli Lilly and Company; EuroImmun; F. Hoffmann-La Roche Ltd and its affiliated company Genentech, Inc.; Fujirebio; GE Healthcare; IXICO Ltd.; Janssen Alzheimer Immunotherapy Research \& Development, LLC.; Johnson \& Johnson Pharmaceutical Research \& Development LLC.; Lumosity; Lundbeck; Merck \& Co., Inc.; Meso Scale Diagnostics, LLC.; NeuroRx Research; Neurotrack Technologies; Novartis Pharmaceuticals Corporation; Pfizer Inc.; Piramal Imaging; Servier; Takeda Pharmaceutical Company; and Transition Therapeutics. The Canadian Institutes of Health Research is providing funds to support ADNI clinical sites in Canada. Private sector contributions are facilitated by the Foundation for the National Institutes of Health (www.fnih.org). The grantee organization is the Northern California Institute for Research and Education, and the study is
coordinated by the Alzheimer's Therapeutic Research Institute at the University of Southern California. ADNI data are disseminated by the Laboratory for Neuro Imaging at the University of Southern California.

\bibliographystyle{elsarticle-harv}
\bibliography{biblio}{}

\begin{thebibliography}{34}
\expandafter\ifx\csname natexlab\endcsname\relax\def\natexlab#1{#1}\fi
\expandafter\ifx\csname url\endcsname\relax
  \def\url#1{\texttt{#1}}\fi
\expandafter\ifx\csname urlprefix\endcsname\relax\def\urlprefix{URL }\fi

\bibitem[{Buckner(2004)}]{buckner2004}
Buckner, R.~L., 2004. Memory and executive function in aging and ad. Neuron
  44~(1), 195--208.

\bibitem[{Cole and Franke(2017)}]{Cole2017}
Cole, J.~H., Franke, K., 2017. Predicting age using neuroimaging: Innovative
  brain ageing biomarkers. Trends in Neurosciences.

\bibitem[{Cole et~al.(2015)Cole, Leech, and Sharp}]{cole2015prediction}
Cole, J.~H., Leech, R., Sharp, D.~J., 2015. Prediction of brain age suggests
  accelerated atrophy after traumatic brain injury. Annals of neurology 77~(4),
  571--581.

\bibitem[{Cole et~al.(2016)Cole, Poudel, Tsagkrasoulis, Caan, Steves, Spector,
  and Montana}]{Cole2016}
Cole, J.~H., Poudel, R.~P., Tsagkrasoulis, D., Caan, M.~W., Steves, C.,
  Spector, T.~D., Montana, G., 2016. {Predicting brain age with deep learning
  from raw imaging data results in a reliable and heritable biomarker}.
\newline\urlprefix\url{http://arxiv.org/abs/1612.02572}

\bibitem[{Courchesne et~al.(2001)Courchesne, Karns, Davis, Ziccardi, Carper,
  Tigue, Chisum, Moses, Pierce, Lord, et~al.}]{courchesne2001unusual}
Courchesne, E., Karns, C., Davis, H., Ziccardi, R., Carper, R., Tigue, Z.,
  Chisum, H., Moses, P., Pierce, K., Lord, C., et~al., 2001. Unusual brain
  growth patterns in early life in patients with autistic disorder an mri
  study. Neurology 57~(2), 245--254.

\bibitem[{Di~Martino et~al.(2014)Di~Martino, Yan, Li, Denio, Castellanos,
  Alaerts, Anderson, Assaf, Bookheimer, Dapretto, et~al.}]{ABIDE}
Di~Martino, A., Yan, C.-G., Li, Q., Denio, E., Castellanos, F.~X., Alaerts, K.,
  Anderson, J.~S., Assaf, M., Bookheimer, S.~Y., Dapretto, M., et~al., 2014.
  The autism brain imaging data exchange: towards large-scale evaluation of the
  intrinsic brain architecture in autism. Molecular psychiatry 19~(6), 659.

\bibitem[{Ellis et~al.(2009)Ellis, Bush, Darby, De~Fazio, Foster, Hudson,
  Lautenschlager, Lenzo, Martins, Maruff, et~al.}]{aibl}
Ellis, K.~A., Bush, A.~I., Darby, D., De~Fazio, D., Foster, J., Hudson, P.,
  Lautenschlager, N.~T., Lenzo, N., Martins, R.~N., Maruff, P., et~al., 2009.
  The australian imaging, biomarkers and lifestyle (aibl) study of aging:
  methodology and baseline characteristics of 1112 individuals recruited for a
  longitudinal study of alzheimer's disease. International Psychogeriatrics
  21~(4), 672--687.

\bibitem[{Fischl(2012)}]{Fischl2012}
Fischl, B., 2012. Freesurfer. Neuroimage 62~(2), 774--781.

\bibitem[{Fjell et~al.(2014)Fjell, Westlye, Grydeland, Amlien, Reinvang, Raz,
  Holland, Dale, Walhovd, and Neuroimaging}]{Fjell2014}
Fjell, A.~M., Westlye, L.~T., Grydeland, H., Amlien, I., Reinvang, I., Raz, N.,
  Holland, D., Dale, A.~M., Walhovd, K.~B., Neuroimaging, D., 2014. {Critical
  ages in the life-course of the adult brain: nonlinear subcortical aging}
  34~(10), 2239--2247.

\bibitem[{Franke et~al.(2010)Franke, Ziegler, Kl{\"{o}}ppel, and
  Gaser}]{Franke2010}
Franke, K., Ziegler, G., Kl{\"{o}}ppel, S., Gaser, C., 2010. {Estimating the
  age of healthy subjects from T1-weighted MRI scans using kernel methods:
  Exploring the influence of various parameters}. Neuroimage 50~(3), 883--892.
\newline\urlprefix\url{http://dx.doi.org/10.1016/j.neuroimage.2010.01.005}

\bibitem[{Gaser et~al.(2013)Gaser, Franke, Kl{\"{o}}ppel, Koutsouleris, and
  Sauer}]{Gaser2013}
Gaser, C., Franke, K., Kl{\"{o}}ppel, S., Koutsouleris, N., Sauer, H., 2013.
  {BrainAGE in Mild Cognitive Impaired Patients: Predicting the Conversion to
  Alzheimer's Disease}. PLoS One 8~(6).

\bibitem[{Good et~al.(2001)Good, Johnsrude, Ashburner, Henson, Friston, and
  Frackowiak}]{Good2001}
Good, C.~D., Johnsrude, I.~S., Ashburner, J., Henson, R.~N., Friston, K.~J.,
  Frackowiak, R.~S., 2001. {A Voxel-Based Morphometric Study of Ageing in 465
  Normal Adult Human Brains}. Neuroimage 14~(1), 21--36.
\newline\urlprefix\url{http://linkinghub.elsevier.com/retrieve/pii/S1053811901907864}

\bibitem[{Guti{\'e}rrez et~al.(2017)Guti{\'e}rrez, Peter, Klein, and
  Wachinger}]{gutierrez2017}
Guti{\'e}rrez, B., Peter, L., Klein, T., Wachinger, C., 2017. A multi-armed
  bandit to smartly select a training set from big medical data. In:
  International Conference on Medical Image Computing and Computer-Assisted
  Intervention. Springer, pp. 38--45.

\bibitem[{Guttmann et~al.(1998)Guttmann, Jolesz, Kikinis, Killiany, Moss,
  Sandor, and Albert}]{Guttmann1998}
Guttmann, C., Jolesz, F.~A., Kikinis, R., Killiany, R.~J., Moss, M.~B., Sandor,
  T., Albert, M.~S., 1998. {White matter changes with normal aging}. Neurology
  50~(4), 972--978.
\newline\urlprefix\url{http://www.neurology.org/cgi/doi/10.1212/WNL.50.4.972}

\bibitem[{Habes et~al.(2016)Habes, Janowitz, Erus, Toledo, Resnick, Doshi, {Van
  der Auwera}, Wittfeld, Hegenscheid, Hosten, Biffar, Homuth, V{\"{o}}lzke,
  Grabe, Hoffmann, and Davatzikos}]{Habes2016}
Habes, M., Janowitz, D., Erus, G., Toledo, J.~B., Resnick, S.~M., Doshi, J.,
  {Van der Auwera}, S., Wittfeld, K., Hegenscheid, K., Hosten, N., Biffar, R.,
  Homuth, G., V{\"{o}}lzke, H., Grabe, H.~J., Hoffmann, W., Davatzikos, C.,
  2016. {Advanced brain aging: relationship with epidemiologic and genetic risk
  factors, and overlap with Alzheimer disease atrophy patterns.} Transl.
  Psychiatry 6~(February), e775.
\newline\urlprefix\url{http://www.ncbi.nlm.nih.gov/pubmed/27045845}

\bibitem[{Hedden and Gabrieli(2004)}]{hedden2004}
Hedden, T., Gabrieli, J.~D., 2004. Insights into the ageing mind: a view from
  cognitive neuroscience. Nature reviews neuroscience 5~(2), 87--96.

\bibitem[{Ingalhalikar et~al.(2014)Ingalhalikar, Smith, Parker, Satterthwaite,
  Elliott, Ruparel, Hakonarson, Gur, Gur, and Verma}]{ingalhalikar2014}
Ingalhalikar, M., Smith, A., Parker, D., Satterthwaite, T.~D., Elliott, M.~A.,
  Ruparel, K., Hakonarson, H., Gur, R.~E., Gur, R.~C., Verma, R., 2014. Sex
  differences in the structural connectome of the human brain. Proceedings of
  the National Academy of Sciences 111~(2), 823--828.

\bibitem[{Jack et~al.(2008)Jack, Bernstein, Fox, Thompson, Alexander, Harvey,
  Borowski, Britson, L~Whitwell, Ward, et~al.}]{adni}
Jack, C.~R., Bernstein, M.~A., Fox, N.~C., Thompson, P., Alexander, G., Harvey,
  D., Borowski, B., Britson, P.~J., L~Whitwell, J., Ward, C., et~al., 2008. The
  alzheimer's disease neuroimaging initiative (adni): Mri methods. Journal of
  magnetic resonance imaging 27~(4), 685--691.

\bibitem[{Kondo et~al.(2015)Kondo, Ito, {Kai Wu}, Sato, Taki, Fukuda, and
  Aoki}]{Kondo2015}
Kondo, C., Ito, K., {Kai Wu}, Sato, K., Taki, Y., Fukuda, H., Aoki, T., 2015.
  {An age estimation method using brain local features for T1-weighted images}.
  2015 37th Annu. Int. Conf. IEEE Eng. Med. Biol. Soc., 666--669.
\newline\urlprefix\url{http://ieeexplore.ieee.org/lpdocs/epic03/wrapper.htm?arnumber=7318450}

\bibitem[{Koutsouleris et~al.(2013)Koutsouleris, Davatzikos, Borgwardt, Gaser,
  Bottlender, Frodl, Falkai, Riecher-R{\"o}ssler, M{\"o}ller, Reiser,
  et~al.}]{koutsouleris2013accelerated}
Koutsouleris, N., Davatzikos, C., Borgwardt, S., Gaser, C., Bottlender, R.,
  Frodl, T., Falkai, P., Riecher-R{\"o}ssler, A., M{\"o}ller, H.-J., Reiser,
  M., et~al., 2013. Accelerated brain aging in schizophrenia and beyond: a
  neuroanatomical marker of psychiatric disorders. Schizophrenia bulletin
  40~(5), 1140--1153.

\bibitem[{Liem et~al.(2017)Liem, Varoquaux, Kynast, Beyer, Masouleh,
  Huntenburg, Lampe, Rahim, Abraham, Craddock, et~al.}]{liem2017predicting}
Liem, F., Varoquaux, G., Kynast, J., Beyer, F., Masouleh, S.~K., Huntenburg,
  J.~M., Lampe, L., Rahim, M., Abraham, A., Craddock, R.~C., et~al., 2017.
  Predicting brain-age from multimodal imaging data captures cognitive
  impairment. NeuroImage 148, 179--188.

\bibitem[{Mann and Whitney(1947)}]{mann1947}
Mann, H.~B., Whitney, D.~R., 1947. On a test of whether one of two random
  variables is stochastically larger than the other. The annals of mathematical
  statistics, 50--60.

\bibitem[{Marcus et~al.(2007)Marcus, Wang, Parker, Csernansky, Morris, and
  Buckner}]{oasis}
Marcus, D.~S., Wang, T.~H., Parker, J., Csernansky, J.~G., Morris, J.~C.,
  Buckner, R.~L., 2007. Open access series of imaging studies (oasis):
  cross-sectional mri data in young, middle aged, nondemented, and demented
  older adults. Journal of cognitive neuroscience 19~(9), 1498--1507.

\bibitem[{Nenadic et~al.(2017)Nenadic, Dietzek, Langbein, Sauer, and
  Gaser}]{Nenadic2017}
Nenadic, I., Dietzek, M., Langbein, K., Sauer, H., Gaser, C., 2017. {BrainAGE
  score indicates accelerated brain aging in schizophrenia, but not bipolar
  disorder}. Psychiatry Res. Neuroimaging 266~(March), 86--89.
\newline\urlprefix\url{http://linkinghub.elsevier.com/retrieve/pii/S0925492717301002}

\bibitem[{Pedregosa et~al.(2011)Pedregosa, Varoquaux, Gramfort, Michel,
  Thirion, Grisel, Blondel, Prettenhofer, Weiss, Dubourg, Vanderplas, Passos,
  Cournapeau, Brucher, Perrot, and Duchesnay}]{scikit}
Pedregosa, F., Varoquaux, G., Gramfort, A., Michel, V., Thirion, B., Grisel,
  O., Blondel, M., Prettenhofer, P., Weiss, R., Dubourg, V., Vanderplas, J.,
  Passos, A., Cournapeau, D., Brucher, M., Perrot, M., Duchesnay, E., 2011.
  Scikit-learn: Machine learning in {P}ython. Journal of Machine Learning
  Research 12, 2825--2830.

\bibitem[{Potvin et~al.(2017)Potvin, Dieumegarde, and Duchesne}]{Potvin2017}
Potvin, O., Dieumegarde, L., Duchesne, S., 2017. {Normative Morphometric Data
  for Cerebral Cortical Areas Over the Lifetime of the Adult Human Brain}.
  Neuroimage.
\newline\urlprefix\url{http://linkinghub.elsevier.com/retrieve/pii/S1053811917304135}

\bibitem[{Rasmussen and Williams(2005)}]{Rasmussen2005}
Rasmussen, C.~E., Williams, C. K.~I., 2005. Gaussian Processes for Machine
  Learning (Adaptive Computation and Machine Learning). The MIT Press.

\bibitem[{Steffener et~al.(2016)Steffener, Habeck, O'Shea, Razlighi, Bherer,
  and Stern}]{steffener2016}
Steffener, J., Habeck, C., O'Shea, D., Razlighi, Q., Bherer, L., Stern, Y.,
  2016. Differences between chronological and brain age are related to
  education and self-reported physical activity. Neurobiology of aging 40,
  138--144.

\bibitem[{Valizadeh et~al.(2017)Valizadeh, Hanggi, Merillat, and
  J\"ancke}]{Valizadeh2017}
Valizadeh, S.~A., Hanggi, J., Merillat, S., J\"ancke, L., 2017. {Age prediction
  on the basis of brain anatomical measures}. Hum. Brain Mapp. 38~(2),
  997--1008.

\bibitem[{Wachinger et~al.(2015)Wachinger, Golland, Kremen, Fischl, and
  Reuter}]{Wachinger2015}
Wachinger, C., Golland, P., Kremen, W., Fischl, B., Reuter, M., 2015.
  Brainprint: a discriminative characterization of brain morphology. NeuroImage
  109, 232--248.

\bibitem[{Wachinger and Reuter(2016)}]{wachinger2016domain}
Wachinger, C., Reuter, M., 2016. Domain adaptation for alzheimer's disease
  diagnostics. Neuroimage 139, 470--479.

\bibitem[{Wachinger et~al.(2016)Wachinger, Salat, Weiner, and
  Reuter}]{Wachinger2016}
Wachinger, C., Salat, D.~H., Weiner, M., Reuter, M., 2016. Whole-brain analysis
  reveals increased neuroanatomical asymmetries in dementia for hippocampus and
  amygdala. Brain 139~(12), 3253--3266.

\bibitem[{Wang et~al.(2014)Wang, Li, Miao, Dai, Hua, and He}]{Wang2014}
Wang, J., Li, W., Miao, W., Dai, D., Hua, J., He, H., 2014. {Age estimation
  using cortical surface pattern combining thickness with curvatures}. Med.
  Biol. Eng. Comput. 52~(4), 331--341.

\bibitem[{Ziegler et~al.(2012)Ziegler, Dahnke, Gaser, Initiative,
  et~al.}]{Ziegler2012}
Ziegler, G., Dahnke, R., Gaser, C., Initiative, A. D.~N., et~al., 2012. Models
  of the aging brain structure and individual decline. Frontiers in
  Neuroinformatics 6~(3).

\end{thebibliography}

\end{document}